%% file: thesis.tex
\documentclass[phd, titlesmallcaps, examinerscopy, copyrightpage, foronline]{mqthesis}

\usepackage{array,multirow}
\usepackage{rotating}  
\usepackage{algorithm}
\usepackage{algcompatible}
\usepackage{bbm}
\usepackage[noend]{algpseudocode}
\usepackage[flushleft]{threeparttable}
\usepackage{tabularx}
\usepackage{theorem}   
\usepackage{bm}  
\usepackage{amsmath,amsfonts,amssymb}
\usepackage{booktabs}  
\pdfoutput=1 
\usepackage{pdfsync}   

\newtheorem{definition}{Definition}

\usepackage[]{inputenc}

\begin{document}
\pdfoutput=1 
\frontmatter

\title{Towards Personalised and Human-in-the-Loop Document Summarisation}

\ifthenelse{\boolean{foronline}}{
  \author{\href{samira.ghodratnama@hdr.mq.edu.au}{Samira Ghodratnama}}
  \department{Computing}
}{
  \author{Samira Ghodratnama}
  \department{Computing}
}

\degrees{PhD in Computer Science}

 \submitdate{May 2021}


\titlepage
\input{dedication}
\input{acknowledge}
\input{Examiners}

\input{listofpublications}

\input{abstract}

\tableofcontents
\listoffigures
\listoftables

\mainmatter

\input{ch_1/chap_intro}
\input{ch_2/ch_background}

\input{ch_3/ch_ExperimentalSetup}
\input{ch_4/FeatureEngineering}
\input{ch_5/InteractiveSummarization}
\input{ch_6/PersonalizedSummarization}
\input{ch_7/Application}
\input{ch_8/Conclusion}

\appendix

 
\backmatter


\bibliography{references}

\end{document}

%% file: dedication.tex
\chapter{Dedication}
I dedicate this dissertation to three beloved people who have meant and continue to mean so much to me.

The first is my father, for his endless love and support.
He always trusts my abilities and respects all my decisions, giving me the confidence to follow my dreams.
He taught me to believe in hard work and inspired me to be strong despite many life obstacles.

In memory of my mother, Shirin, who, though no longer in this world, continues to regulate my life. 
Her words of encouragement and push for commitment always remain in my mind.

Last but not least, I dedicate this to my husband, Mehrdad, who always encourages me to accept every adventure life offers, especially this one.
He taught me to think big and view every challenge as an opportunity to achieve what I aspire.
I am genuinely thankful for having you in my life and exploring the world with you, Mehrdad!

%% file: acknowledge.tex
\chapter{Acknowledgements}
I would like to take this opportunity to acknowledge those who have impacted my PhD accomplishment and doctorate journey.

First, I express my  deep gratitude and sincere appreciation to my Principal Supervisor, Dr Amin Beheshti, for his guidance, encouragement and support he has provided throughout my PhD program. 
His positive outlook inspired me and gave me the confidence to explore new research directions.
He gave me the academic freedom to follow my ideas, and his insightful feedback pushed me to sharpen my thinking and elevated my work to a higher level.
I have been extremely lucky to have a supervisor who cared so much and enriched my growth as a student, a researcher and a scientist.

I would like to thank my associate supervisor, Dr Jia Wu, whose expertise were valuable in formulating the research questions and methodology. 
I would also like to thank Dr Fariborz Sobhanmanesh and Dr Hamid Reza Motahari-Nezhad for their valuable guidance and thoughtful comments throughout my studies.
I extend further thanks to Dr Mohamed Sarwat at Arizona State University for being such a kind and understanding host during my research visit, especially given COVID-19. 

My sincere appreciation also goes to Macquarie University and the AI-enabled Processes Research Centre for generously funding this PhD research through the Macquarie Research Excellence Scholarship and the Postgraduate Research Fund.
These scholarships made it possible for me to pursue my doctoral studies.

I am thankful to every member of Data Analytics Research Lab at Macquarie University, who supported me in my research.
In addition, I would like to thank the anonymous reviewers who provided helpful feedback and suggestions on my publications.

I would also like to acknowledge Capstone Editing for providing copyediting and proofreading services, according to the guidelines laid out in the university-endorsed national `Guidelines for Editing Research Theses'.

Finally, I am thankful to my family and my in-laws for their love and patience over the years and during my research endeavours\textemdash especially my cute and sweet niece, Kimia, for her love, patience and understanding when I was so busy doing my research and I did not have time to spend with her.
They are my source of strength, and without their love and endless support this dissertation would have neither been started nor completed.\newline

\noindent Samira Ghodratnama\newline
\noindent May 2021

%% file: Examiners.tex
\chapter{Dissertation Examiners}

\begin{itemize}
    \item Professor Ladjel Bellatreche, ISAE-ENSMA, France
    \item Professor Ismail Khalil, Johannes Kepler University Linz, Austria
    \item Professor Salil Kanhere, UNSW Sydney, Australia
\end{itemize}

%% file: listofpublications.tex
\chapter{List of Publications}

\begin{itemize}

	
        
\item[$\bullet$] \textbf{Ghodratnama, Samira}, Mehrdad Zakershahrak, and Amin Beheshti. "Summary2vec: Learning Semantic Representation of Summaries for Healthcare Analytic.", IJCNN 2021: International Joint Conference on Neural Network, 2021. (\textbf{CORE Rank: A}). 

\item[$\bullet$] \textbf{Ghodratnama, Samira}, Amin Beheshti, Mehrdad Zakershahrak, and Fariborz Sobhanmanesh. "Intelligent Narrative Summaries: \\From Indicative to Informative Summarization", Big Data Research Journal, 2021. (\textbf{Q1, IF: 3.58}). 
\item[$\bullet$] \textbf{Ghodratnama, Samira}, Amin Beheshti, Mehrdad Zakershahrak, and Fariborz Sobhanmanesh. "Extractive Document Summarization based on Dynamic Feature Space Mapping.", IEEE Access, 8, pp.139084-139095, 2020. (\textbf{Q1, IF: 3.74}). 

\item[$\bullet$] \textbf{Ghodratnama, Samira}, Mehrdad Zakershahrak, and Fariborz Sobhanmanesh. "Am I Rare? An Intelligent Summarization Approach for Identifying Hidden Anomalies." AI-PA 2020 - 18th International Conference on Service Oriented Computing (ICSOC) Workshop, 2020. (\textbf{CORE Rank: A}). 

\item[$\bullet$] \textbf{Ghodratnama, Samira}, Mehrdad Zakershahrak, and Fariborz Sobhanmanesh. "Adaptive Summaries: A Personalized Concept-based Summarization Approach by Learning from Users' Feedback. AI-PA 2020 - 18th International Conference on Service Oriented Computing (ICSOC) Workshop, 2020. (\textbf{CORE Rank: A}). 

\item[$\bullet$] \textbf{Ghodratnama, Samira}, Amin Beheshti, and Mehrdad Zakershahrak. "A Personalized Reinforcement Learning Summarization Method for Learning Structure from Unstructured Data." submitted to Web Information Systems Engineering (WISE), 2021. (\textbf{CORE Rank: A}). 

\item[$\bullet$]\textbf{Ghodratnama, Samira}, Amin Beheshti, Mehrdad Zakershahrak, and Hamid Reza Motahari-Nezhad. "SumRecom: A Personalized Interactive Summary Recommendation System by Learning from Users' Feedback using Reinforcement Learning." submitted to Machine Learning Journal, 2021.  (\textbf{Q1, IF: 2.67}). 


\item[$\bullet$] Beheshti, Amin, Francesco Schiliro, \textbf{Samira Ghodratnama}, Farhad Amouzgar, Boualem Benatallah, Jian Yang, Quan Z. Sheng, Fabio Casati, and Hamid Reza Motahari-Nezhad. \emph{iProcess: enabling IoT platforms in data-driven knowledge-intensive processes.}, In International Conference on Business Process Management, pp. 108-126. Springer, Cham, 2018. (\textbf{CORE Rank: A}).   

\item[$\bullet$] Amouzgar, Farhad, Amin Beheshti, \textbf{Samira Ghodratnama}, Boualem Benatallah, Jian Yang, and Quan Z. Sheng. "iSheets: a spreadsheet-based machine learning development platform for data-driven process analytics." In International Conference on Service-Oriented Computing, pp. 453-457. Springer, Cham, 2018. (\textbf{CORE Rank: A}). 

\item[$\bullet$] Schiliro, Francesco, Amin Beheshti, \textbf{Samira Ghodratnama}, Farhad Amouzgar, Boualem Benatallah, Jian Yang, Quan Z. Sheng, Fabio Casati, and Hamid Reza Motahari-Nezhad. "iCOP: IoT-enabled policing processes." In International Conference on Service-Oriented Computing, pp. 447-452. Springer, Cham, 2018. (\textbf{CORE Rank: A}). 

\item[$\bullet$] Amin Beheshti, Boualem Benatallah, Hamid Reza Motahari-Nezhad, \textbf{Samira Ghodratnama}, and Fred Amouzgar,. "BP-SPARQL: A Query Language for Summarizing and Analyzing Big Process Data", Springer, 2020. (\textbf{Book Chapter}).

\end{itemize}

%% file: abstract.tex
\chapter{Abstract}
The ubiquitous availability of computing devices and the widespread use of the internet have generated a large amount of data continuously.
Therefore, the amount of available information on any given topic is far beyond humans’ processing capacity to properly process, causing what is known as ‘information overload’.
To efficiently cope with large amounts of information and generate content with significant value to users, we require identifying, merging and summarising information.
Data summaries can help gather related information and collect it into a shorter format that enables answering complicated questions, gaining new insight and discovering conceptual boundaries.

This thesis focuses on three main challenges to alleviate information overload using novel summarisation techniques.
It further intends to facilitate the analysis of documents to support personalised information extraction.
This thesis separates the research issues into four areas, covering (i) feature engineering in document summarisation, (ii) traditional static and inflexible summaries, (iii) traditional generic summarisation approaches, and (iv) the need for reference summaries.
We propose novel approaches to tackle these challenges, by:
\begin{itemize}
    \item enabling automatic intelligent feature engineering
    \item enabling flexible and interactive summarisation
    \item utilising intelligent and personalised summarisation approaches.
\end{itemize}
The experimental results prove the efficiency of the proposed approaches compared to other state-of-the-art models. 
We further propose solutions to the information overload problem in different domains through summarisation, covering network traffic data, health data and business process data.


%% file: ch_1/chap_intro.tex
\begin{savequote}[10cm] 
\sffamily
‘It’s not information overload. It’s filter failure.’
\qauthor{Clay Shirky}
\end{savequote}
\chapter{Introduction}
The way information is being generated, stored, manipulated, retrieved and disseminated has changed dramatically in recent decades.
The ubiquitous availability of computing devices and the widespread use of the internet have seen many data being generated continuously.
Therefore, the amount of available information on any given topic is far beyond humans’ processing capacity to manage, causing ‘information overload’.
This is commonly caused by~\cite{hoq2014information}:
\begin{itemize}
    \item multiple information sources presence, or information overabundance
    \item information managing challenge, or irrelevant obtained information
    \item users’ lack of time to analyse and understand gathered data.
\end{itemize}

Although this considerable amount of data can be useful in analytic applications, information is not valuable unless the knowledge can be derived.
However, no processing technique is possible on raw data until it becomes meaningful and relevant.
Indeed, there are many practical scenarios in which users need to process extensive textual documents with a specific goal.
Examples include lawyers facing an extensive collection of legal documents to find and process relevant information for a case to derive arguments and conclusions.
Researchers also need to read vast amounts of published scientific articles and find connections and trends across the content.
Obviously, the more documents which one has to face, the more challenging the problem of finding what they need is due to information overload. 
In these scenarios, information-seeking processes are more beyond the searching of facts, aiming to extend one’s knowledge about a topic, gain new insights and discover conceptual boundaries~\cite{Exploratory1121979}.
Therefore, new tools and techniques are required to facilitate understanding of big data.

To efficiently cope with a large amount of information and generate significant value to users, one must identify the data and merge it as a whole entity, specifically when the question comprises a distributed and complicated topic.
This process commonly includes three activities, including (i) discovery, to explore new information; (ii) filtering, to summarise new information; and (iii) adapting, to make information accessible to new users~\cite{pollar2004surviving}.
Therefore, summaries are helpful when huge amounts of information need processing.

Data summaries can help gather related information and collect it into a shorter format that enables answering complicated questions, gaining new insight and discovering conceptual boundaries.
A large body of research exists to address this problem, specifically regarding summaries that need to be generated in textual form, known as ‘text summarisation’. 
Automatic document summarisation is a long-studied area covering different perspectives to articulate the effects and needs of data reduction for analysis, commercialisation, management and personalising purposes.

The goal of automatic text summarisation is to identify and highlight the critical aspects of one or multiple input document(s) within a defined size limit~\cite{gupta2010survey}.
A good summary conveys the key ideas efficiently, allowing users to quickly gain an overview of the text without expending much effort.
Despite the body of research, it is still difficult to produce summaries that are comparable to human-generated ones~\cite{gupta2010survey}.
The massive collection of documents (volume), the speed of generated documents (velocity), and the unstructured format (variety) in which documents appear make summarisation challenging.
Various text summarization models have been studied in the natural language processing community, including extractive and abstractive approaches, single-document or multi-document summarisation (MDS), and query-focused or update summarisation. 
Existing MDS approaches produce a uniform summary for all users without considering individual interests.
Optimising a system to produce one ‘best’ summary that satisfies all variants of users is also highly impractical.
Therefore, a more significant challenge is the subjectivity aspect of summarization in selecting the content in summaries for different users.
Besides, current approaches optimise their system based on gold-standard summaries generated by human experts, which are expensive and time consuming.

This dissertation proposes novel approaches towards intelligent, personalised and interactive human-in-the-loop summarisation.
We first introduce concepts central to the work described in this dissertation, and identify significant research issues in personalising document summarisation and information-seeking processes. 
This is followed by a description of the critical research issues tackled in the thesis.
Next, we summarise our contributions to the area and outline the thesis structure.


\section{Key Research Issue}
This thesis focuses on three main challenges proposing novel summarisation techniques to alleviate information overload.
We intend to facilitate the analysis of documents to support personalised information extraction.
Therefore, we separate the research issues into three areas covering (i) feature engineering in document summarisation, (ii) interactive summarisation and (iii) personalised summarisation.

\subsection{Feature Engineering in Document Summarisation}
The essential part of any text-related problem is feature engineering, (i.e., the process of creating features from raw text data).
Feature engineering in machine learning is more than selecting the appropriate features and transforming them. 
It prepares a dataset to be compatible with an algorithm, improves a model’s performance, and reveals any hidden information, highlighting what is important in summarisation.
Text summarisation is an old challenge that dates back to the 1950s when such features as word and phrase frequency were used to select essential sentences. 
However, these factors are still required for feature engineering in summarisation tasks, due to its applicability.

\subsection{Traditional Static and Inflexible Summaries}
\label{}
In general form, summarisation takes a topic-related document set as input and generates a summary that bears the most crucial aspect.
Research on MDS, in general, ignores the usefulness of the approach for users. Instead, the literature mostly focuses on the accuracy of their produced summaries, resulting in short (3–6 sentences), single, inflexible and flat summaries for all users.
Therefore, these approaches are incapable of producing more extended summaries since all the details are omitted, even if users are interested in obtaining more information.
Besides, the produced summaries are unstructured and, therefore, are improperly equipped for allowing further analysis.
Unfortunately, a single summary is unlikely to serve all users in a large population. 
Therefore, there is an urgent demand to make summaries that can be changed upon a user’s request.

\subsection{Traditional Generic Summarisation Approaches}
Existing summarisation approaches produce a general summary in the form of a few selected sentences to serve all the users’ needs.
In contrast to a generic summary that is unique for all users, there is a lack of user-centric summarisation approaches in which a user can specify their desired content in making summaries. 
Several reports~\cite{berkovsky2008aspect,park2010automatic} demonstrate that users prefer personalised summaries that accurately reflect their interests\textemdash hence the demand. 
However, the system should have background knowledge about the user to achieve this, making tailored summarisation a challenging task.
Challenges include acquiring relevant information about a user, aggregating and integrating the information into a user model, and using the provided information to make a personalised summary.

\subsection{Need for Reference Summaries}
State-of-the-art summarisation approaches train their system towards one single summary, referred to as reference or gold-standard summaries made by a human~\cite{gupta2010survey}.
A report by Lin~\cite{lin2004rouge} shows that three thousand hours of human effort were needed for summaries' evaluation for the Document Understanding Conferences (DUC)\textemdash a subjective, costly and time-consuming task.

\section{Contribution Overview}
Our goal is to facilitate the creation of personalised and comprehensive text summarisation. 
The research discussed in this dissertation aims to answer the following high-level research questions:
\begin{itemize}
    \item How useful are the current state-of-the-art methods in automatically creating and detecting features in the context of document summarisation? How can we distinguish the efficiency of different features?
    \item What structures are required to help users seek their desired information? How can we use individual feedback such that summarisation approaches adapt to the users’ needs?
    \item Can we simulate user behaviour to predict their desired summary or information-seeking path? Can we provide predictability for unseen situations in terms of application?
    \item How can we eliminate the need for reference summaries to reduce the cost of summarisation? How can the task of persoanlised summarization be better regulated, formulated, and comparable to contribute more to the research field?
\end{itemize}
To answer these questions, we propose novel computational methods and extensive experiments for the creation of intelligent and personalised summaries.
In detail, this thesis makes the following contributions to tackle this problem, categorised as enabling:
\begin{itemize}
    \item automatic intelligent feature engineering
    \item flexible and interactive summarisation
    \item intelligent and personalised summarisation.
\end{itemize}

\subsection{Enabling Automatic Intelligent Feature Engineering}
To enable automatic intelligent feature engineering in document summarisation, we propose two solutions, which are also fundamental to other contributions made by this thesis.

\subsubsection{Contribution 1: Dynamic Feature Space Mapping Through a Novel Traditional Summarisation Approach}
We present a novel intelligent approach for summarisation, which benefits from using supervised and unsupervised algorithms simultaneously and in an interpretable manner.
We combine the clustering and classification algorithm into a single objective function.
Clustering aims to discover the underlying structure of data, and then feeds the processed information to the classification stage through a single objective function.
The goal is to improve the performance of the summarisation algorithm.
Besides, features are dynamically weighted by optimising the process in each cluster.
The proposed approach is capable of measuring the role of features through the summarisation process by making various feature spaces.

\subsection{Enabling Flexible and Interactive Summarisation}
Compared to gold-standard summaries, previous approaches mainly optimise their system to enhance summaries’ accuracy rather than users’ need, resulting in short and static summaries.
We present an interactive narrative framework based on the feature extracted from users’ engagement to tackle these issues.
Therefore, we propose to hierarchically structure the summary output to improve navigation and provide users with more information upon request. 
Based on the proposed structure, we present two intelligent and interactive summarisation approaches: semi-structured narrative summaries (SNARS) and fully structured narrative summaries (FNARS).
The proposed approaches aim to engage users in the summarisation process to guarantee interactive speed even for extensive text collections, and to eliminate the need for reference summaries.

\subsubsection{Contribution 2: An Intelligent Semi-structured Narrative Summaries (SNARS) for Interactive Summarisation}
We propose a new task for MDS, called narrative summaries (NARS); this technique gathers the related information and collects it into a coherent hierarchy.
We called this approach ‘narrative summaries’ because it provides information in a logical order, ranked from the most indicative sentences to more informative sentences, to prevent users from being cognitively overwhelmed with a complete summary at once.
This is important because a user might be interested in different aspects of a topic based on their background knowledge, situation and context, as well as cognitive bias.
Therefore, we enable users to explore their desired information with minimal reading required.
Since sentences are the representative unit in this structure, the proposed model is considered an extractive and semi-unstructured approach.

\subsubsection{Contribution 3: An Intelligent Fully-structured Narrative Summaries (FNARS) for Interactive Summarisation}
Studies have shown that when people are exposed to many documents, they rarely make a fully formulated summary. 
Instead, they attempt to extract concepts and the relationships between them~\cite{pirolli2005sensemaking,chin2009exploring,yimam2016new}.
Therefore, access to structured data is increasingly critical in every domain~\cite{kirkpatrick2015putting,van2017integrated,jackson2016desiderata}.
The proposed model is presented as a hierarchical concept map, as this structured presentation style is suitable for summarisation.
A concept map is a labelled graph.
Nodes in a concept map denote the concepts and edges are the relations between them ~\cite{falke2019automatic}.
Therefore, a summary in concept map form provides a concise overview of the contents of document collections, reveals interesting connections, can provide better navigational structure for greater exploration, and facilitates use of the summaries for further analysis.
The representation unit is a sentence in SNARS and a concept in FNARS. Therefore, FNARS provides a more concise overview of information, while SNARS provides more detail.
Conversely, FNARS is a fully structured model and can be used for further analysis.

\subsection{Enabling Intelligent and Personalised Summarisation}
Traditional summarisation approaches produce a generic summary that fits all users.
Therefore, the subjectivity aspect of summarisation (i.e., what is deemed valuable for different users) is ignored, making these approaches impractical in real-world scenarios.
Crafting a user-specific summary for an input document cluster is a challenging task.
In a personalised approach, the system needs to know about a user’s background knowledge or interests.
When we do not have access to prior knowledge, including profile or background knowledge, the system requires user interaction to procure feedback for modelling individual interests.
Therefore, this thesis provides human-in-the-loop approaches to generate personalised summaries that better understand the users’ needs.
This eliminates the need for reference summaries, which is a challenging issue for summarisation tasks.

\subsubsection{Contribution 4: Adaptive Summarisation: A Personalised and Interactive Concept-based Summarisation Approach}
Producing a user-desired summary is a challenging task.
To tackle this problem, we propose a novel optimisation algorithm that directly reflects users’ interests in making extractive summaries.
The proposed approach, called ‘adaptive summaries’, learns from the information users provide to the system gradually through interaction.
Users are allowed to provide their feedback in an iterative loop, and can choose important concepts while defining their degree of importance.
Moreover, they can define their confidence level in their choices and can even select which concepts not to include in the output.
The proposed approach learns to select sentences that maximise the summary score according to user feedback.
By doing this, we allow even novice users to interactively explore, manipulate and analyse unstructured text document collections to integrate their own notion of importance.
The proposed approach can guarantee interactive speed to keep the user engaged in the process.

\subsubsection{Contribution 5: Novel Personalised Summarisation Approach to Predict Users’ Desired \textit{Unstructured} Summaries Based on Feedback}
The proposed approach considers the summarisation problem as a recommender system, where the goal is to suggest a personalised summary to a user based on their preferences. 
Therefore, we keep humans in the loop by providing feedback through interaction.
We also propose a preference-based interactive summarisation model to extracts users’ interests to learn and generate user-adapted results.
We present the content in a condensed fashion using summarisation while interacting with the system.
The proposed method, ‘SumRecom’, learns to predict users’ preferences to generate summaries utilising their feedback by creating a behavioural model that relies on reinforcement learning (RL).
We also use an inverse RL (IRL) algorithm to automatically evaluate a summary based on a domain expert’s previous history evaluation.
The representation unit is a sentence, meaning the model follows an extractive approach~\cite{ghodratnama2020extractive}.

\subsubsection{Contribution 6: Personalised Summarisation Approach to Predict Users’ Desired \textit{Structured} Summaries Based on the Proposed Hierarchical Structure}
Predicting users’ desired \textit{structured} summary is another challenge addressed in this dissertation.
We examine the automatic production of personalised and structured summaries to dynamically maintain a federated summary view incrementally, resulting in a unified framework for an intelligent summary generation.
We first propose a hierarchical, personalised, concept-based summarisation approach, which sums up a collection of documents in a concise hierarchical concept map based on users’ feedback.
Instead of providing a short and static summary, the approach engages users by querying their preferences and utilising an RL algorithm to learn how to generate user-adapted content through a hierarchy.
The proposed model improves the deficiency of traditional approaches in various aspects such as subjectivity aspect and finding interesting patterns and relationships in different collection parts interactively.

\subsection{Dissertation Organisation}
This dissertation is organised as follows.
We first discuss the fundamental background and a discussion of the current state-of-the-art approaches in Ch.~\ref{Ch_2}.
We explain what document summarisation is in more depth, and offer a new perspective on the related work for describing different categories towards personalising summaries.
The summarization task is defined formally, and the challenges are discussed in the following.

Ch.~\ref{ch_3} discusses in depth the experimental set-up,  evaluation metrics, and the required data to train and evaluate the proposed summarisation models.
Ch.~\ref{ch_4} then focuses on the central task of document summarisation and the role of feature engineering.
We propose a novel approach to dynamic feature space mapping to tackle the problem.
We then evaluate the proposed approaches experimentally to assess the applicability of modelling for future tasks.

Ch.~\ref{ch_5} explores the problem of automating intelligent and interactive summarisation.
We propose a hierarchical structure for this purpose. 
Based on the proposed structure, we provide two solutions for interactive summarisation\textemdash SNARS and FNARS\textemdash followed by human and automatic evaluation of the proposed approaches.

Ch.~\ref{ch_6} discusses the problem of automating intelligent and personalised summarisation.
We focus on alternative models for personalising summarisation and predicting desired summaries for specific users. 
Different solutions for a personalised and interactive concept-based summarisation approach are posited.
Each predict users’ desired extractive and unstructured and structured summaries based on their feedback, as well as the personalised information-seeking path.
Then, a comprehensive evaluation of each approach is given.

Ch.~\ref{ch_7} develops different applications of the technology discussed in this thesis.
We discuss how our model facilitates the representation and analysis of documents in various domains and address three specific applications, including:
\begin{itemize}
    \item the use of summarisation in detecting anomalies in network traffic data
    \item application of summarisation in healthcare analytic problems
    \item using summarisation to narrate business process data.
\end{itemize}

We also propose a novel approach called ‘Summary2vec’, which presents each summary by a fixed-length vector using a novel architecture.
The goal is to create a numeric representation of summaries as an individual related unit of meaning conveying one or more aspects of information space. 
Summary2vec is remedial, in that it helps to design automatic services for various analytic purposes that require information-seeking activities.

Finally, Ch.~\ref{ch_8} summarises the findings and outlines promising directions for future research.
We close the chapter by highlighting the remaining challenges to encourage and guide possible future directions on this research topic.

%% file: ch_2/ch_background.tex
\chapter{Background and Related Work}
\label{Ch_2}
This dissertation investigates the problem of intelligent, personalised and human-in-the-loop document summarisation.
This chapter discusses the fundamental concepts in the field and state-of-the-art document summarisation techniques.
We start with an overview of document summarisation and the definition, followed by a comparison with document compression (Sec.~\ref{overview_ch2}).
We conclude that there is no ideal summarisation approach, as summarisation is defined according to the needs of the application and individual users.

Sec.~\ref{TraditionalCategories} categorises state-of-the-art approaches based on the common categorisation style, including input type, purpose type and output type.
Each category and the following subcategories are also explained.
We then propose a new categorisation schema including traditional summarisation, structured summarisation, and interactive and personalised summarisation.
Sec.~\ref{featureEngineer} next discusses the need for feature engineering in document summarisation and explores the various existing features (with their respective definition), including term-level, sentence-level, paragraph-level and corpus-level features.

Sec.~\ref{TraditionalApproach} presents various works in traditional automatic summarisation, especially more recent ones.
We also explain the need for structured (Sec.~\ref{structured}), interactive, and personalised (Sec.~\ref{InteractiveApproach}) summarisation approaches.
The chapter is concluded in Sec.~\ref{ch2_summary}.
We present both the challenges and limitations that prevent progress in the field, and further highlight the remaining problems and gaps in personalised document summarisation.
These challenges are interesting for researchers as a guide for future research.

\section{Document Summarisation: Overview}
\label{overview_ch2}
Producing a summary is complicated even for a domain expert, yet it can be even more difficult for machines.
The machine should have the ability for learning NLP and producing summaries and background knowledge suitable for humans.
This section explains different summarisation definitions and compares the summarisation task with the compression task.

\subsection{Definition}
\label{definition}
Summarisation creates the best representation of the original data, enabling efficient storage, quick browsing, and retrieval of a large collection of data without loss~\cite{borko1975abstracting}.
However, there is no unique definition for summarisation, meaning it can be understood based on application or user. 
For instance, summarisation can be defined as the process of reducing data size or finding the important part(s) of data while eliminating redundant or non-relevant data.

The most general definition for summarisation is the automatic mechanism of generating brief and condensed representations of the content~\cite{luhn1957statistical}.
A summary can also be defined as concise, informational, and grammatically correct text without redundancy~\cite{aries2019automatic}.
A summary is also defined as a shorter version of a document generated by a machine to draw the most significant information in a more concise format without human assistance~\cite{das2007survey}.

Radev et al.~\cite{radev2002introduction} provided a more recent definition, framing summarisation as ‘a text that is produced from one or more texts, that conveys the important information in the original text(s) and usually significantly less than that’.
According to this definition, three important issues should be considered:
\begin{itemize}
    \item Summaries are produced from single or multiple documents.
    \item Summaries should preserve the important parts of the original text.
    \item Summaries have to reduce the original text by at least 50\%.
\end{itemize}

Defining what is important in this definition is a challenging task.
For example, in query-based summarisation, a good summary should cover topics related to the query. 
Conversely, it is more challenging to decide which aspects are interesting to readers when it comes to generic summarisation systems\textemdash finding a balance between the main and other related topics~\cite{zhang2012mutual,song2011fuzzy}.
Size is also an essential aspect of summarisation.
However, size is the main aspect in developing compression-based approaches.
Sec.~\ref{Compression} compares summarisation and compression approaches for clarification.

\subsection{Document Summarisation v. Document Compression}
\label{Compression}
While text summarisation aims to discover relevant information from multiple documents and prepare this in a concise readable format for users, document compression seeks to reduce the amount of data required to represent documents.
Therefore, document compression employs filtering approaches to tackle information overload.
The document compression concept derives from the idea of compressing data using encoding techniques used in information theory~\cite{shannon1948mathematical}.
Text compression techniques condense sentences while keeping the important information.
Text compression has a wide variety of practical applications such as compressing microblogs and generating headlines for news articles.
Text compression approaches are classified into two groups, including deletion-based and abstractive procedures.

\subsubsection{Deletion-based Text Compression}
Deletion-based text compression approaches work based on the idea of \textit{reducing without significant loss }~\cite{jing2000sentence}.
Therefore, the goal is to remove as many extraneous words from a (set of) document(s) without diminishing the text’s main content or sentence transformations~\cite{marsi2009limits}.
‘Data-intensive processing’ and making data ‘lean’ are two main subcategories of deletion-based text compression.

Early approaches belong to the lean category since they follow an unsupervised paradigm, such as an integer linear programming-based approach~\cite{clarke2007modelling}.
Consequently, they do not require training data but rather employ a language model to extract the most compressed sentences instead of using training sentence–compression pairs.
Conversely, the data-intensive approaches are supervised and require sentence-compression pairs for training.
Other proposed modelling approaches include the noisy-channel model~\cite{knight2000statistics,radev2002introduction,knight2002summarization}, variational auto encoders~\cite{miao2016language} and Seq2Seq models~\cite{filippova2015sentence}.
A recent approach proposed by Zhaoet al.~\cite{zhao2018language} is based on a new language-model-based evaluator.
They use trial-and-error deletion operations and the RL algorithm to get the most desirable target compression.

\subsubsection{Abstractive Text Compression}
Abstractive models produce more compressed texts by inserting, reordering or deleting operations\textemdash similar to abstractive text summarisation techniques~\cite{knight2002summarization}.
A recent abstractive compression model is a tree-to-tree transduction model using synchronous tree-adjoining grammars~\cite{shieber1991synchronous} to obtain all different formats for rewriting a sentence
Other approaches include using attentive long short-term memory (LSTM) models used for caption compression~\cite{wubben2016abstractive}, and a Seq2Seq model decoder~\cite{yu2018operation}.

\section{Document Summarisation Categories}
\label{TraditionalCategories}
\begin{figure*}
\centering
\centerline{\includegraphics[width=1.1\textwidth]{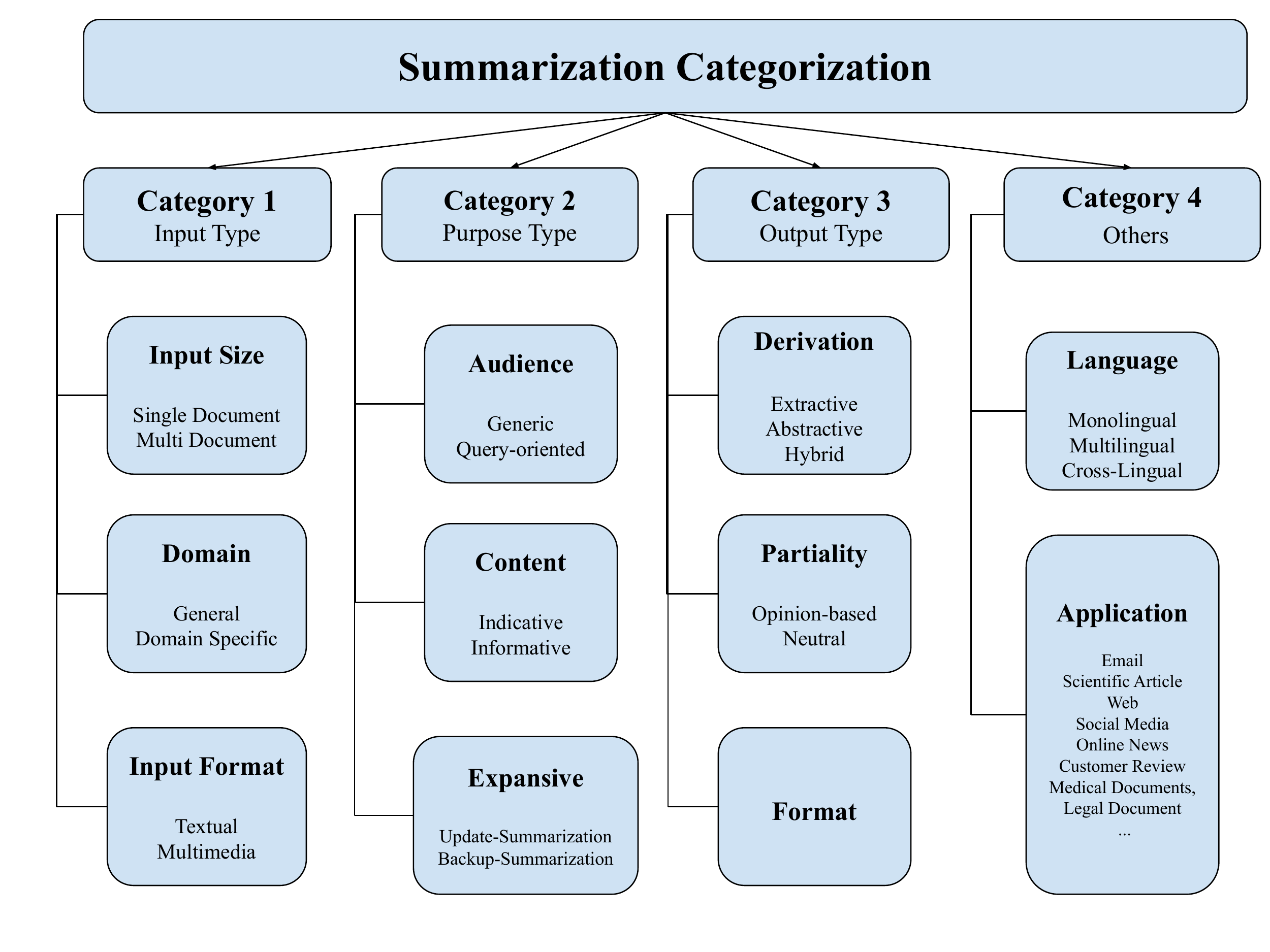}}
\caption{Document summarisation categorisation.}
\label{Tex-Summarization-category}
\end{figure*}

Document summarisation approaches are categorised based on the input, purpose and output type. A graphical structure of these categories is depicted in Fig.~\ref{Tex-Summarization-category}.

\subsection{Categorising Based on Input Sources’ Properties}
We examined the input documents using three criteria: (i) input size (how many documents a system can have as input), (ii) domain specificity (domains that the model can handle), and (iii) input format (the structure of documents).

\subsubsection{Single-document Summarisers v. Multi-document Summarisers}
Early works on summarisation focused on a single document.
However, recently, due to the fast development of documents, MDS has gained more attention~\cite{gupta2010survey}.
Single-document summarisation approaches process just one input document, and the first work in this category returns to 50s~\cite{luhn1958automatic,baxendale1958machine,edmundson1969new}.
Conversely, multi-document summarisers gather many documents on the same topic as input, enabling diversification of information sources and redundancy simultaneously~\cite{mckeown1995generating}.

\subsubsection{Domain Specificity: General Input v. Domain-specific Input}
The input documents can be general or related to the same domain, such as legal documents or biomedical documents.
However, it is more suitable to develop a summarisation system specific to a domain to reduce term ambiguity, use grammar and formatting schemes, and facilitate the use of domain knowledge by enhancing relevancy detection.
An example is a medical text summarisation where the authors use domain-specific cue phrases and other features to find relevant sentences~\cite{sarkar2009using}.
Another example is LetSum, a summarisation system for legal~\cite{farzindar2004legal} and biomedical texts~\cite{reeve2007use}.

\subsubsection{Input Format: From Textual to Multimedia Documents}
Input documents are in varied forms based on their structures, mediums and scales.
The structure relates to the document organisation.
For example, the structure of a scientific article typically includes an introduction, related works, methods, experiments and conclusions.
Focusing on the structure is the basis of some approaches to generate summaries, as this increases summarisation performance~\cite{farzindar2004legal}.

Scale is another aspect that affects performance.
For example, in many known summarisation systems, term frequency is used for measuring relevancy.
Consequently, the input document should be large enough to detect essential concepts. 
Conversely, common techniques do not work for summarising low-scale documents such as tweets and microblogs,~\cite{sharifi2010summarizing}.
Instead, trending phrases or a phrase specified by users, such as a hashtag, can be selected as a topic~\cite{duan2012twitter}.

The medium is another critical aspect that carries information.
While most summarisation systems focused on textual support, there exists some work on summarising images, audio and videos ~\cite{kim2014joint}. 
Recently, there has been increased interest in multimedia summarisation.
Examples include interactive football summarisation~\cite{moon2009interactive}, summarising important events in a football video~\cite{zawbaa2012event}, video summarisation using web images~\cite{khosla2013large}, video summarisation using a given category~\cite{Potapov2014}, and video summarisation by learning~\cite{Gygli_2015_CVPR}.

\subsection{Categorising Based on Summarisation Purpose}
According to the summarisation goal, summarisation approaches are categorised based on the audience, content and expansiveness.

\subsubsection{Audience: Generic v. Query-oriented Summarisation Approaches}
The concept of importance is different for different users.
For example, a user may need to focus on specific aspects of a document rather than the input document’s main idea.
In these scenarios, the interest can be defined by a query.
In a query-oriented summarisation approach, a good summary is judged according to a user’s query.
One common technique in this category is to adapt existing summarisation approaches to answer a query.
Topic signature words or graph-based approaches~\cite{qazvinian2010citation} and submodular approaches~\cite{lin2011class} are examples of common approaches in this category.
Other approaches also exist, which are explicitly designed for answering queries~\cite{abdi2018qmos}. 
In contrast to a query-based approach, a generic summarisation method tries to preserve important information presented by the author from an input document~\cite{vanetik2015,thomas2015,Vicente2015}.

\subsubsection{Content: Indicative v. Informative Summarisation Approaches}
‘Informative summaries’ include the primary information of the original documents and help users find their interests by extracting the main idea(s).
Most existing summarisation approaches are informative ones.
Conversely, ‘indicative summaries’ only include a global representation of the original document(s) and can be helpful to decide whether to refer to the original source.

\subsubsection{Expansiveness: Update Summarisation v. Background Summarisation Approaches}
A generated summary may focus on the original document’s background or compare it to past documents, referred to as the ‘expansiveness’ metric. 
A ‘background summarisation’ algorithm produces summaries based on the input content without eliminating information from prior related documents~\cite{mani2001summarization}.
Conversely, ‘update summarisation’ should convey information beyond what is already known.
The goal is to generate summaries from recent documents that do not include prior information, known as ‘novelty’.
Extracting novelty in addition to salience can be modelled using latent Dirichlet allocation or through incremental hierarchical clustering~\cite{delort2012dualsum}.

\subsection{Categorising Based on Output Summary Properties}
According to the output summary, summarisation algorithms are classified based on three measures: (i) the derivation process (the process applied to generate a summary from the primary document), (ii) partiality (how a summary handles the original document’s opinions), and (iii) the summary’s format.

\subsubsection{Derivation: Extractive Summarisation v. Abstractive Summarisation}
The derivation measure refers to the process of obtaining a summary\textemdash that is, extracting important parts or producing a new summary.
Abstractive summaries are generated by interpreting the main concepts of a document, and then stating those contents using clear and natural language~\cite{erkan2004lexrank,barros2019natsum}.
Abstraction techniques are a substitute for the original documents rather than a part of them.
Therefore, abstractive approaches require deep NLP such as semantic representation and inference.
However, these are challenging to produce due to the current limitation in linguistic techniques.

Conversely, extractive methods are more popular because of their comparative simplicity.
This approach usually contains three steps, involving (i) representation of the original text document, (ii) sentence scoring, and (iii) selecting high-scoring sentences in the summary.
Extractive summaries are generated by selecting units as the representative of the original documents, usually measured in sentences, whose grammatical structure is easy to maintain.
These sentences are then concatenated into a shorter text to produce a meaningful and coherent summary~\cite{mehta2018effective}.

\subsubsection{Partiality: Opinion-based Summarisation v. Neutral Summarisation}
Produced summarisation can be neutral or opinion based.
Neutral summarisation algorithms produce summaries that reflect the input documents’ content without judgement or evaluation, even if they are judgemental in nature.
Most existing summarisation works belong to this category~\cite{Vicente2015,thomas2015}.
In contrast, opinion-based summarisation algorithms include automatic judgements either implicitly or explicitly.
An explicit judgement includes some opinion statements in the summary, while the implicit one uses bias to add and/or omit material.
With the growth of interest in users’ opinions, this kind of summarisation is more popular.
For example, one approach summarises customer opinions through Twitter by extracting the different product features and the conversation messages’ polarity~\cite{othman2016customer} .

\subsubsection{Format: Output Format}
Produced summaries can feature in various formats, including structured in a concept map format~\cite{falke2019automatic} or unstructured in the form of some sentences using a different template~\cite{heu2015fodosu}.
Summaries may also focus on users’ preferences or goals. For example, the OntoSum system~\cite{bontcheva2005generating} uses device profiles such as mobile phones and web browsers to adjust the summary formatting and length.

\subsection{Language-based and Application-based Categorisation}
This section discusses different categories of summarisation algorithms based on language and application of document summarisation.

\subsubsection{Language Model}
Based on the input documents’ language, three different categories exist for summarisation: monolingual, multilingual or cross-lingual. 
In a monolingual summarisation algorithm the source and the target language are the same and specific.
Multilingual summarisation approaches accept documents in different languages and produce the output in the same language~\cite{jevzek2008automatic}.
For example, SUMMARIST~\cite{hovy1999automated} is a multilingual summariser that is available in English, Japanese, Spanish, Arabic, Indonesian and Korean and FarsiSum~\cite{hassel2004farsisum} (which is a monolingual text summarisation system for Persian).
In a cross-lingual summarisation system, the source text is written in one language and the user can choose the language of the output summary~\cite{gambhir2017recent}.

\subsubsection{Application}
Document summarisation has many real-world applications, which have been studied at length.
Examples include summarisation for emails, online news and social media; customer reviews; scientific articles; and both medical and legal document summarisation.
The following describes email and scientific paper summarisation in more detail.

Emails have some distinct characteristics such as the interactive dialogue nature, as in verbal communications.
Nenkova et al.~\cite{nenkova2004facilitating} offer initial progress in this field by generating a summary of thread discussions.
Newman et al.~\cite{newman2003summarizing} further proposed a system to summarise a full inbox rather than a single thread.
They clustered messages into topical groups and then obtaining summaries for individual clusters.

In scientific paper summarisation, there is a substantial amount of research to be used to identify essential sentences in an original article.
One approach is to use a language model that indicates a probability to words mentioned in the citation context sentences, following by scoring the sentence importance in an original paper using the Kullback–Leibler divergence method~\cite{mei2008generating}.

\subsection{Proposed Categorisation Schema}
\label{proposedcategories}
The problem of automatic document summarisation is an old challenge, and many approaches exist for various purposes to solve one key drawback of previous methods. 
The following lists different metrics to evaluate a good summary from various perspectives~\cite{falke2019automatic}.
Based on the goal and the application, a summarisation approach should:
\begin{itemize}
    \item contain an overview of the chosen documents
    \item remove redundancy
    \item present the various relations in documents
    \item generate summaries with great detail while also covering the main topics
    \item be user friendly
    \item be multidimensional and cover different aspects of documents
    \item produce personalised summaries based on the user’s goal(s).
\end{itemize}

\begin{figure*}[t]
\centering
\centerline{\includegraphics[width=\textwidth]{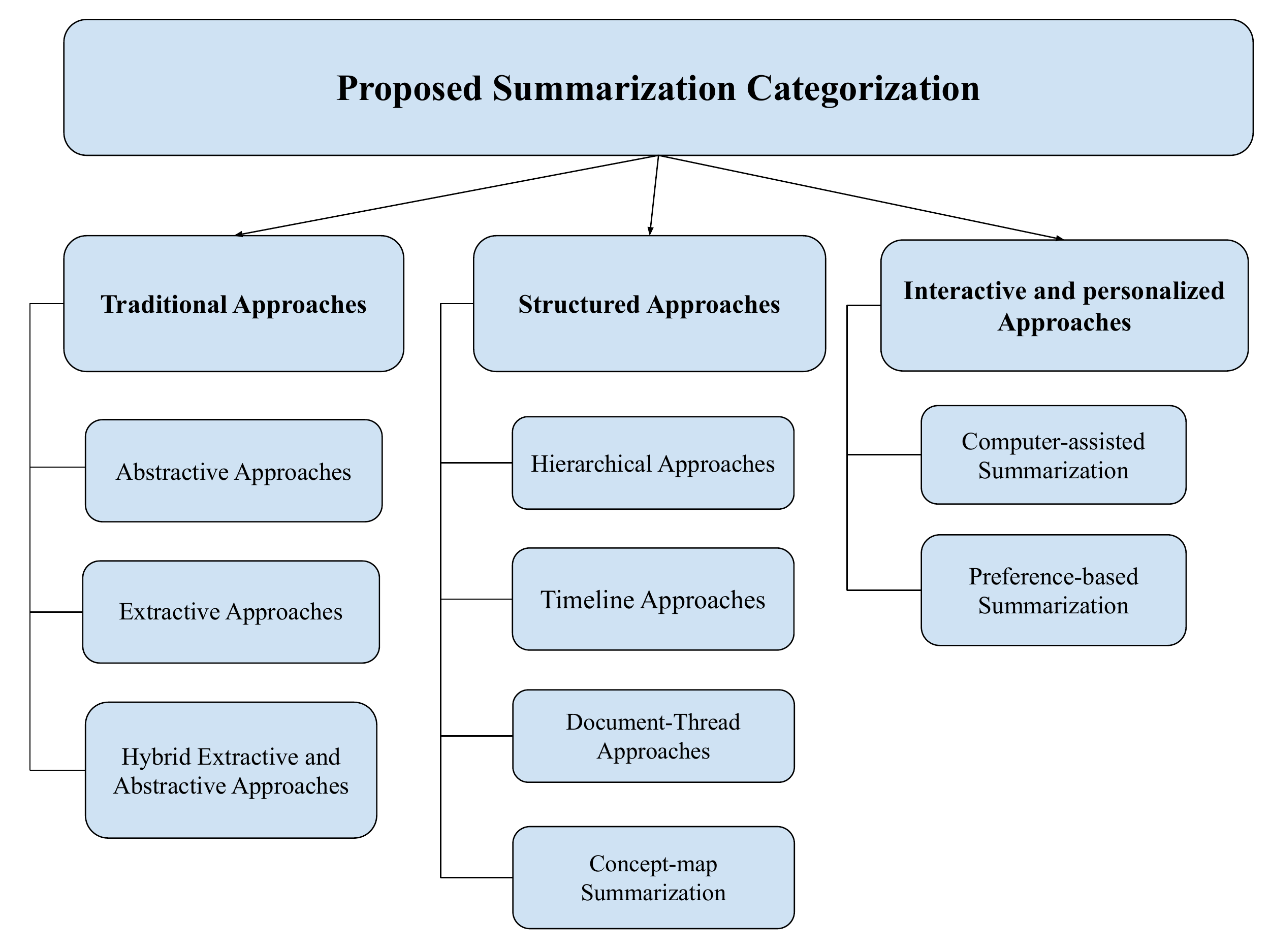}}
\caption{Proposed categorisation schema for document summarisation.}
\label{fig:Proposedcategory}
\end{figure*}

Based on these criteria, the existing gaps and the requirements of good document summarisation, we categorised state-of-the-art approaches as (i) traditional, (ii) structured, and (iii) interactive and personalised.
The proposed hierarchy is depicted in Fig.~\ref{fig:Proposedcategory}.
Before explaining each category and the leading approach in each category, we first describe feature engineering in document summarisation, required to explain state-of-the-art techniques.

\section{Feature Engineering in Document Summarisation}
\label{featureEngineer}
Abstractive approaches require deep NLP such as semantic representation and inference.
However, they are challenging to produce and require maturation.
The goal of automatic extractive summarisation is to rank the important sentences in a document set (i.e., sentences containing maximum information about the document’s topics).
Therefore, sentences should be represented based on certain \textit{features}.
This section explains state-of-the-art features and the proposed approaches to learn optimal feature sets.

\subsection{State-of-the-art Features}
Luhn~\cite{luhn1958automatic} was the first to address the automatic document summarisation problem in the 1950s using the keyword frequency feature.
Other features were later proposed, including cue words, title words and sentence location~\cite{edmundson1969new}.
Church et al.~\cite{church1999inverse} proposed other new features such as sentence length cut-off, fixed phrases, paragraph features, thematic word features and upper-case word features.
Later, Hovy and Lin~\cite{lin1997identifying} verified that the position method is not applicable and efficient for all domains.
Therefore, other features such as term frequency–inverse document frequency (TF–IDF),cue words, sentence location, and longest common subsequences (LCSs) were proposed for sentence scoring~\cite{radev2001experiments}.

Many other features have been also proposed over the years to improve the performance of automatic document summarisation.
We categorised existing features summarisation in four groups, covering term-level features, sentence-level features, paragraph-level features and corpus-level features~\cite{meena2015optimal}.
Table ~\ref{tab:termfeatures},~\ref{tab:sentencefeatures},~\ref{tab:parafeatures} and ~\ref{tab:Corpusfeature} present some important features in each group, with brief descriptions.

\begin{table}[b!]
    \centering
    \caption{Term-level Features}
    \vspace{1mm}
    \label{tab:termfeatures}
    \begin{tabular}{|l|l|l|l|}
    \hline
    \multicolumn{2}{|l|}{\textbf{Feature}}& \multicolumn{2}{|l|}{\textbf{Description}} \\ 
    \hline
    \multicolumn{2}{|l|}{Term Frequency}& \multicolumn{2}{|l|}{Frequent words mentioned in the document.} \\ 
    \hline
    \multicolumn{2}{|l|}{TF-IDF}& \multicolumn{2}{|l|}{Frequent words considering other terms.} \\ 
    \hline
    \multicolumn{2}{|l|}{Cue Words}& \multicolumn{2}{|l|}{\begin{tabular}{@{}c@{}}Sentences includes cue words.\end{tabular}}\\
    \hline
    \multicolumn{2}{|l|}{Title Similarity}& \multicolumn{2}{|l|}{Sentences containing words from the title.} \\ 
    \hline
    \multicolumn{2}{|l|}{Uppercase word}& \multicolumn{2}{|l|}{Sentences include upper-case words.} \\ 
    \hline
    \multicolumn{2}{|l|}{Positive Keyword}& \multicolumn{2}{|l|}{\begin{tabular}{@{}c@{}} Frequent keywords occurring in the summary.\end{tabular}}\\
    \hline
    \multicolumn{2}{|l|}{Negative Keyword}& \multicolumn{2}{|l|}{\begin{tabular}{@{}c@{}} Keywords not frequently occurring in the summary.\end{tabular}}\\
    \hline
    \multicolumn{2}{|l|}{Residual IDF}& \multicolumn{2}{|l|}{\begin{tabular}{@{}c@{}} The residual IDF of a word.\end{tabular}}\\
    \hline
    \multicolumn{2}{|l|}{Gain}& \multicolumn{2}{|l|}{\begin{tabular}{@{}l@{}} Features based on hypothesising which moderately frequent\\ words are most important in a document.\end{tabular}}\\
    \hline
    \multicolumn{2}{|l|}{Term Co-occurrence}& \multicolumn{2}{|l|}{\begin{tabular}{@{}c@{}} Clusters of important words are identified and weighted.\end{tabular}}\\
    \hline
    \multicolumn{2}{|l|}{Query Score}& \multicolumn{2}{|l|}{\begin{tabular}{@{}c@{}} Sentences are scored according to the number of query terms.\end{tabular}}\\
    \hline
    \multicolumn{2}{|l|}{Synonyms}& \multicolumn{2}{|l|}{\begin{tabular}{@{}c@{}} Synonyms are matched using WordNet or other tools.\end{tabular}}\\
    \hline
    \multicolumn{2}{|l|}{Significant Word}& \multicolumn{2}{|l|}{\begin{tabular}{@{}c@{}} Relative significance of words.\end{tabular}}\\
    \hline
    \multicolumn{2}{|l|}{Title Similarity}& \multicolumn{2}{|l|}{\begin{tabular}{@{}l@{}} Squares the number of common terms between a document’s \\ title and each sentence.\end{tabular}}\\
    \hline
    \end{tabular}
\end{table}

\begin{table}
    \centering
    \caption{Sentence-level Features}
    \vspace{1mm}
    \label{tab:sentencefeatures}
    \begin{tabular}{|l|l|l|l|}
    \hline
    \multicolumn{2}{|l|}{\textbf{Feature}}& \multicolumn{2}{|l|}{\textbf{Description}} \\ 
    \hline
    \multicolumn{2}{|l|}{Sentence location}& \multicolumn{2}{|l|}{\begin{tabular}{@{}c@{}} Position of sentences determine the weights.\end{tabular}}\\
    \hline
    \multicolumn{2}{|l|}{Semantic structure}& \multicolumn{2}{|l|}{\begin{tabular}{@{}l@{}} Using a graph structure where node are sentences, and related sentences \\ are recognised by edges between them.
    \end{tabular}}\\
    \hline
    \multicolumn{2}{|l|}{Length Cut Off}& \multicolumn{2}{|l|}{\begin{tabular}{@{}c@{}}Too short or too long sentences are eliminated.\end{tabular}}\\
    \hline
    \multicolumn{2}{|l|}{Fixed Phrase}& \multicolumn{2}{|l|}{\begin{tabular}{@{}l@{}} Sentences containing some fixed phrases, known as indicator\\ phrases, are given priority.\end{tabular}}\\
    \hline
    \multicolumn{2}{|l|}{Concept Signature}& \multicolumn{2}{|l|}{\begin{tabular}{@{}c@{}} Topic words and the associated pairs are selected (co-occurrence feature).\end{tabular}}\\
    \hline
    \multicolumn{2}{|l|}{Concept Count}& \multicolumn{2}{|l|}{\begin{tabular}{@{}c@{}} Counts the concepts' occurrence instead of individual verbs and nouns.\end{tabular}}\\
    \hline
    \end{tabular}
\end{table}

\begin{table}
    \centering
    \label{tab:parafeatures}
    \caption{Paragraph-level Features}
    \vspace{1mm}
    \begin{tabular}{|l|l|l|l|}
    \hline
    \multicolumn{2}{|l|}{\textbf{Feature}}& \multicolumn{2}{|l|}{\textbf{Description}} \\ 
    \hline
    \multicolumn{2}{|l|}{Paragraph Position}& \multicolumn{2}{|l|}{\begin{tabular}{@{}c@{}} Sentences are weighted according to their paragraph' position.\end{tabular}}\\
    \hline
    \multicolumn{2}{|l|}{Optimal Position}& \multicolumn{2}{|l|}{\begin{tabular}{@{}c@{}}A sequence of most important sentences are identified.\end{tabular}}\\
    \hline
    \end{tabular}
 \end{table}

\begin{table}
    \centering
    \label{tab:Corpusfeature}
    \caption{Corpus-based Features}
    \vspace{1mm}
    \begin{tabular}{|l|l|l|l|}
    \hline
    \multicolumn{2}{|c|}{\textbf{Feature}}& \multicolumn{2}{|c|}{\textbf{Description}} \\ 
    \hline
    \multicolumn{2}{|l|}{Signature word}& \multicolumn{2}{|l|}{\begin{tabular}{@{}c@{}} Frequency of word occurrence averaged across a large corpus is used.\end{tabular}}\\
    \hline
    \multicolumn{2}{|l|}{Baseline Probability}& \multicolumn{2}{|l|}{\begin{tabular}{@{}l@{}}Using baseline documents, we define a term’s importance\\such that more frequent words have higher probability.\end{tabular}}\\
    \hline
    \multicolumn{2}{|l|}{Document Probability}& \multicolumn{2}{|l|}{\begin{tabular}{@{}c@{}} Estimates a term's likeliness of within a document.\end{tabular}}\\
    \hline
    \end{tabular}
 \end{table}

\subsection{Learning Optimal Feature Set}
Features are applied individually or collectively according to the application and the proposed model.
Typically, a different set of combinations is required to obtain optimal results.
Rafael et al.~\cite{ferreira2014context} tried combining different word-, sentence- and graph-level features for scoring sentence.
Meena and Gopalani~\cite{meena2014analysis} evaluated available features and analysed the results of combining different features.
As discovered, finding the optimal feature set remains a challenge.
One solution is to test different combinations of features and report the best feature for each document~\cite{meena2015optimal}.
However, as the number of features increases, the approaches become less practical.
Besides, most existing approaches give equal weight to all features\textemdash a solution in which each feature is weighted based on its respective importance.
To the best of our knowledge, there is no feature weighting algorithm for extractive document summarisation.

In addition to features, different models also have been suggested for recommending features.
Fattah and Ren~\cite{fattah2009ga} proposed supervised models including genetic algorithms (GAs), probabilistic neural networks, feed-forward neural networks, mathematical regression, and a Gaussian mixture model.
The authors used various features to train the summarisation model, including sentence position, relative sentence length, positive and negative keywords, sentence resemblance to the title, named entity in the sentence,  sentence centrality, numerical data, and aggregate similarity. 
Elsewhere, Prasad and Kulakarni~\cite{shardan2010implementation} used word similarity among paragraphs, iterative query scoring, word similarity among sentences, as well as a format-based score, term frequency, cue words, and tile similarity as features to score sentences.
Abuobieda et al.~\cite{abuobieda2012text} further used title feature, sentence position, numerical data, sentence length and thematic words.
In another study, Mendoza et al.~\cite{mendoza2014extractive} used title similarity, sentence position, cohesion sentence length and coverage as the features.
Optimisation techniques were used with evolutionary algorithms to generate summaries. 
However, the goal of weighting in these problems is to measure the importance of sentences and to select them, not to weight the features~\cite{garcia2013single,mendoza2014extractive}.
 
Recent approaches use deep learning methods such as convolutional neural networks~\cite{cao2015learning,DBLP:journals/cee/NiuXAPBA20}, a recurrent neural network (RNN)~\cite{cheng2016neural,nallapati2017summarunner}, or a combination of the two~\cite{wu2018learning,narayan2018ranking}.
Consequently, they do not need handmade features.
Although these approaches could gain outstanding results in terms of performance, they are neither efficient nor interpretable, and even require a large volume of training data\textemdash one of the key challenges in summarisation.
Besides, they cannot estimate each feature’s role in the summarisation task and for each class separately.

\section{Traditional Summarisation}
\label{TraditionalApproach}
The main categorisations in document summarisation are abstractive, extractive and hybrid approaches.
Each of these categories and their subcategories is explained in this section.
Since the focus of this dissertation is on extractive approaches, they are analysed in greater detail.

\subsection{Abstractive Approaches}
Abstractive summaries are generated by interpreting the main concepts of a document, and then stating those contents in a clear and natural manner~\cite{erkan2004lexrank,barros2019natsum}.
Abstraction techniques are a substitute for the original documents rather than a part of them.
Therefore, abstractive approaches require deep NLP such as semantic representation and inference. 
However, they are challenging to produce and need time to mature.
The reason is that today’s systems and computing devices cannot provide semantic representation, inference and natural language to such a level that is equivalent to humans.
Abstractive summarisation approaches are mainly classified into two groups: structure-based approaches and semantic-based approaches~\cite{heu2015fodosu,mohamed2019srl}.
The structure-based approach aims to find a schema that can describe the document, including template-based methods, rules-based methods, and ontology- and tree-based procedures. 
In semantic-based strategies, a document’s semantic structure is used, including linguistic data (i.e., noun and verb phrases) and semantic graph-based approaches.

\subsection{Extractive Approaches' Structure}
Extractive text summarisation approaches select some sentences as representative of the original document(s).
These sentences are then concatenated into a shorter text format to produce a meaningful and coherent summary~\cite{mehta2018effective}.
An extractive approach usually contains three steps, including representation of the original text document, sentence scoring, and then selecting high-scoring sentences.
These three steps are the basis of various categories of extractive approaches.

\subsubsection{Representation Model}
Before performing any summarisation algorithm, the original text requires to be converted to an intermediate representation model.
Generally, there are two main representation models: ‘topic’ representations and ‘indicator’ representations~\cite{allahyari2017text} which are different considering the representation model and complexity.
Indicator representation approaches represent sentences in the form of features (indicators), discussed in Sec.~\ref{featureEngineer}.
Topic representation approaches convert the text into an intermediate model and interpret the text.

\subsubsection{Sentence Scoring}
\label{sentenceScoring}
When an intermediate representation is generated, each sentence is scored based on its importance.
A sentence’s score is computed based on the covered topic in the original documents in a topic representation approach.
Conversely, indicator representation methods aggregate the evidence from different indicators using techniques such as machine learning.

\subsubsection{Generating Summaries}
\label{generateSum}
Eventually, the summariser selects the top most important sentences based on a size limit parameter to generate the final summary.
Greedy algorithms such as converting the sentence selection into an optimisation problem are also used to select important sentences.
In this setting, a set of sentences is chosen to maximise overall importance and minimise redundancy.

\subsection{Extractive Approach Categorisation}
This dissertation categorises extractive approaches as statistical approaches, graph-based approaches, knowledge-based approaches, and machine learning approaches.

\subsubsection{Statistical (Early) Approaches}
Early approaches on document summarisation combined various features (defined in Sec.~\ref{featureEngineer}) to score the relevancy of selected text units.
However, combining some features does not guarantee summary improvement.
Two significant approaches in this category are Lead-3 and the phrase-based ILP model~\cite{woodsend2010automatic}.
Lead-3 selects the three leading sentences as the summary, and the phrase-based ILP model~\cite{woodsend2010automatic} is based on a linear programming formulation that learns to combine phrases considering such features as coverage, length and grammar constraints.
Although this category relies on shallow features, these two approaches still report promising results.

\subsubsection{Graph-based Approaches}
Graph theory has been used in many approaches for representing the semantic structure of a document~\cite{vodolazova2013extractive}; therefore, it is appropriate for document summarisation tasks.
For graph-based methods, each text element (words or sentences) is treated as a node~\cite{das2007survey,hariharan2009studies,erkan2004lexrank,hariharan2013enhanced}.
Two nodes or specifically two sentences are connected with an edge if they share some similarities.
Two types of graphs are used to represent text: these are 'lexical graphs', which use the lexical features of the text and 'semantic graphs', which use the text’s semantic properties, such as the ontological relationship between words.
The relationship between a set of words represents the sentences’ syntactic structure (dependency tree and syntactic trees).
TGRAPH~\cite{parveen2015topical} and URANK~\cite{wan2010towards} are two significant approaches that use a graph model and ranking scores for each sentence obtained in a unified ranking process.
Further, TextRank~\cite{mihalcea2004textrank} and LexRank~\cite{erkan2004lexrank} use lexical features to create a graph.
The difference between the two is that the similarity measure in TextRank is estimated by counting the number of similar words between two sentences, whereas LexRank utilises the cosine similarity among sentences ~\cite{alami2019enhancing}.
Other approaches to find similarity measures between the nodes involve discounting, the cumulative sum method~\cite{hariharan2009studies} and position weight~\cite{hariharan2013enhanced}.

\subsubsection{Clustering-based and Frequent Term Approaches} 
Clustering-based approaches group together the related information retrieved from similar documents and passages.
After clustering, sentences are ranked within each cluster.
Their salience scores are calculated, and high-scoring sentences from each cluster are extracted to form the summary~\cite{hariharan2009studies}.

Conversely, frequent term approaches seek the frequent and semantically similar terms in documents~\cite{Naresh}.
Semantic similarity checks the path’s length linking the terms and further measures the content difference and similarity.
The summariser selects sentences including the most frequent and semantically related terms.

\subsubsection{Knowledge-based Summarisation Approaches}
The automatic text summarisation aims to create summaries similar to human-created summaries.
One solution is to combine summarisation techniques with knowledge bases (semantic-based or ontology-based summarisers) such as Wikipedia, YAGO and DBpedia to consider the semantics of words.
For example, Sankarasubramaniam et al.~\cite{sankarasubramaniam2014text} employed Wikipedia along with a graph-based ranking technique.
Another example is the YAGO-based summariser [100] that utilises YAGO ontology~\cite{suchanek2007yago} to classify key concepts in a document set.

\subsubsection{Optimisation-based and Machine Learning Approaches}
This dissertation characterises recent, state-of-the-art summarisation approaches as an optimisation problem, a fuzzy problem or a machine learning problem.

\textbf{\textit{Summarisation as an optimisation problem.}}
Optimisation algorithms have been widely used for various summarisation purposes. 
The most common algorithms are the GA~\cite{sanchez2020experimental} and particle swarm optimisation ~\cite{sanchez2018extractive}.
A GA is a search-based optimisation algorithm inspired by two concepts of evolution and population genetics.
It generates random solutions and optimises population by applying natural operations such as selection, mutations and crossover.
GAs also create an initial population randomly.
The algorithm evaluates each population member based on a defined fitness measure and assesses according to some preferred requirements.
The algorithm then selects some individuals while favouring higher fitness (selection) and generating new samples by combining the chosen individuals’ features (crossover).

Particle swarm optimisation is another commonly applied bio-inspired algorithm used to obtain an optimal solution, inspired by birds’ social movement~\cite{mandal2019pso,priya2019enhanced}.
It initiates with a randomly discovered population of particles with a random position and a velocity. 
Then, a velocity vector is calculated for each individual.
Their respective position is updated using its former position and the recently renewed velocity vector before converging.

\textbf{\textit{Summarisation as a fuzzy problem.}}
Fuzzy logic has been widely used in summarisation systems.
Fuzzy systems take text features as input, and the algorithm transforms them to a fuzzy linguistic values (fuzzifier).
Fuzzy rules in the form of ‘if-then’ statements are used to generate the outputs~\cite{el2020automatic,patel2019fuzzy}.
he fuzzy evolutionary optimisation method (FEOM)~\cite{song2011fuzzy} is another approach in this category, which clusters the documents and selects the most important sentence in each group.

\textbf{\textit{Machine learning for summarisation.}}
Machine learning algorithms are widely used for summarisation using supervised, unsupervised and semi-supervised approaches.
Supervised learning algorithms require a large amount of labelled data for training.
These algorithms model summarisation as a binary classification sorted either as ‘in summary’ or ‘not in summary’.
Therefore, different machine learning algorithms used for classification can also be used for this purpose.
Commonly used supervised learning algorithms include regression, support vector machines, naïve Bayes classification, and decision trees~\cite{das2007survey,gupta2010survey}.

Unsupervised learning algorithms aim to discover the hidden structure of data without the need for labelled data.
Commonly used techniques include clustering and the hidden Markov model (HMM).
The ‘query, cluster and summarise’ technique~\cite{dunlavy2007qcs} is another system based on the HMM that computes the probability of each sentence as being appropriate for the summary set.
Meanwhile, semi-supervised learning algorithms require both labelled and unlabelled data, whereas conditional random fields~\cite{shen2007document} consider the summarisation task a sequence-labelling problem.

Recently, the focus has shifted to neural network-based and deep RL methods, which could show promising results.
Both employ word embedding~\cite{pennington2014glove} to represent words at the input level, and then feed this information to the network to gain the output summary. 
These models mainly use a convolutional neural network~\cite{cao2015learning}, an RNN~\cite{cheng2016neural,nallapati2017summarunner} or a combination of the two~\cite{wu2018learning,narayan2018ranking}.
Although these approaches could gain outstanding results in terms of performance, they are not efficient and interpretable.
Neural network sentence extraction (NN-SE)~\cite{cheng2016neural} is a neural network model composed of a hierarchical document encoder and an attention-based extractor; SummerRuNNer~\cite{nallapati2017summarunner} is an RNN-based sequence model; and the hierarchical structured self-attentive model for extractive document summarisation (HSSAS)~\cite{al2018hierarchical} is a neural network model with a hierarchical structured self-attention mechanism to create both sentence and document embedding.
Last is BanditSum~\cite{dong2018banditsum}, a neural network model that considers summarisation a contextual bandit (CB) problem.

Lately, RL approaches have been proposed for both extractive and abstractive summarisation~\cite{ryang2012framework,pasunuru2018multi,paulus2018deep}.
Most existing RL-based approaches use heuristic and greedy functions as the reward function and, therefore, do not require reference summaries~\cite{ryang2012framework,rioux2014fear}.
Other approaches use different recall-oriented understudy for gisting evaluation (ROUGE) measure variants as the reward function and, therefore, require reference summaries to reward RL~\cite{pasunuru2018multi,paulus2018deep, kryscinski2018improving}.
The reward quality is a bottleneck for RL-based summarisation approaches~\cite{gao2019reward}.

\subsection{Hybrid Extractive and Abstractive Approaches}
Hybrid approaches combine both abstractive and extractive techniques.
A hybrid approach commonly consists of an extractive phase to extract the key sentences from the input text and an abstractive phase to generate the final abstractive summary. 
The two approaches are interrelated and the overall summarisation performance is enhanced.
However, the research community focuses more on extractive techniques since their abstractive counterparts are highly complex and need extensive NLP.
One example is an approach that uses a graph model to obtain the key sentences in the extraction phase, and an RNN-based encoder–decoder for abstraction phase~\cite{wang2017integrating}.

SumItUp~\cite{bhat2018sumitup} is another example that employs some statistical features and a semantic feature (emotion described by the text) to generate a summary.
For removing redundant sentences, Cosine similarity is used.
A language generator takes the extracted sentences to transform the extractive summary into an abstractive summary.
Sentences are reordered to retain the original sequence.

\section{Structured Summarisation}
\label{structured}
Traditional summarisation approaches are incapable of producing more extended summaries since all details are omitted, even if the user is interested in obtaining more information.
Besides, the produced summaries are unstructured and, therefore, difficult to unpack for further analysis.
This prompts the need for structured summarisation.
Structured summaries are defined by generating Wikipedia articles and biographies~\cite{liu2010biosnowball}.
We categorised structured summarisation approaches into four groups, covering (i) timeline summarisation approaches, (ii) document thread summarisation, (iii) hierarchical summarisation, and (iv) concept map summarisation.

\subsection{Timeline Summarisation}

Timeline approaches commonly produce a short summary to form a story based on dates.
Using partial ordering relations~\cite{kolomiyets2012extracting} to link the events in a narrative and a temporal representation of events according to time intervals~\cite{do2012joint} are examples of timeline summaries.

Some approaches emphasise the summarisation aspect for generating timelines from multiple articles.
One example is to formalise generating timeline problems as an optimisation problem that balances coherency, diversity and summary quality~\cite{yan2011evolutionary}.
Another model is a summarisation-based approach to create timelines based on the inter-date and intra-date sentence dependencies~\cite{yan2011timeline}.
Other approaches identify the most important dates and the bursts of news that surround them, and then categorise events based on the burst time~\cite{swan2000automatic,chieu2004query,akcora2010identifying,hu2011generating,kessler2012finding}.

\subsection{Document Thread Summarisation}
Discovering threads of \textit{related} documents is another category of structured summaries.
These mostly employ a machine algorithm to find the threads using a supervised approach.
Features include the temporal sector of stories for recognising events and order of time to capture dependencies~\cite{nallapati2004event}.
Others used a hybrid clustering and topic modelling approach to cluster news articles~\cite{ahmed2011unified}, or statistical models to detect trends and topics from document~\cite{tang2012tut}.

Identifying \textit{coherent} threads of documents is another category among structured summarisation algorithms.
One proposed algorithm formulated components of an article chain and connected two specified articles~\cite{shahaf2010connecting}.
In the literature, Gillenwater et al.~\cite{gillenwater2012discovering}  a probabilistic technique was proposed to extract a set of threads from a document set.
Shahaf et al.~\cite{shahaf2012trains} extended the work by implementing the idea of metro maps in scientific areas~\cite{shahaf2012metro}.

\subsection{Hierarchical Summarisation}
The relationship between summarisation and hierarchies are analysed in the literature~\cite{haghighi2009exploring,takahashi2007hierarchical,lawrie2001finding}.
However, the hierarchy in these approaches is the relation between different elements of a document, such as words or phrases~\cite{haghighi2009exploring,takahashi2007hierarchical,lawrie2001finding}.
A hierarchy is also defined as a structure prioritising more general information~\cite{celikyilmaz2010hybrid,ouyang2009integrated} or spreading the summary out across the hierarchy~\cite{yang2003fractal,wang2006multi}.
A recent hierarchical summarisation approach is SUMMA, which produces a hierarchy of relatively short summaries.
The hierarchy is based on time intervals and, therefore, can also be considered a timeline approach~\cite{christensen2014hierarchical}.

\subsection{Concept Map Summarisation}
Concept map MDS approaches produce structured summaries in the form of concept maps.
A concept map that extends the mind map idea introduced by Novak and Gowin~\cite{novak1984learning} is a labelled graph showing concepts as nodes and the relations between them as edges.
Different techniques have been suggested for single documents~\cite{oliveira2001automatic,villalon2009concept,leake2006jump,aguiar2016automatic} and multiple documents~\cite{rajaraman2002knowledge,zouaq2008building,zubrinic2012automatic,qasim2013concept}.
Different document models have also used concept maps, including in scientific papers~\cite{qasim2013concept}, legal documents~\cite{zubrinic2012automatic}, student essays~\cite{villalon2009concept} and general webpages~\cite{rajaraman2002knowledge}.

The first step in creating a concept map is to extract the concepts and relation spans from the input documents.
Extracted mentions refer to the same concept or the relations that require grouping.
Concept and relation labelling and importance estimation are the final steps in creating a summarised concept map.
The most recent approach in concept summarisation was proposed by Falke~\cite{falke2019automatic}.
This approach learns to identify and merge coreferent concepts to reduce redundancy, determine their importance with a robust supervised model, and find an optimal summary through ILP.

\section{Interactive and Personalised Summarisation}
\label{InteractiveApproach}
Interactive NLP algorithms ask users to provide certain feedback forms to refine the model and generate higher-quality outcomes tailored to the user. 
Multiple forms of feedback have also been studied for different applications including mouse clicks for information retrieval~\cite{borisov2018click}, post-edits and ratings for machine translation~\cite{denkowski2014learning, kreutzer2018can}, error markings for semantic parsing~\cite{lawrence2018counterfactual}, and preferences for translation~\cite{kingma2014adam}.

Most existing computer-assisted summarisation tools present essential parts of documents to the user using a traditional automatic summarisation algorithm, and then ask users to refine the results without further interaction.
The refining process includes cutting, pasting and reorganising the essential elements to formulate a final summary~\cite{orasan2006computer,craven2000abstracts,narita2002web}.
Other works present automatically derived hierarchically ordered summaries that allow users to drill down from a general overview to detailed information~\cite{christensen2014hierarchical,shapira2017interactive}.
Therefore, these systems are neither interactive nor consider user feedback to update their internal summarisation models.
Other interactive summarisation systems include the iNeATS~\cite{leuski2003ineats} and IDS~\cite{jones2002interactive}, which allow users to tune several parameters (e.g., size, redundancy and focus) for customising the produced summaries.
Avinesh and Meyer~\cite{avinesh2017joint} proposed a more recent interactive summarisation approach that asks users to label important bigrams within candidate summaries.
Their system can achieve near-optimal performance in 10 rounds of interaction in simulation experiments, collecting up to 350 critical bigrams.
However, labelling important bigrams is an enormous burden on users, as they have to read many potentially unimportant bigrams.

There is increasing research interest in using preference-based feedback and RL algorithms in summarisation.
For example, one approach learns a sentence ranker from human preferences on sentence pairs~\cite{zopf2018estimating}.
The ranker is then used to evaluate the quality of summaries by counting the number of high-ranked sentences included in a summary.
This preference-based RL algorithm has also been used in summarisation. The structured prediction from partial information (SPPI)~\cite{sokolov2016stochastic,kreutzer2017bandit} is a policy-gradient RL algorithm that receives rewards from the preference-based feedback.
The problem is that SPPI suffers heavily from the high sample complexity problem.

Another recent preference RL approach is APRIL~\cite{gao2018april}, which has two stages.
First, the user’s ranking over candidate summaries is retrieved, and then a neural RL agent is used to search for the optimal summary.
However, favouring one summary to another in both approaches places considerable burden on users.
It is worth re-mentioning that summarisation aims to provide users with a summary that reduces the need to read multiple documents.
However, asking users to prefer a summary to another in multiple rounds among a summary space that includes all randomly possible combinations of sentences only adds more cognitive load.

\section{Summary}
\label{ch2_summary}
This chapter provides an overview of document summarisation (Sec.~\ref{overview_ch2}) and explores several different categories based on various parameters, including input size, purpose type, output properties, and language and applications, with brief descriptions of each (Sec.~\ref{TraditionalCategories}). 
Also proposed was a new categorisation schema based on current gaps in state-of-the-art approaches to summarisation (Sec.~\ref{proposedcategories}).
The proposed categorisation schema includes traditional approaches, structured approaches, and interactive and personalised summarisation approaches.

Sec.~\ref{featureEngineer} outlined the existing methods and techniques for feature engineering in document summarisation, including the various proposed features.
Traditional approaches\textemdash including extractive, abstractive, and hybrid extractive and abstractive approaches\textemdash are discussed in Sec.~\ref{TraditionalApproach}.
Sec.~\ref{structured} then explored the different structured summarisation approaches, including timeline and document thread summarisation, hierarchical summarisation, and concept map summarisation, followed by an explanation of various interactive and personalised approaches (Sec.~\ref{InteractiveApproach}).
In summary, there are four main limitations facing the current summarisation approaches, addressed in the following chapters:
\begin{itemize}
    \item There is a need for intelligent automatic feature engineering in summarisation that can capture each feature’s importance based on the defined score.
    \item Flexible and interactive approaches are needed to facilitate improved summary navigation based on users’ interests.
    \item There is a lack of personalised summarisation approaches that can predict the desired summary for a specific user based on their individual behaviour.
\end{itemize}

%% file: ch_3/ch_ExperimentalSetup.tex
\chapter{Experimental Setup}
\label{ch_3}
This chapter outlines the experimental set-up, the datasets used in our experiments, and the evaluation metrics to evaluate the proposed models.
These factors are considered when comparing each of the proposed approaches in this thesis.

\section{Dataset}
Different datasets are created for various summarisation applications.
We used the same dataset to evaluate the proposed approaches in this dissertation: these are the DUC dataset and the CNN/Daily Mail dataset.

\subsection{DUC Datasets}
DUC datasets are the most common datasets used in text summarisation research, provided by the National Institute of Standards and Technology.
DUC datasets are released online as part of the summarisation shared task hosted at the DUC each year.~\footnote{https://duc.nist.gov/}
However, access to the data requires permission.
Each dataset contains both documents and their corresponding summaries, which are created manually or automatically, either as baseline summaries or generated by challenge participants’ systems.
These datasets are in English and are sourced from the news domain. They can be used for both single-document summarisation and MDS.
In this dissertation, we used DUC2001, DUC2002 and DUC2004.
DUC2001 contains 60 sets of approximately 10 documents that cover various subjects.
DUC2002 contains 567 document summary pairs divided into 59 clusters.
DUC2004 contains 100 sets of approximately 10 documents that cover various subjects.
To forge a valid comparison between the approaches, we used an experimental setting similar to the state-of-the-art approaches explained in each section.

\subsection{CNN/Daily Mail Datasets}
The CNN/Daily Mail dataset~\cite{hermann2015teaching} is an English dataset created for passage-based question-answering tasks.
However, a modified version of this corpus has been extensively used for evaluating summarisation systems~\cite{nallapati2017summarunner}.
The dataset is in English, is publicly available, and contains online news articles.~\footnote{https://github.com/abisee/cnn-dailymail}
Each dataset comprises a set of document clusters accompanied by several human-generated summaries used for training and evaluation purposes.
The dataset includes news articles (781 tokens on average) matched with multi-sentence summaries (3.75 sentences or 56 tokens on average), 287,226 training pairs, 13,368 validation pairs and 11,490 test pairs.

\section{Evaluation Metric}
One approach to evaluate produced summaries is by comparing the generated summary and the reference summary.
Comparing summaries to the original text helps to understand the measures, including information loss.
Conversely, comparison to a reference summary will quantify the quality of summaries against humans.
In both situations, the evaluation strategy is deemed ‘intrinsic’ in nature, since it is compared against itself as a content evaluation method or to verify linguistic aspects of the output summary, including the grammar, coherence and reference clarity. coherence~\cite{steinberger2012evaluation}. 
To perform a comprehensive evaluation of the proposed approach, we categorised evaluation as ‘automatic evaluation’ and ‘human evaluation’ (discussed in Sec.~\ref{AutoEval} and ~\ref{HumanEval}).
Automatic summarisation approaches are also used to perform other tasks such as information retrieval, translation, or question answering.
Therefore, one strategy is to evaluate the summarisation approach towards a specific task, known as an ‘extrinsic method’.

\subsection{Automatic Evaluation}
\label{AutoEval}
There are some conferences with a primary role in designing evaluation standards for automatic scoring of summaries and human evaluation~\cite{over2007duc}.
ROUGE~\cite{lin2004rouge} is the most commonly accepted metric for evaluating summaries, which automatically determines the summary quality by comparing it to human (reference) summaries.
It computes the number of common units (n-grams) in both the system’s summary and the reference summary.
ROUGE-N is a recall-based measure and is based on a comparison of n-grams.
Eq.~\ref{rouge} describes how ROUGE-N is calculated.

\begin{equation}
\label{rouge}
    ROUGE_n= \frac{\sum_{S\in \{Reference Summaries\}} \sum_{gram_n \in S}{Count_{match}(gram_n)}}{\sum_{S\in \{Reference Summaries\}} \sum_{gram_n \in S}{Count(gram_n)}},
\end{equation}

where $n$ is the n-gram size, $Count_{match}(gram_n)$ is the number of common n-grams in the candidate and the reference summaries, and $Count(gram_n)$ is the number of n-grams in the reference summary.

ROUGE-L employs the concept of longest common sequence between the two sequences of text.
Although this metric is more flexible, its drawback is that all n-grams must be consecutive~\cite{lin2004rouge}.
Three variants of ROUGE (ROUGE-1, ROUGE-2 and ROUGE-L) are used in this dissertation.
ROUGE-1 and ROUGE-2 were used to evaluate informativeness, and ROUGE-L (LCS) was used to assess fluency. 
We used the limited-length ROUGE recall-only evaluation (75 words) to compare the proposed approach and DUC dataset to avoid bias.
Besides, the full-length F1 score was used to evaluate the CNN/Daily Mail dataset.

\subsection{Human Evaluation}
\label{HumanEval}
While ROUGE serves as a rough measure of coverage, it only compares the n-gram units~\cite{ermakova2019survey}.
Since our goal is to advance personalised approaches, ROUGE cannot be a useful measure.
Therefore, we conducted human experiments to evaluate the proposed models based on various criteria.
The details of the experiments are explained in each subsection. The general setting is explained as follows.

We hired Amazon Mechanical Turk (MTurk)~\footnote{https://www.mturk.com/} workers to attend tasks without any specific prior background required.
We designed a series of micro-tasks for each experiment. 
Not recently published articles were selected for the experiments to avoid any bias in understanding the topics.
To ensure the human subjects understood the study’s objective, we asked workers to complete a qualification task first.
Participants were asked to write a summary of a news article.
The results that did not have logical meaning or structure were noted as spam and, thus, removed manually from the results.
For example, in the qualification tasks, we asked users to write a summary explaining the main parts of a document.
Some results could not pass the qualification task.
Another example is the short response time allocated for the tasks, intended to prove that the answers recorded are random or not provided in advance by the workers.
Four evaluation aspects were analysed.
These are (i) information coverage (how much information the summary covers), (ii) knowledge extraction (how much users can learn from summaries), (iii) effectiveness (the speed at which users learn) and (iv) user preference (the users’ preference(s) compared to other approaches).

\section{Baseline}
We compared the proposed approaches to various previously published models known for their significant performance on the datasets.
The basic state-of-the-art approaches are explained as follows.

Two early approaches with particular significance are ‘Lead-3’, which selects the three leading sentences as the summary, and the ‘phrase-based ILP model’~\cite{woodsend2010automatic}, which is based on a linear programming formulation that learns to combine phrases considering such features as coverage, length and grammar constraints. 
Although these approaches rely on shallow features, they still report promising results.

Next are TGRAPH~\cite{parveen2015topical} and URANK~\cite{wan2010towards}.
These two graph-based approaches use graph modelling and ranking scores for each sentence obtained in a unified ranking process.
Differential evolution~\cite{aliguliyev2009new} is another approach in basic machine learning algorithms that optimises the allocation of sentences.
The selection of sentences is based on a recursive scheme.

Next is FEOM~\cite{song2011fuzzy}, which clusters documents in a dataset and then selects the most important sentence of each group.
Querying, clustering and summarising~\cite{dunlavy2007qcs} are based on the hidden HMM, which computes the probability of each sentence’s appropriateness for the summary set.
Finally, the conditional random field~\cite{shen2007document} considers summarisation tasks as sequence-labelling problems.

Most prominent neural network models include the NN-SE~\cite{cheng2016neural}, which harnesses a hierarchical document encoder and an attention-based extractor, and SummaRuNNer~\cite{nallapati2017summarunner}, an RNN.
Further, HSSAS~\cite{al2018hierarchical} uses a hierarchical structured self-attention mechanism to create both sentence and document embedding, and BanditSum~\cite{dong2018banditsum} considers summarisation a CB problem.
This model receives a document and chooses a sequence of sentences to include in the summary.
The goal is to maximise the ROUGE score.

\section{Summary}
This chapter discussed the experimental set-up, the datasets used in our experiments, and the evaluation metrics to evaluate the proposed models.
We explained different evaluation perspectives, including intrinsic and extrinsic approaches.
Based on these two measures, we defined automatic evaluation using the ROUGE measure and human evaluation, and further presented the general setting using MTurk for the human evaluation component.
Also introduced were the baselines for comparison.

%% file: ch_4/FeatureEngineering.tex
\chapter{Towards Intelligent Feature Engineering}
\label{ch_4}
Automatic extractive summarisation approaches aim to rank sentences based on some defined features that reflect their importance.
Various features have been suggested for this purpose. 
However, it remains a challenge to produce a summary that best represents documents in a dataset.
This chapter focuses on answering the following questions about enabling automatic intelligent feature engineering for document summarisation purposes:
\begin{itemize}
    \item How good are the state-of-the-art methods in automatically creating and detecting features in the context of document summarisation?
    \item How can we evaluate the efficiency of different features in making summaries?
\end{itemize}

We propose a novel approach, called ExDoS, which benefits from supervised and unsupervised algorithms simultaneously and in an interpretable manner.
We combined the clustering and classification algorithm into a single objective function.
Clustering aims to discover the underlying structure of data, and then feeds the processed information to the classification stage through a single objective function.
ExDoS iteratively minimises the classifier’s error rate in each cluster by proposing a dynamic local feature weighting schema.
Moreover, ExDoS specifies each feature’s contribution in discriminating each class\textemdash a challenging task in summarisation.
The unique contributions of this chapter are as follows.

We introduce and formalise a theoretically grounded method based on the idea of combining supervised and unsupervised learning, and employ this for the task of document summarisation.
Since clustering aims to discover the underlying structure of data and feed this information to the classification stage through a single objective function, it improves the performance of the summarisation algorithm.
This architecture allows us to develop clusters of sentences that can help in selecting summaries.
Specifically, we designed a ranking measure that determines whether a document sentence matches a highlight, and should be labelled with ‘1’ (must be in the summary) or ‘$0$’ (not in the summary).

Second, ExDoS can measure the role of features in discriminating each class individually through the summarisation process by making various feature spaces.
Features are dynamically weighted through the optimisation process in each cluster.
These weights represent the role of each feature in discriminating each label individually while summarising documents.

Third, sentences are selected in a way that produced summaries are coherent and non-redundant.
The most crucial sentence will be at the top, and then other sentences are chosen to cover all critical information without redundancy.

Finally, ExDoS has an additional advantage which is interpretability.
The separated terms in the optimisation process allow us to track the output summary.
Such visualisation is beneficial to explain decisions made by the system to the end user.

\section{Introduction}
Among the various categories for extractive summarisation approaches are machine learning approaches, which, although a recent phenomenon, have been widely used.
Extractive summarisation can be either unsupervised or supervised.
In unsupervised approaches, the goal is to find representative sentences.
In supervised methods, the problem is a binary classification task where classes are defined as being/not being included in a summary~\cite{li2020text,rautray2019performance}.
Our proposed approach, ExDoS, benefits from using both supervised and unsupervised algorithms simultaneously and in an interpretable manner.
The rationale behind this is to harvest the advantages of both classification and clustering algorithms.
While classification uses the knowledge of labels, clustering extracts the hidden information based on features.
Therefore, combining these two approaches can provide many advantages for different problems~\cite{ghodratnama2015efficient,ghodratnama2020content}.

ExDoS obviates the need for feature engineering in the summarisation task.
Although the most critical phase in machine learning algorithms is feature extraction, prior work has mainly focused on the sentence selection process.
Recently, some attempts have been made to find an optimum feature set for the summarisation process.
These approaches consider each feature’s relevance as a binary problem (i.e., whether a feature is included in the feature set or not)~\cite{meena2015optimal}. 
In addition to summarising a set of documents, ExDoS can measure the importance of different features with the help of local feature weighting.
The local weights of features indicate how each feature contributes to making each cluster. ExDoS transforms the feature space into a new feature space by weighting features locally in each cluster.
This feature weighting process aims to close up the same-label samples and push different-label samples further.

\begin{figure*}[t]
    \centerline{\includegraphics[width=\textwidth]{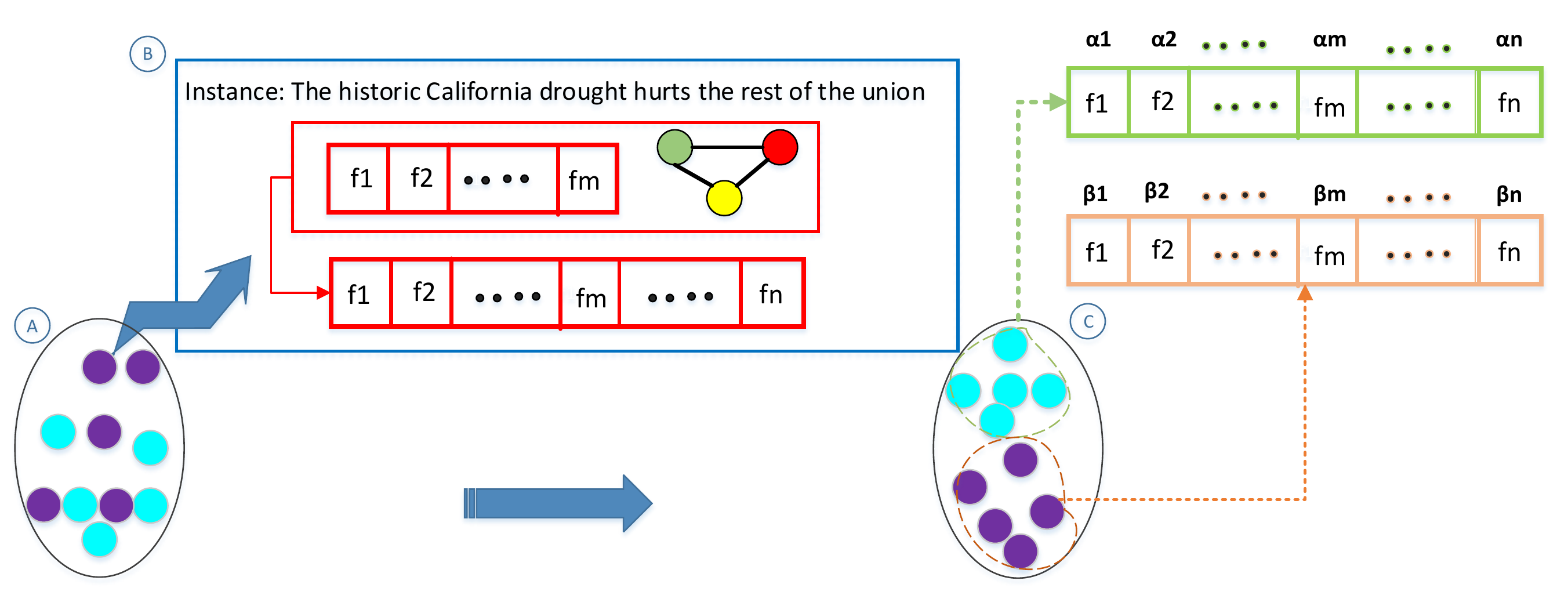}}
    \caption{Overview of the proposed approach (ExDoS). (A) A simple dataset with two classes. (B) Each instance is a sentence. We combine both surface and linguistic features (extracted from the semantic graph) to make a unified feature set. (C) The final output is groups of similar samples in which features are locally weighted in each group.}
    \label{ViewApproach}
\end{figure*}

\begin{figure*}[t]
    \centerline{\includegraphics[width=\textwidth]{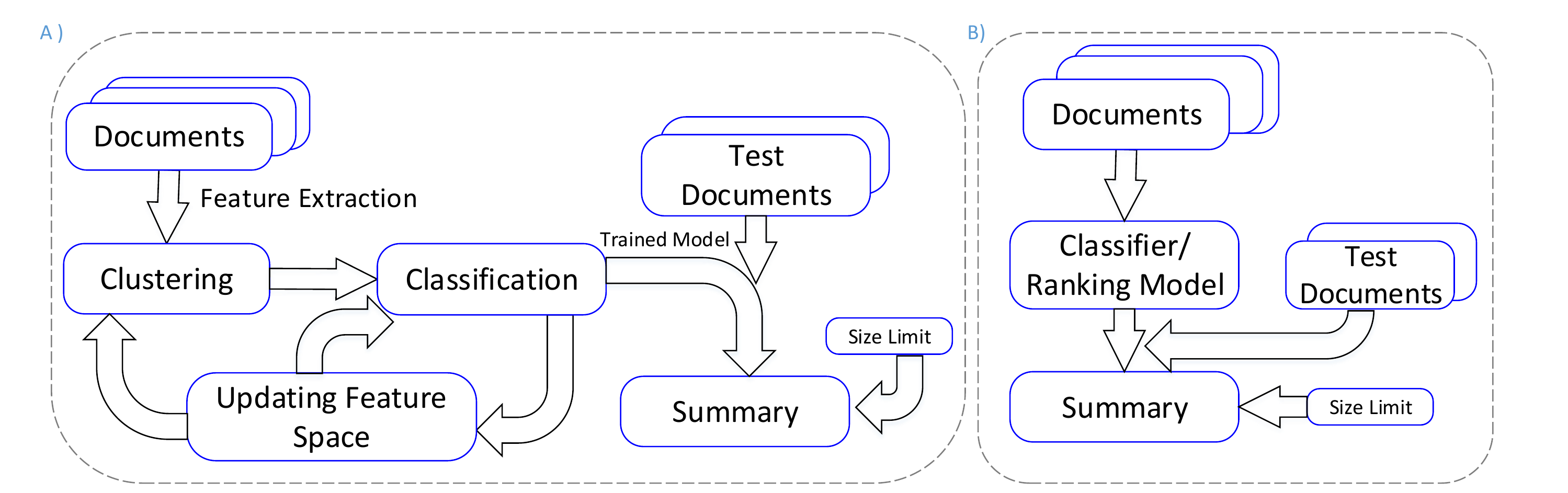}}
    \caption{(A) ExDos architecture. (A) The weights of features in each cluster are updated iteratively to bring similar samples closer to each other in the new feature spaces by minimising the error of classifiers in clusters. (B) The architecture of state-of-the-art approaches.}
    \label{ProposedArchitecture}
\end{figure*}

An overview of ExDoS is illustrated in Fig.~\ref{ViewApproach}, where a sample is a sentence modelled as a vector of features.
The final output is groups of similar samples where features are locally weighted in each group.
The weights of features illustrate the importance of each feature in subspaces (clusters).
Since the algorithm performs in an iterative manner using gradient descent, the simplest clustering (k-means) and k-nearest neighbour classification (KNN) algorithms are used to support efficiency.
However, k-means is one of the most reliable and widely used clustering algorithms.
Besides, the KNN classifier has been successfully used in many pattern-recognition applications.
It has been statistically proven that when $K=1$ (1NN), the probable error of 1NN would be less than twice the Bayes classifier error.
This proof states that 1NN is capable of generating near-optimal results.
Fig.~\ref{ProposedArchitecture} shows the architecture of ExDoS and how supervised and unsupervised approaches are combined to make new feature spaces.
As shown, weights of features in each cluster are updated iteratively to bring similar samples closer to each other in the new feature spaces by minimising classifier error in clusters.

\begin{figure*}[t]
    \centerline{\includegraphics[width=0.9\textwidth]{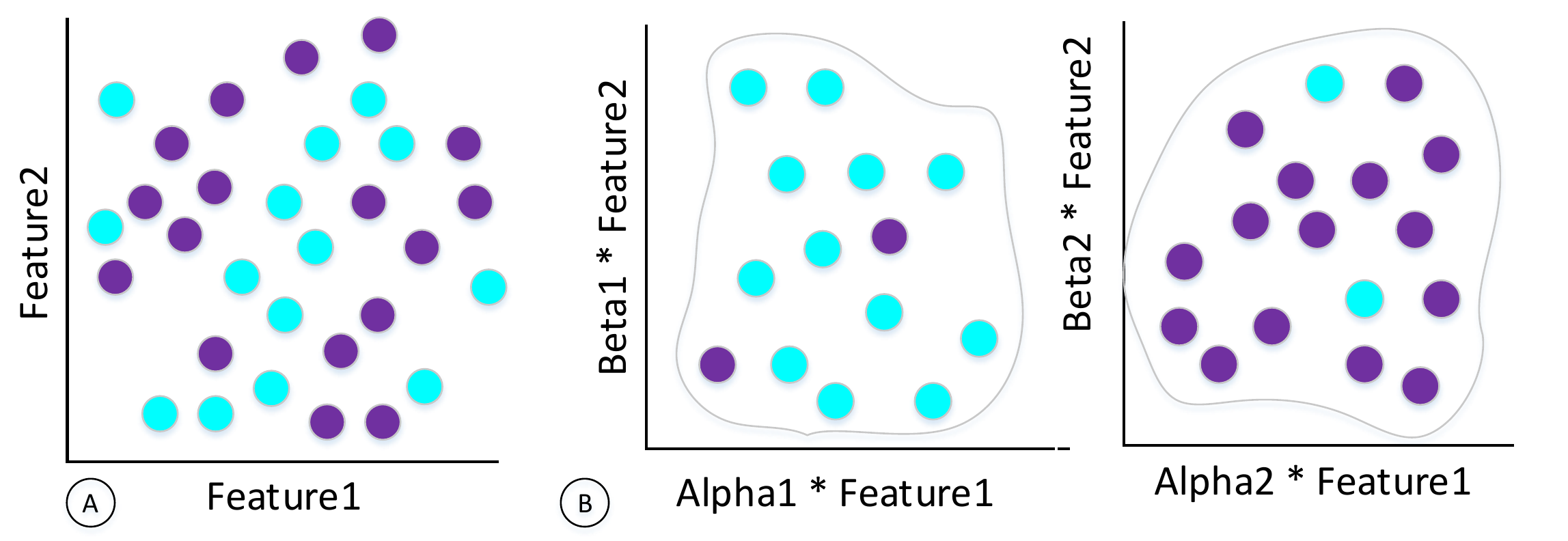}}
    \caption{(A) Distribution of synthetic data (2-class). (B) The output of the ExDoS is two new feature spaces where the weights (alpha and beta) are updated in such a way that similar samples are close to each other.*\newline
    *The details of this transformation are illustrated in Fig.~\ref{detailSamples}.}        
    \label{SampleExample}
\end{figure*}

\begin{figure*}[t]
    \centerline{\includegraphics[width=\textwidth]{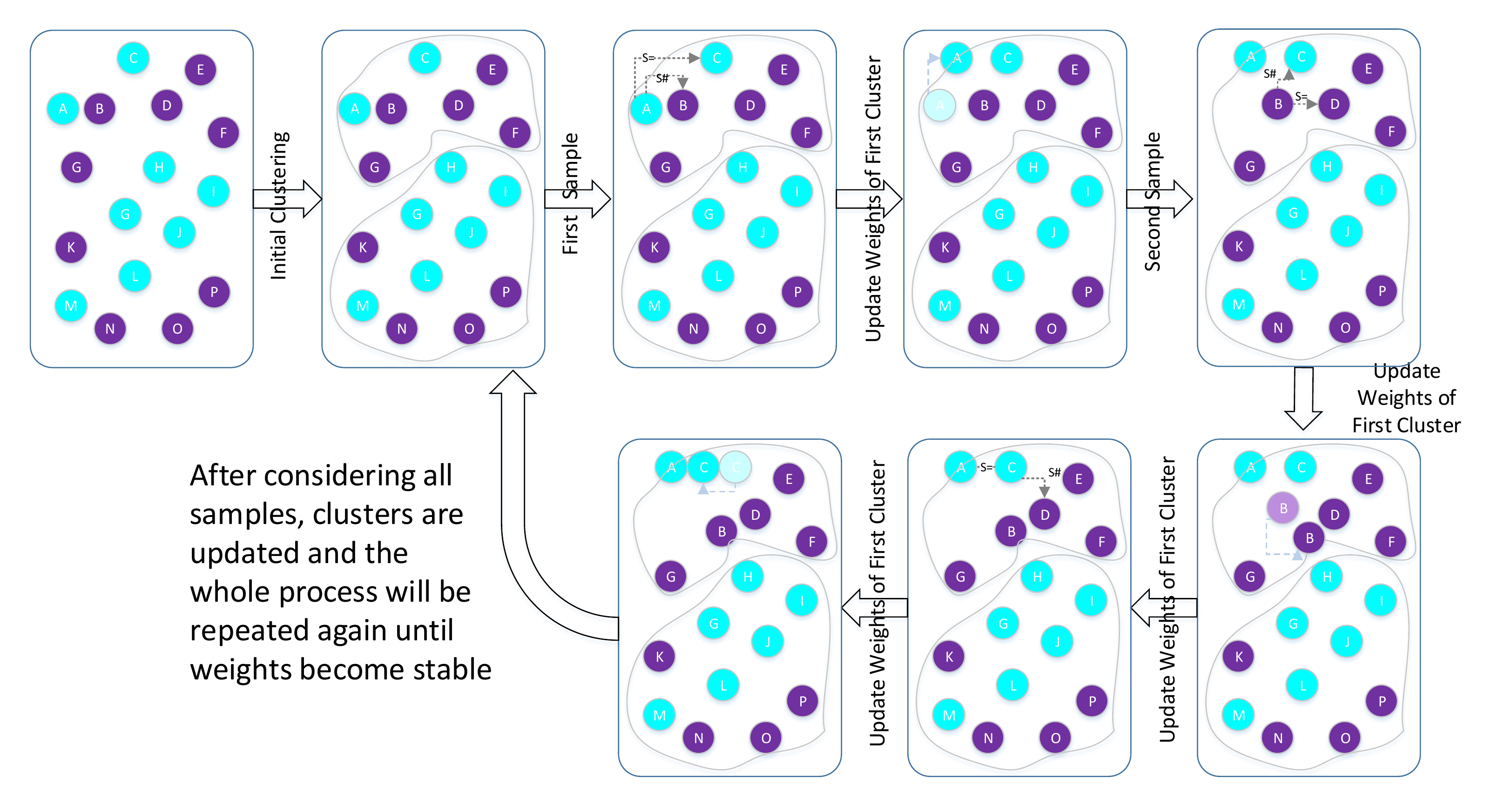}}
    \caption{
    Single iteration of ExDoS.
    All samples are considered in each iteration, and weights are updated to bring the nearest same-label sample ($s_=$) closer and push different-label samples ($s_{\neq}$) further.
    The local weights of features are subsequently updated.
    }                   
    \label{detailSamples}
\end{figure*}

We provide a detailed technical description of the proposed summarisation system throughout this chapter and illustrate its functionality using a working example.
A synthetic example of the ExDoS is illustrated in Fig.~\ref{SampleExample} and ~\ref{detailSamples}.
In Fig.~\ref{SampleExample}, the distribution of synthetic two-dimensional data (2-class) is depicted.
The output of the ExDoS is two new feature spaces where the weights (alpha and beta) are updated such that similar samples are close to each other.
The details of this transformation and how the output is derived are illustrated step by step in Fig.~\ref{detailSamples}.
We also evaluated our model both automatically (in terms of ROUGE factor) and empirically (human analysis) on two benchmark datasets (DUC2002 and the CNN/Daily Mail).
The accuracy of the approach, the importance of features, the effect of local feature weighting, the method’s complexity, and the parameters are all evaluated.

\subsection{Problem Statement}
The input is a set of documents $D=\{D_{1},D_{2}, ... ,D_{n}\}$, and each document consists of a sequence of sentences $S=[s_1,s_2,$$...$$,s_N]$.
Each sentence $s_i \in R^d$ is a sample vector corresponding to the $i$-th sentence, and $d$ is the number of features.
$Y=[Y_1, Y_2]$ is the class labels with two possible values of ‘1’ (being in summary) and ‘0’ (not being in summary).
$K$ is the number of clusters, and cluster centroids are denoted as  $C$, where $C_k$ is the centre of $k$-th cluster.
The sample $s_=$  is the closest sample with the same class label, and the sample $s_{\neq}$ is the closest sample with a different class label.
Also, $d_w$ denotes the weighted Euclidean distance.
Then, the goal is to learn a function $f:S\rightarrow Y$, which is defined on a given dataset $\{(s_1,y_1),(s_2,y_2),$$...$$,(s_m,y_m)\}$.

\subsection{Feature Set}
We explored a broad range of features that are commonly used for summarisation.
Two feature sets were defined to represent documents: these are surface-level sets and linguistic-level sets.
The first sets were extracted directly from the document, and the document was transformed into a semantic graph for the latter.

Essentially, ‘surface features’ contain frequency-based features (TF–IDF, residual IDF [RIDF], gain and word co-occurrence), word-based features (upper-case words and signature words), similarity-based features (Word2Vec and Jaccard measure), sentence-level features (position, length cut-off and length), and named entities.
Conversely, ‘linguistic features’ are categorised based on semantic graphs.
That is, for each sentence, a parse tree is constructed using the Stanford NLP tool~\cite{manning-EtAl:2014:P14-5}.
Each sentence is then summarised as a subgraph, which is a triple form.
To make the triples, we used an algorithm that extracts triples in the form of subject, predicate and object~\cite{rusu2007triplet}.
Subgraphs are connected to each other, where edges are annotated with similarity weights.
Similar or synonymous verbs (using Wordnet) are merged and subjects are concatenated.
Then, weights update as the average weights of two merged sentences.
Thus, linguistic features are composed of the average weights of connected edges, the merge status of a sentence as a binary feature, the number of sentences merged with a sentence, and the number of sentences connected with a sentence.

\subsection{Methodology}
ExDoS aims to discover the data’s underlying structure in the clustering phase and then feed this information to the classification stage in an iterative manner.
Therefore, a continuous objective function is defined for analytically optimising both clustering and classification stages by incorporating a new local feature weighting technique.
The nearest neighbour classifier’s error rate is minimised using the weighted distance, which overcomes the deficiency of popular Euclidean distance.
Moreover, the captured space for decision-making (in 1NN) by the Euclidean distance is a hyper-sphere.
The overall objective function is defined in Eq.~\ref{EqEDU}.
\begin{equation}
\label{EqEDU}
J(\textbf{W},\textbf{C})=J_1{(\textbf{W},\textbf{C})}+J_2{(\textbf{W})},
\end{equation}

where the first term ($J_1$) is the estimation error of clustering, and the second term ($J_2$) is the summation of the classification errors over the $K$ clusters.
These equations are expanded in Eq.~\ref{Eq:j1}, where $N_k$ is the number of samples in $k$-th cluster.

\begin{equation}
\label{Eq:j1}
J(\textbf{W},\textbf{C})=  \sum_{k=1}^{K}\sum_{i=1}^{\mid N_k \mid} d_{w_{}}^2 (s_i,C_k) + \frac{1}{N}\sum_{k=1}^{K}\sum_{i=1}^{\mid N_k \mid}S_{\beta}(\frac{d_w(s_i,s=)}{d_w(s_i,s_{\neq})})
\end{equation}

To estimate the error of 1NN, the following approximation function is used~\cite{paredes2006learning}:

\begin{equation}
\label{Eq:j2}
\frac{1}{N}\sum_{s\in S}^{}S_{\beta}(\frac{d_w(s,s_=)}{d_w(s,s_{\neq})})
\end{equation}

The sample $s_=$ is the nearest same-class sample, and the sample $s_{\neq}$ is the nearest different-class sample to the input sample $s$.
Respectively, $d_w$ is the weighted Euclidean distance and $S_{\beta}$ is the sigmoid function.
Two parameters are optimised in this objective function.
The feature-dependent weights associated with the sample $s$ are trained to make the $s_=$  closer to $s$ while making the sample $s_{\neq}$ further.
Then, the cluster centres update using the learnt weighted distance.
Since this function is differentiable, we can analytically use gradient descent, guaranteeing convergence for estimating the matrix $W$ and the centres.
The iterative optimisation of learning parameters are given in Eq.~\ref{Eq:WUp} and ~\ref{Eq:CUp}, where $\alpha$ and $\gamma$ are learning parameters.

\begin{equation}
\label{Eq:WUp}
W^{t+1}=W^{t}-\alpha(\frac{J(\textbf{W},\textbf{C})}{\delta(W)}) 
\end{equation}

\begin{equation}
\label{Eq:CUp}
C^{t+1}=C^{t}-\gamma(\frac{J(\textbf{W},\textbf{C})}{\delta(C)})
\end{equation}

To simplify the formula, the function $R(x)$ is defined in Eq.~\ref{Eq:Rx}~\cite{paredes2006learning}.

\begin{equation}
\label{Eq:Rx}
R(s_i)=(\frac{d_w(s_i,s_{i,=})}{d_w(s_i,s_{i,\neq})})
\end{equation}

The partial derivative of $J(W,C)$ with respect to $W$ is calculated in Eq.~\ref{Eq:JJ}.

\begin{equation}
\label{Eq:JJ}
        {\frac{\delta J(\textbf{W},\textbf{C})}{\delta W_k}} \cong \sum_{i=1}^{\mid N_k \mid} 2W_k \odot (x_i-C_k)^2+\frac{1}{N}\sum_{i=1}^{\mid N_k \mid} S_{\beta}^{'}(R(s_i))\frac{\delta R(s_i)}{\delta W_k},
\end{equation}

where $\odot$ is the inner product, and $\frac{\delta R(x_i)}{\delta W_k}$ is defined in Eq.~\ref{Eq:delta}.

\begin{equation}
\label{Eq:delta}
    \frac{\delta R(s_i)}{\delta W_k}=\frac{1}{{}d_{W_k}^2(s_i,s_{i,\neq})}({\frac{1}{R(s_i)}W_k\odot (xs_i-s_{i,=})^2}-R(s_i)W_k\odot{(s_i-s_{i,\neq})^2)}
\end{equation}

The derivative of $S_{\beta}(z)$ is defined in Eq.~\ref{Eq:beta}.

\begin{equation}
\label{Eq:beta}
S_{\beta}(z){'}=\frac{\delta S_{\beta}(z)}{\delta z}=\frac{\beta e^{\beta(1-z)}}{(1+e^{\beta(1-z)})^2}
\end{equation}

And the partial derivative of $J(\textbf{W},\textbf{C})$ with respect to $C$ is calculated using Eq.~\ref{Eq:derivC}.

\begin{equation}
\label{Eq:derivC}
\frac{J(\textbf{W},\textbf{C})}{\delta C_k} \cong \sum_{i=1}^{\mid N_k \mid} -2W_k^2 \odot(x_i-C_k)
\end{equation}

Since we need to optimise the features’ weight for cluster samples along with the centre of clusters, we first update $W$ in each cluster, and then update the centres ($C$).

\subsubsection{Generating Summary}
After training the model, we defined three measures, including coverage, coherence and redundancy, to generate summaries.

\subsubsection{Coverage}
The sentences’ coverage based on the proposed architecture is defined using Eq.~\ref{Eq:cov}.

\begin{equation}
\label{Eq:cov}
Cov(s_i)=\mid{d_{w}(c_{+},s_i)-d_{w}(c_{-},s_i)}\mid
\end{equation}

For each sentence, the weighted distance to cluster centres is estimated.
The coverage is defined as the difference between data points and two cluster centres.
$c_{+}$ is the cluster where the majority of samples belong to the positive class (being in summary), and ($c_{-}$) is the cluster where samples mostly belong to the negative class (not being in summary).
The summary coverage is the sum of all sentences’ coverage in that summary.

\subsubsection{Coherence}
A critical aspect of a good summary is coherence, or the summary order. For this purpose, we used G-Flow~\footnote{\url{http://knowitall.cs.washington.edu/gflow/}}~\cite{christensen_naacl13}, a graph model for selection and ordering that balances coverage and coherence.
G-Flow relies on the approximate discourse graph, where each node is a sentence, and edges indicate whether a sentence coherently follows one other.
The indicators include coreference, discourse cues, de-verbal nouns, and more.
The coherence is defined in Eq.~\ref{Eq:coh} as the sum of the edge weights between successive summary sentences.

\begin{equation}
\label{Eq:coh}
Coh(s_i)= w_{G+}(s_i,s_{i+1})+w_{G-}(s_i,s_{i+1}),
\end{equation}

where $w_{G+}$ and $w_{G-}$ represent positive and negative edges, respectively.
Since this formula considers the coherence between adjacent sentences, the produced summary may lack topic coherence compared to human-generated summaries.
However, the outcome of the experiments does not indicate this problem.
The coherence of a summary is the sum of the coherence of all sentences in the summary.

\subsubsection{Redundancy}
The redundancy measure is defined as the combination of a sentence’s embedding similarity with the previously selected sentences.
The overall score of each sentence is defined in Eq.~\ref{Eq:sim}.

\begin{equation}
\label{Eq:sim}
Red(s_i)=\sum_{s\in Summary} sim(s_i,s) 
\end{equation}

\subsubsection{Objective Function}
We propose our objective function to balance defined criteria by having all coverage, coherence, redundancy and the limit size ($B$) in one objective function (Eq.~\ref{Eq:score}).

\begin{equation}
\label{Eq:score}
\begin{gathered}
     maximize: \vspace{10mm} Score(S)\triangleq Cov(S)+\lambda Coh(S)-\phi Red(S)\\  
     s.t.  \hspace{1cm} \sum_{s\in Summary} length(s) <B  
\end{gathered}
\end{equation}

\begin{figure*}
    \centerline{\includegraphics[width=\textwidth]{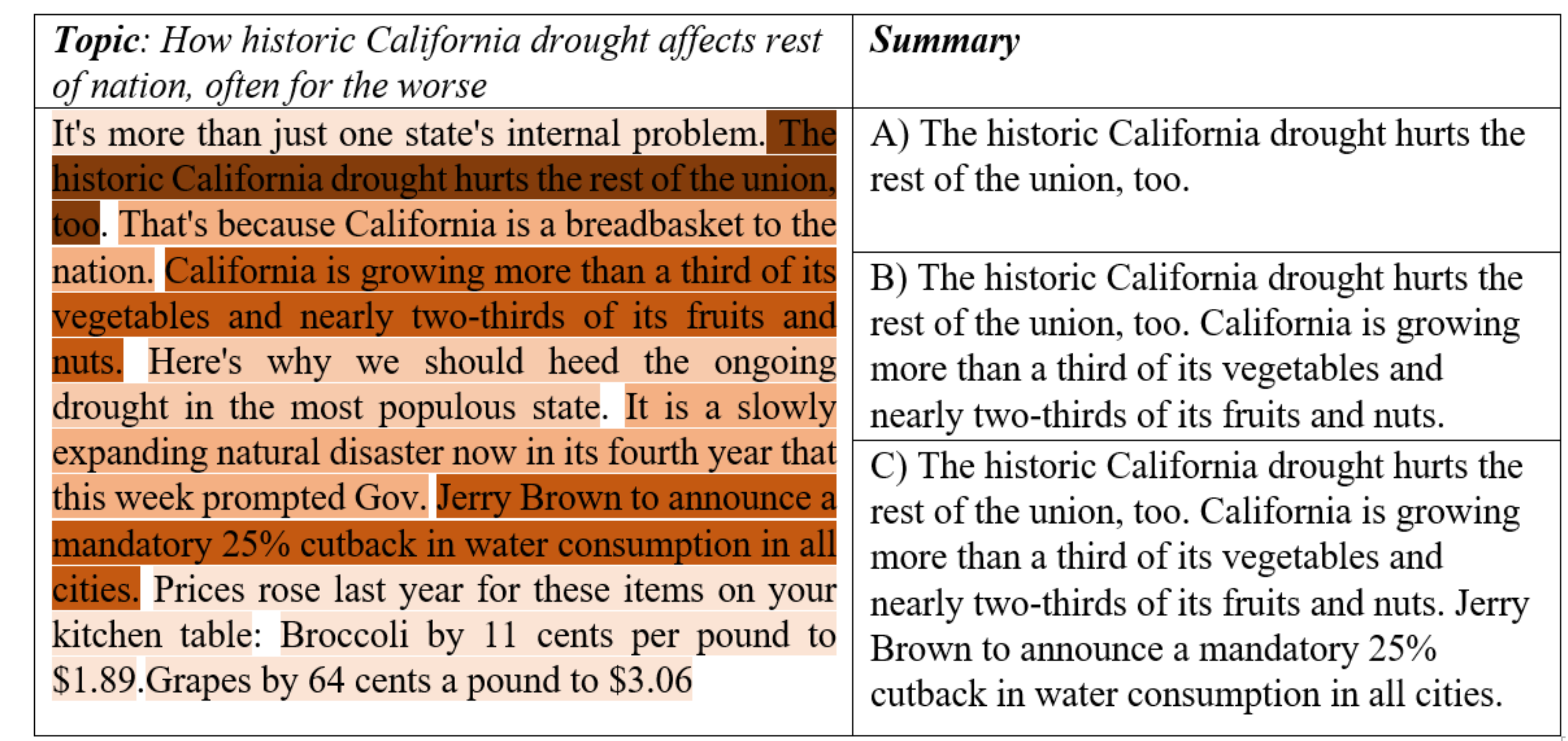}}
    \caption[Caption for LOF]{Visualisation of the summarisation process for a CNN article about the California droughts.~\protect\footnotemark \hspace{1mm} The left box contains the original text, and the right box is the summarisation process (three iterations). To visualise the sentence score, we divide the ranks of an iteration into four portions, each coloured differently. (The darkest colour shows the most important one.) However, it should be noticed that ranks are changed in each iteration.}
    \label{test}
\end{figure*}

To solve this objective function, we need to use a local search to approximate the optimum value.
For this purpose, we used a hill-climbing algorithm with a random start~\cite{stuart2003artificial}.
Adding, removing or replacing a sentence is permitted in each step, and the parameters are trained in the process.
An example of ranking sentences is depicted in Fig.~\ref{test}.

\subsection{Experiments and Evaluation}
This section presents the experimental set-up for assessing the performance of our summarisation model.
The three variants of ROUGE (ROUGE-1, ROUGE-2 and ROUGE-L) were used.
We employed the limited-length ROUGE recall-only evaluation (75 words) for comparison of DUC to avoid bias, and the full-length F1 score to evaluate the CNN/Daily Mail dataset.
We used this measure to compare the produced summary with state-of-the-art approaches and to analyse the effect of local feature weighting in the same approach.

\footnotetext{\url{http://www.cnn.com/2015/04/03/us/california-drought/}}

We evaluated ExDos from various perspectives, including automatic accuracy evaluation of the results, human preference evaluation, the effect of local feature weighting, parameter analysis and efficiency analysis.
The initial number of clusters was set to the best value estimated by the silhouette approach~\cite{rousseeuw1987silhouettes}.

\begin{table}
    \centering
    \caption{ROUGE score (\%) Comparison on DUC2002 Dataset}\label{tab1:duc2002}
    \begin{tabular}{|l|c|c|c|}
        \hline
        Model & ROGUE-1 Score & ROGUE-2 Score  & ROGUE-L Score \\
        \hline
        Lead-3 & 43.6 & 21.0 & 40.2 \\ 
         \hline
        ILP & 45.4 & 21.3 & 40.3 \\ 
         \hline
        TGRAPH~\protect\footnotemark & 48.1 & 24.3 &  N/A \\
         \hline
        URANK & 48.5 & 21.5 &  N/A \\
         \hline
        NN-SE & 47.4 & 23.0 & 43.5\\
         \hline
        \begin{tabular}{@{}c@{}}SummaRuNNer\end{tabular} & 46.6 & 23.1 &43.03 \\
         \hline
        HSSAS & 52.1 & 24.5 & 48.8\\
         \hline
        \textbf{ExDoS} & 52.5 & 24.7 & 48.8\\
        \hline
    \end{tabular}
\end{table}
\footnotetext{\href{http://www.cnn.com/2015/04/03/us/california-drought/}{Rouge-L results for TGRAPH and URANK are not reported.}}

The ROUGE results are illustrated in Table~\ref{tab1:duc2002} and \ref{tab1:cnn}.
According to Table~\ref{tab1:duc2002} (DUC2002 dataset), ExDoS outperforms most state-of-the-art approaches and competes with HSSAS.
Results on the CNN/Daily Mail dataset follow the same trend as DUC2002. Note that the score is generally lower compared to DUC2002.
This is because the gold-standard summaries include paraphrasing. Meanwhile, HSSAS~\cite{al2018hierarchical} is a neural network model that has a hierarchical structured self-attention mechanism to create both sentence and document embedding, and BanditSum~\cite{dong2018banditsum} is a neural network model that considers summarisation as a CB problem.
The latter receives a document and chooses a sequence of sentences to include in the summary, where the policy is to maximise the ROUGE score.
Our model is a simple, efficient model that achieves better results in terms of ROUGE score in most cases, while offering other benefits such as interpretability.
We performed an analysis of variance test to evaluate the significant supremacy of our approach statistically.
Results show that ExDoS outperforms the baselines, including ILP, TGRAPH, URANK and NN-SE, with a significant margin ($p<0.01$), while competing with HSSAS and BanditSum.

\begin{table}
\centering
\caption{ROUGE score (\%) Comparison on CNN/Daily Mail Using F1 Variant of ROUGE}
\label{tab1:cnn}
    \begin{tabular}{|l|c|c|c|}
        \hline
        Model & ROGUE-1 Score & ROGUE-2 Score & ROGUE-L Score \\
        \hline
        Lead-3 & 39.2 & 15.7 & 35.5  \\
         \hline
        NN-SE & 35.4 & 13.3 & 32.6\\
         \hline
        SummaRuNNer & 39.9 & 16.3 & 35.1 \\
         \hline
        HSSAS & 42.3 &  17.8 & 37.6 \\
         \hline
        BanditSum & 41.5 & 18.7 & 37.6 \\
         \hline
        \textbf{ExDoS} & 42.1 & 18.9 & 37.7 \\
        \hline
    \end{tabular}
\end{table}
\vspace{-2mm}

\subsubsection{Feature Importance Evaluation} 
In addition to being modern, ExDoS learns the relevance of features separately for each class, as reported in Table~\ref{tab:featureImportanceyy}.
The reported weights are the average weights in each feature set.
Based on observations, we concluded that in DUC2002, the position-based features play a major role in selecting summaries.
Evidently, the most important features are the frequency-based ones in the ‘summary’ class and the similarity features for those in ‘not summary’.
In the CNN/Daily Mail dataset, the similarity-based feature has a major effect on discriminating both classes, probably due to the paraphrased standard summaries.

\begin{table}
\centering
\caption{Estimation of Features Importance}
\label{tab:featureImportanceyy}
     \begin{tabular}{|c|l|l|l|c|c|c|}
     \hline
     & \multicolumn{1}{c|}{\begin{tabular}{@{}c@{}}Feature\\set\end{tabular}} & \multicolumn{1}{c|}{\begin{tabular}{@{}c@{}}Freq\\based\end{tabular}} & \multicolumn{1}{c|}{\begin{tabular}{@{}c@{}}Word\\based\end{tabular}} & \multicolumn{1}{c|}{\begin{tabular}{@{}c@{}}Similarity\\based\end{tabular} } & \multicolumn{1}{c|}{ \begin{tabular}{@{}c@{}}Position\\based\end{tabular}} &  \multicolumn{1}{c|}{ \begin{tabular}{@{}c@{}}Linguistic\\based\end{tabular}} \\
    \hline
    {\multirow{2}{*}{\rotatebox[origin=c]{90}{\begin{tabular}{@{}c@{}}DUC\\2002\end{tabular}}}} & Summary & 0.39 & 0.06 & 0.35 & 0.51 & 0.22\\
      & Not summary & 0.21 & 0.09 & 0.25 & 0.42 & 0.20\\
      \hline
    {\multirow{2}{*}{\rotatebox[origin=c]{90}{\begin{tabular}{@{}c@{}}CNN\end{tabular}}}} & Summary & 0.33 & 0.04 & 0.46 & 0.31 & 0.29\\
     & Not summary & 0.24 & 0.01 & 0.37 & 0.38 & 0.44\\
    \hline
    \end{tabular}
\end{table}

\subsubsection{Evaluating Effect of Local Feature Weighting}
To evaluate the effect of local feature weighting, we conducted an ablation study to compare results to the global weighting of the same procedure.
The results are reported in Table~\ref{tab:withoutExDoS}.
Evidently, local feature weighting significantly affects the summarisation result in both datasets.

\begin{figure*}[t]
    \centerline{\includegraphics[width=\textwidth]{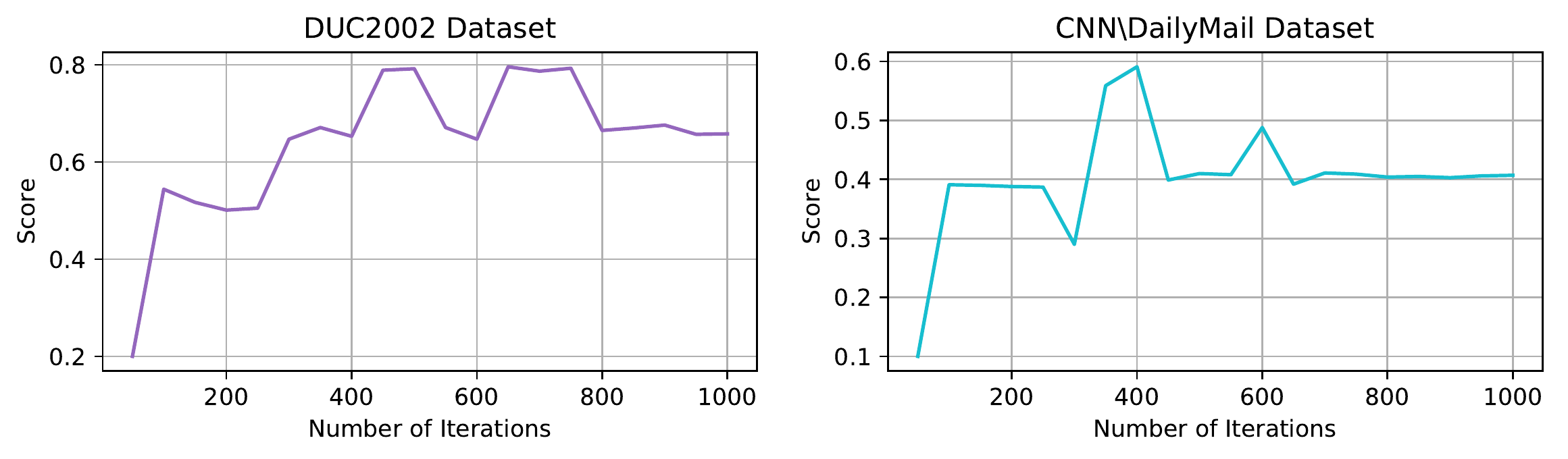}}
    \caption{Both graphs show the number of iterations versus the score in both datasets.}
    \label{1iterations}
\end{figure*}

\begin{table}[t]
    \centering
    \caption{Effects of Dynamic Local Feature Weighting}
    \label{tab:withoutExDoS}
    \begin{tabular}{|l|c|c|c|c|}
    \hline
     & \begin{tabular}{@{}c@{}}DUC2002-\\ROUGE-1\end{tabular} &
    \begin{tabular}{@{}c@{}}DUC2002-\\ROUGE-2\end{tabular} &
    \begin{tabular}{@{}c@{}}CNN/Mail-\\ROUGE-1\end{tabular} &
    \begin{tabular}{@{}c@{}}CNN/Mail-\\ROUGE-2\end{tabular}\\
    \hline
   \begin{tabular}{@{}c@{}}ExDoS +\\ weighting\end{tabular}  & 51.7 & 24.7 & 41.1 & 18.5\\
    \hline
    \begin{tabular}{@{}c@{}}ExDoS -\\ weighting\end{tabular}  & 43.3 & 20.1 & 38.7 & 14.3\\
    \hline
    \end{tabular}
\end{table}

\subsubsection{Efficiency Evaluation}
ExDoS is an efficient approach in terms of its complexity. The computational complexity of ExDoS is determined as $O(K~\times~N_{k}~\times I)$, where $K$ is the number of clusters, $N_{k}$ is the number of samples in the most populated cluster (Max=$N$), and $I$  represents the maximum number of iterations where $I<<N_{K}$.
In Fig.~\ref{1iterations}, the number of iterations versus score value is reported to illustrate the efficiency of ExDoS based on the number of iterations needed to converge the algorithm.

\subsubsection{Parameter Analysis}
As in other parametric models, ExDoS has certain hyper-parameters that need to be tuned.
The learning-rate parameters of weights and centres $(\alpha,\gamma)$ control the speed of convergence in the gradient-descent algorithm. When the learning rate is sufficiently small, the algorithm achieves linear convergence; when it is large, the probability of converging to a suitable local optimum decreases. 
$\beta$  is another key hyper-parameter that regulates the slope of the sigmoid function, where $S_{\beta}(R(s))=1/(1+e^{(\beta(1-R(s)))})$.
For small values of $\beta$, the sigmoid derivative is almost constant.
Conversely, for large values of $\beta$, learning happens when the distance ratio $(R(s))$ is close to $1$.
Two other parameters are $\phi$ and $\lambda$, which control the coherency.

\begin{figure*}
    \centerline{\includegraphics[width=\textwidth]{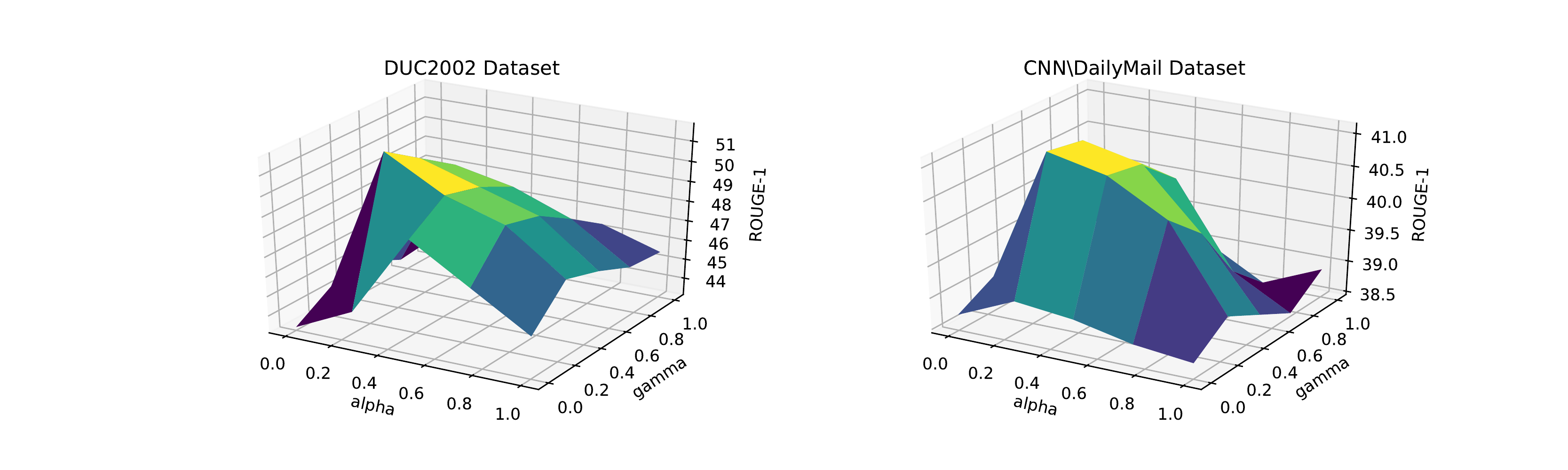}}
    \caption{Both graphs show the learning parameters $(\alpha,\gamma)$ and the corresponding ROUGE-1.}
    \label{ROUGE}
\end{figure*}

\begin{figure*}
    \centerline{\includegraphics[width=\textwidth]{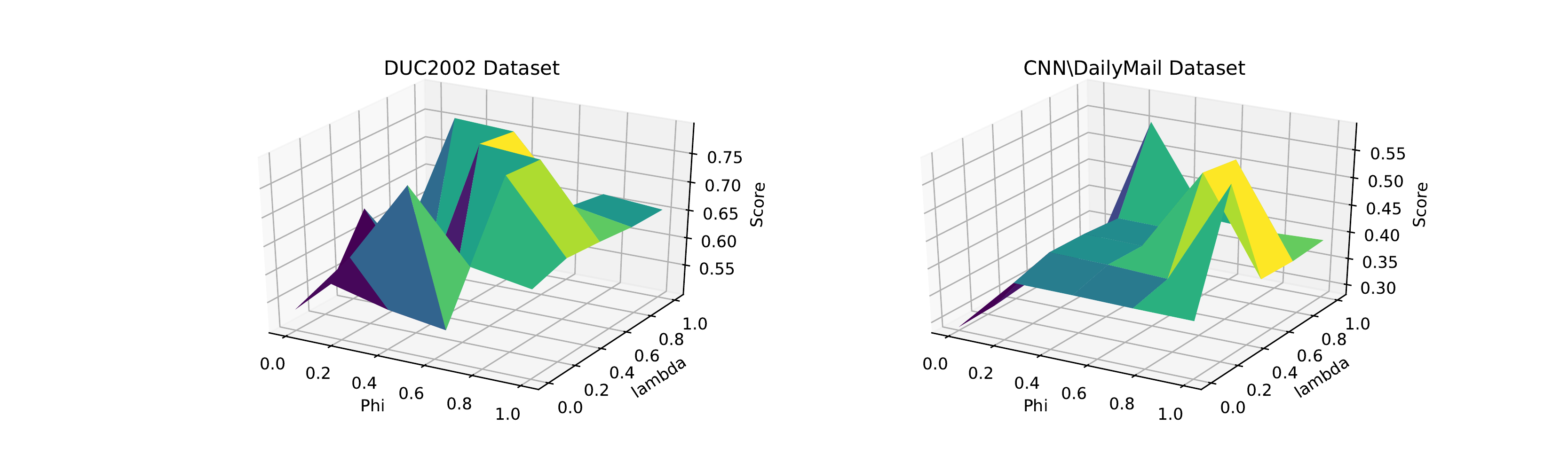}}
    \caption{Both graphs show the learning parameters $(\phi,\lambda)$ and the corresponding score value.}
    \label{Score}
\end{figure*}

To find the best parameters, we tested different combinations of learning rates $(\alpha,\gamma)$. 
These combinations and the corresponding evaluation metric (ROUGE-1) are reported in Fig.~\ref{ROUGE}.
It is noteworthy that the gradient-descent-based learning schemes always converge to a local optimum. When running the algorithm, we empirically observed that it has an effective convergence rate.
The two other parameters $(\phi,\lambda)$ were also tested using different values, reported in Fig.~\ref{Score}.
Since these two parameters control the coherency and redundancy, they do not significantly affect ROUGE.
Therefore, the combination of these variables is reported in terms of score value.

\begin{table}
\centering
\caption{Human Evaluation Result}
\vspace{1mm}
\label{tab:humanEvalutation}
    \begin{tabular}{|c|c|c|c|}
        \hline
        Model & Informativeness & Non-redundancy & Overall\\
        \hline
        Lead-3          & 13\% & 21\%&  20\%\\
        \hline
        SummaRuNNer      & 17\% & 19\%&  16\%\\
        \hline
        HSSAS            & 20\% & 16\%&  21\%\\
        \hline
        BanditSum       & 23\% & 22\%&  18\%\\
        \hline
        \textbf{ExDoS}   & 27\%&  22\%&  25\%\\
        \hline
    \end{tabular}
\end{table}

\subsubsection{Human Evaluation} 
While ROUGE serves as a rough measure of coverage, it only compares the n-gram units.
Therefore, using 20 random sample DUC2002 test documents, we conducted a human experiment to evaluate the model based on other criteria, such as informativeness, redundancy and overall quality.
Twenty-five MTurk participants attended the task, each without any specific prior background.
Participants were presented with a news article and summaries generated using different approaches.
The output of these systems was shown to them, and participants were asked to rank the summaries based on the aforementioned criteria.
Human results reported in Table~\ref{tab:humanEvalutation} represent the voting percentage of participants for each approach (ties were not allowed).
Evidently, ExDoS performed better than most state-of-the-art methods in all measures, but, in terms of redundancy, competes with BanditSum.
Overall, ExDoS achieved significant performance.
This is an interesting result and demonstrates that ExDoS performs well using only clustering information without sophisticated constraint optimisation (ILP, TGRAPH) or the complex architecture of a neural network (HSSAS and BanditSum).

\section{Summary}
This chapter proposed a general-purpose extractive approach for summarising documents. 
We evaluated the proposed model automatically and empirically (human analysis) on the two common benchmark datasets, CNN/Daily Mail and DUC2002.
As shown, the algorithm achieved better results than most state-of-the-art methods in terms of efficiency and performance. 
The human evaluation also proves that the proposed model is proficient in generating instructive and compelling summaries.
Besides, the post-trained weights represent the importance of each feature in discriminating against each class. 
To understand the role of local feature weighting and new feature spaces, we consider the performance of ExDoS through local weighing and without weighting.
Estimating the features’ importance is a fundamental step in summarisation.

%% file: ch_5/InteractiveSummarization.tex
\chapter{Towards Interactive Document Summarisation}
\label{ch_5}
Automatic document summarisation is a long-studied area covering different perspectives. 
It is necessary to articulate the effects and needs of data reduction for analysis, management, commercialisation and personalising purposes.
Summarisation facilitates perceiving and extracting embedded insights that are hidden within data.
However, understanding data is challenging due to the subjectivity aspect of the analysis goal.
Users seek to find only information relevant to a topic and in an organised and coherent structure. 
In the general form, summarisation takes a topic-related set of articles and generates a summary that bears the most crucial information.
The produced summary is, in general, a few selected sentences.

The main drawbacks of existing MDS are as follows:
\begin{itemize}
    \item MDS produces a single, general and flat summary for all users.
    Therefore, summaries are neither interpretable nor personalised, but rather unstructured and, therefore, ill-suited for further analysis.
    \item Existing methods are designed to create short summaries (3–6 sentences) and are incapable of producing more extended outputs.
    Therefore, all details are omitted even if a user is interested in more information.
    \item MDS depends on reference/gold-standard summaries made by humans, which are subjective and costly.
\end{itemize}

Studies have shown that when people are exposed to several documents at once, they rarely make a fully formulated summary~\cite{gupta2010survey}.
Instead, their first attempt is to find a general idea and then gradually go in depth if they find it interesting. 
One study in the literature demonstrates that the most common search strategy among participants is to provide an ‘overview first, filter and selection’, and then discuss the details~\cite{kang2010can}.
That said, each user has different information needs that should be considered when making summaries.
Moreover, they might be interested in exploring different directions based on background knowledge, situation and context, due to personal bias. 
Indeed, these high-level interests will vary over time.
For example, when a researcher wants to read a paper, the first step is to read the title and the abstract. 
The researcher would continue to study the details and methodology only if they are interested.
As an example of context, take the litany of information available on the internet about COVID-19. 
While one might be interested in reading about the symptoms, another might want to research the outbreak locations or perhaps the death toll.
The same applies for researchers investigating summarisation.
A researcher might be eager to know what summarisation is and, thus, focus their interest on different categories of summarisation, such as extractive or abstractive approaches.
Another important issue in this context regards structured summaries, which make further analysis possible.
For example, a user might select the summary length or analyse summaries based on different categories.

In contrast to a generic summary that is unique for all users, this chapter provides user-based hierarchical summaries.
The motivation for this approach is based on how our brain efficiently categorises the perceived information.
The proposed approach help users with general knowledge about a topic to explore a wide range of information.

\begin{figure*}[t]
    \centerline{\includegraphics[width=\textwidth]{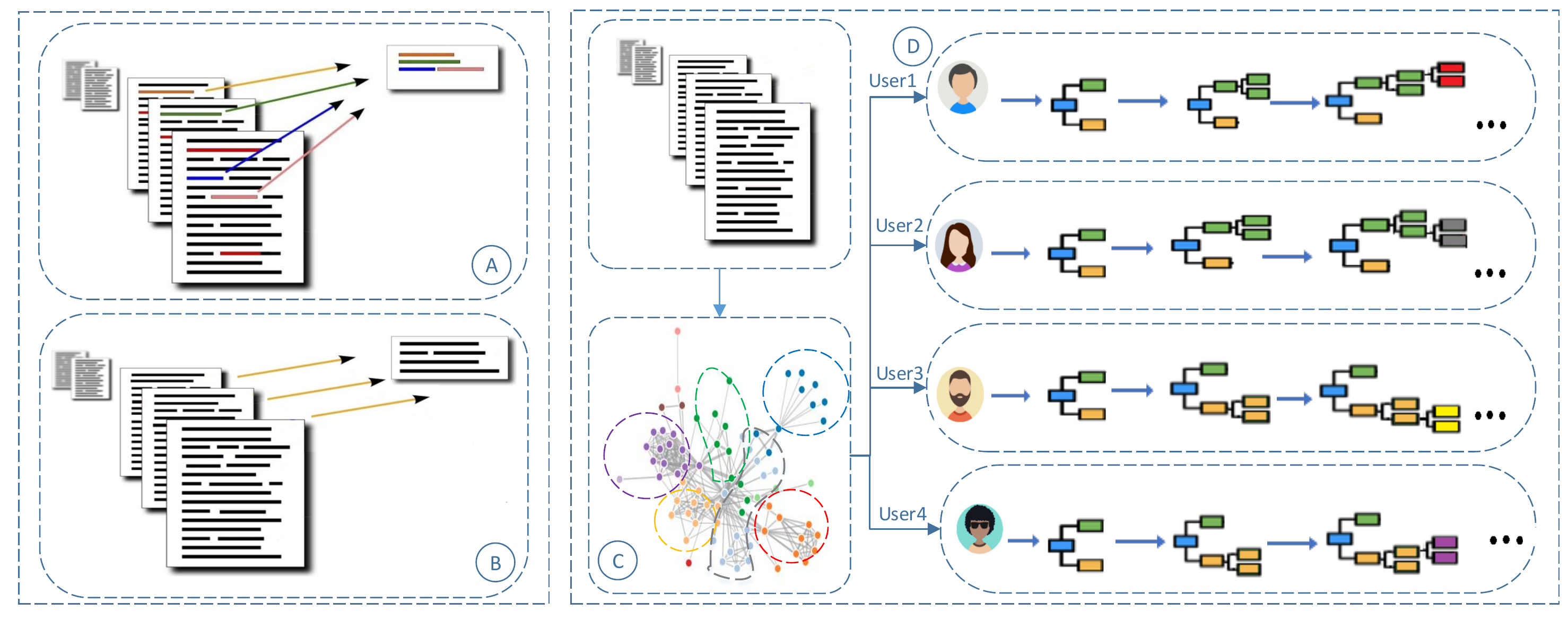}}
    \caption{Traditional summarisation approaches are depicted on the left. (A) Extractive approach (the most informative sentences are selected); (B) Abstractive approach (summaries are generated in the form of new sentences). (C, D) NARS process (a hierarchical summarisation approach).}
    \label{ViewApproaches}
\end{figure*}

We propose a general hierarchical personalised summarisation framework, called NARS, to improve the drawbacks of traditional summarisation methods in various aspects.
We also propose two variants of NARS\textemdash a SNARS and a FNARS.
The goal is to develop intelligent narrative summaries employing the features extracted from users’ engagement. 
To achieve this goal, we propose a hierarchical structure to prevent users from becoming overwhelmed with less important information at first glance, and to facilitate the selection process.
Instead of providing a short and static summary, we present an intelligent and interactive summarisation approach that enables users to navigate through the summary hierarchy and retrieve more in-depth information upon request.
Overall, our approach aims to (i) engage users in the summarisation process and guarantee interactive speeds even for extensive text collections, and (ii) eliminate the need for reference summaries, which is one of the most challenging issues facing the summarisation problem.
A comparison of NARS with traditional state-of-the-art summarisation approaches is shown in Fig.~\ref{ViewApproaches}.

The unique contributions of this chapter are as follows. 
First, we propose a formal definition of multi-aspect hierarchical summarisation.
We focused on users’ desire to better \textit{understand} document summaries rather than obtain only accuracy.
We then introduce NARS, a hierarchical personalised summarisation approach through which users can specify the levels of detail that benefit them in various ways. This includes:
\begin{itemize}
    \item customised summary length (i.e., users determine the length of the summary)
    \item generality v. specificity (The structured output and navigation ability of the approach mean that users learn fast without being overwhelmed by information. This also helps users gradually create a hierarchy by navigating through the summary. The organised output also clearly highlights both minor details and main concepts.)
    \item interaction (i.e., users interact with the summary to better understand one or multiple topics.)
\end{itemize}

Next are reference summary requirements, which see the summaries’ dynamic structure eliminate the need for reference summaries.
This is possible because optimisation of an algorithm that is based on certain gold-standard summaries is not required.

We then propose two models, the SNARS and FNARS.
The output is a well-structured summary that helps users by producing both organised output (i.e., coherent and collated information in one centralised summary) and multifaceted summaries.
This means the models generate concentrated summaries that can answer a user’s query by filtering the hierarchy branches upon request.
Users can also trace the hierarchy based on various criteria in SNARS.

\section{NARS}
\label{NARS}
Previous approaches have used supervised or unsupervised methods that show promising results in terms of accuracy (compared to gold-standard summaries) to tackle the summarisation problem.
Research on MDS mainly ignores the usefulness of the approach for the user.
Instead, the literature mostly focuses on the accuracy of the produced summaries, resulting in inflexible summaries.
Besides, most will optimise their system based on gold-standard summaries generated by human experts, which are costly, subjective and time consuming.
In contrast to previous approaches, we propose a new task for MDS, called NARS, which gathers the related information and collates it into a shorter format.
The proposed approach was called ‘narrative summaries’ since it provides information in a logical order, from the most indicative sentences to more informative sentences.
We also propose a hierarchical structure to prevent users from becoming cognitively overwhelmed when receiving a complete summary at once.
The proposed problem has all the MDS requirements, as well as additional complexities of multifaceted and hierarchical summarisation.

\begin{figure*}[t]
    \centerline{\includegraphics[width=1\textwidth]{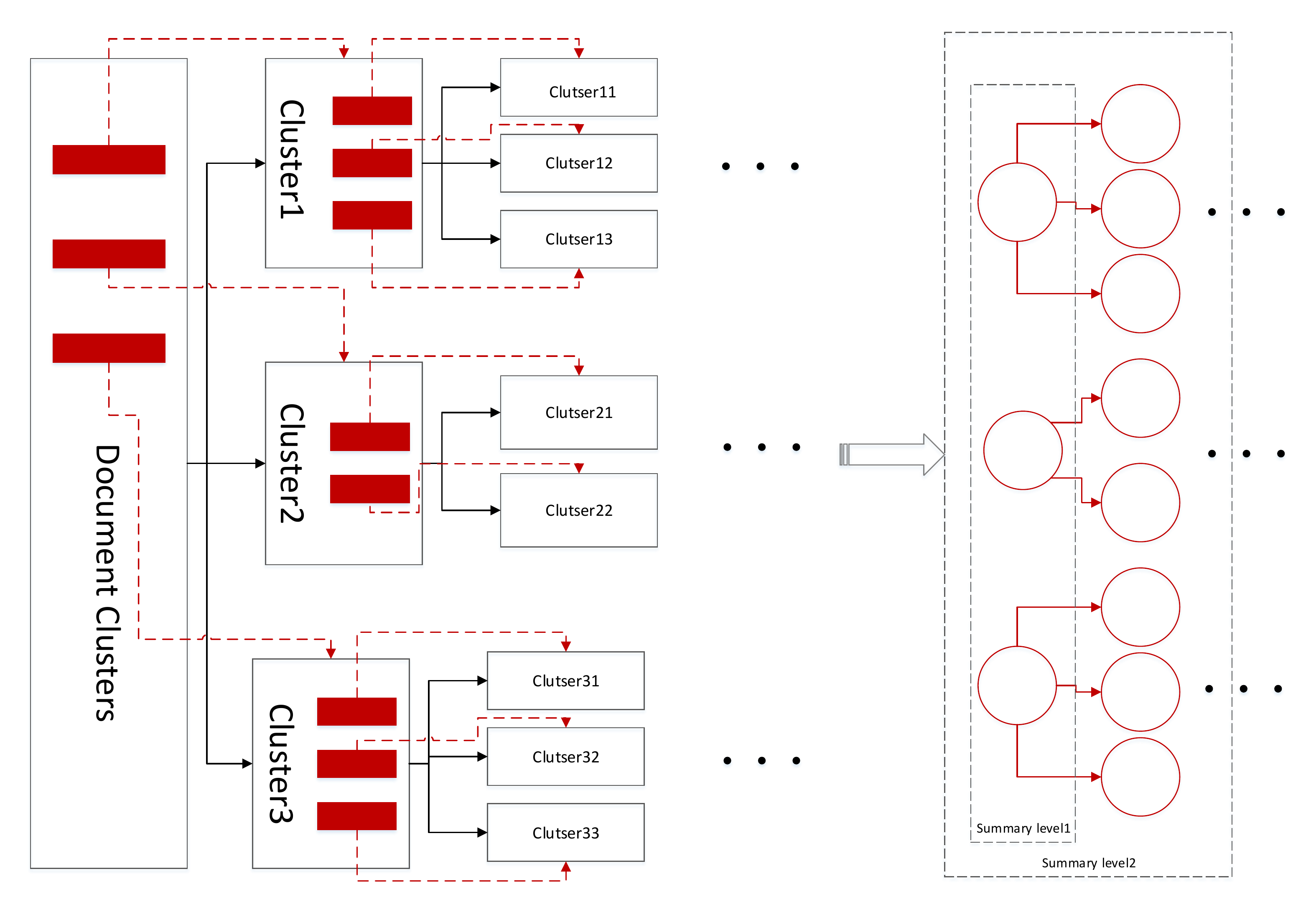}}
    \caption{NARS’s hierarchical structure, where nodes correspond to sentences in SNARS and concepts in FNARS.}
    \label{ch5_diagram}
\end{figure*}

We define a general framework with two main phases involving (i) hierarchical clustering and (ii) hierarchical summarisation over said clustering (Fig.~\ref{ch5_diagram}).
Further proposed are two models based on these phases\textemdash the SNARS and FNARS.
In a semi-structured model, the representation unit is a sentence.
We then define three objective functions to cluster sentences based on topic, time, location and a combination of the three, in such a way that is both coherent and in a logical order.
Since the individual cluster summaries for a given level should be logically distinct, we also propose another objective function in the summarisation phase to maximise coherence and salience while avoiding redundant sentences.
The summary is also required to fit within the given budget size (user parameter).

In a fully structured model, the unit is a concept.
We defined an objective function to cluster related concepts hierarchically in different levels of generalisation and specification.
Then, the summary takes the form of a hierarchical concept map, where each level is a summary.
The proposed solution conveys critical information logically, allowing the user to quickly gain an idea and overview of the content without much reading.

\subsection{Problem Definition}
Given a set of related documents $D=\{D_{1},D_{2}, ... ,D_{n}\}$, the output is a hierarchical summary where the top-level nodes indicate more general information, and the ‘children’ indicate more detailed information related to the ‘parent’ nodes.
The defined structure has the following properties relative to previous approaches:
\begin{itemize}
    \item It is a storytelling process in which the information present at the top level is general and abstract. The summary grows upon the user’s request, particularly if they are interested in answering ‘when’, ‘where’ and ‘who’ questions.
    \item By navigating down the hierarchy, the user perceives the summary effectively by understanding the relationship between parent and child nodes.
    \item Summaries are selected based on measures of fluency, redundancy and being indicative.
\end{itemize}


\subsection{Proposed Framework}
\begin{figure*}[t]
    \centerline{\includegraphics[width=1.1\textwidth]{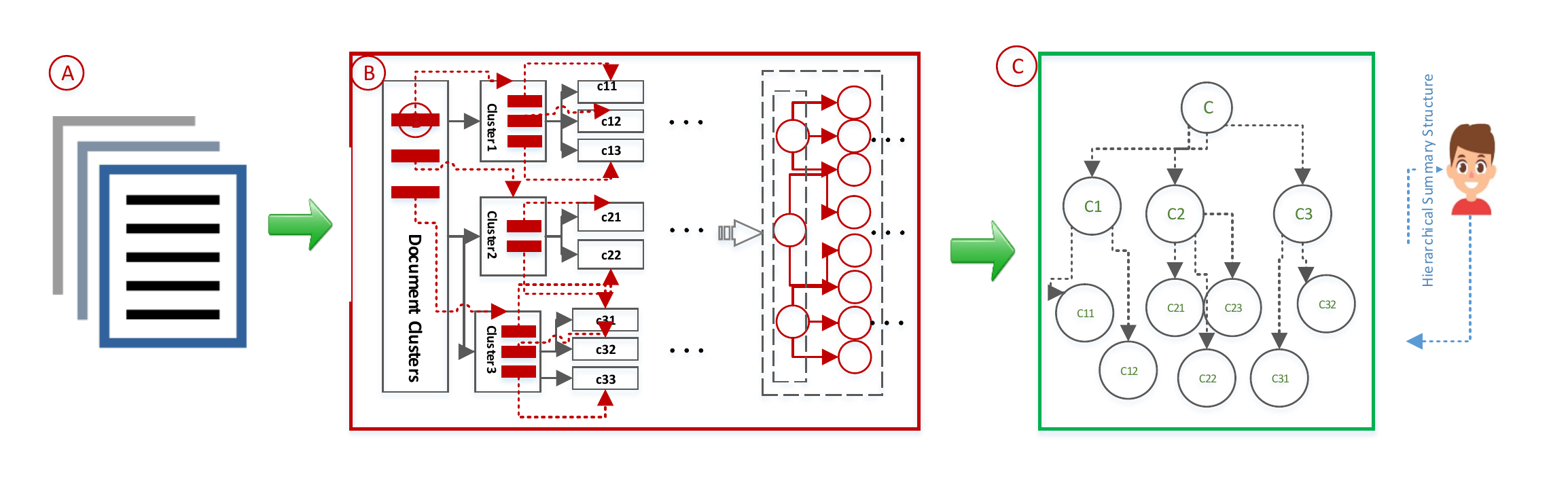}}
    \caption{(A) Documents are pre-processed. (B) Hierarchical clustering is performed. (C) Created summaries are shown to users for interaction, where nodes correspond to sentences in SNARS and concepts in FNARS.}
    \label{system}
\end{figure*}

Overall, NARS is both a semi-structured and fully structured summarisation approach with two main components: (i) hierarchical clustering and (ii) the ability to summarise over the produced clustering to generate hierarchical summaries.
The clustering part creates the boundaries for the summarisation task in the second step.
In a semi-structured model (SNARS), the unit of representation is one sentence.
In this model, time, topic and location are the primary measures in forming hierarchies.
Therefore, this approach is appropriate for storytelling and to answer where, when and what type questions.
In the second model (FNARS), concepts are the representation unit, with a concept map constructed hierarchically for summarising entire contents.
This model is best suited when there are related topics with a redundant concept.
The model's architecture is depicted in Fig.~\ref{system}, and Algorithm~\ref{alg1} demonstrates the general framework.

\section{SNARS}
\label{SNARS}
This section defines the SNARS tasks considered in our approach.
Since sentences are the representative unit in this structure, this model is considered an extractive summarisation approach.

\begin{algorithm}[t]
\begin{algorithmic}[1]
\State \textbf{Input}:  Hierarchical summaries
\State \textbf{Output}: Hierarchical summaries
\State \textbf{Pre-processing}
    \State \quad $\textit{Semi-structured NARS (SNARS): Sentence labelling}$
     \State \quad $\textit{Fully-structured NARS (FNARS): Concept and relation extraction}$
\State\textbf{Hierarchical clustering}
    \State \quad $\textit{SNARS: Multifaceted entropy-based clustering of sentences}$
    \State \quad $\textit{FNARS: Co-reference clustering of concepts and relations}$
\State \textbf{Summarization over hierarchical clusters}
    \State \quad $\textit{SNARS: Optimising based on redundancy, fluency, and being indicative}$
    \State \quad $\textit{FNARS: Optimising based on entropy loss function}$
 \State \textbf{return} Hierarchical summary
 \caption{Narrative summaries}
 \label{alg1}
 \end{algorithmic}
\end{algorithm}

\subsection{Problem Definition} 
Given a set of related documents  $D=\{D_{1},D_{2}, ... ,D_{n}\}$ and the limited size of summaries $b$ as input, the output is a multi-aspect hierarchical textual summary called $S$.

Hierarchical multi-aspect summaries ($S$) have a graph structure, where each node represents a sentence.
The top-level nodes indicate more general information, while the children indicate more specific information related to the parent nodes.
As mentioned, SNARS has two main components\textemdash hierarchical multifaceted clustering and the ability to summarise over the produced clustering.
The clustering component defines the boundaries for the summarisation task in the second step.
Then, for each cluster, a sentence is selected as the representative of that cluster.
Hence, the number of sentences at each level is equivalent to the cluster numbers.
The output of hierarchical clustering serves as the input to the summarisation process.
Thus, functional clustering should have three properties, including:

\begin{itemize}
    \item considerably high enough distance between clusters (inter-cluster distance)
    \item comparatively low distance inside each cluster (intra-cluster distance)
    \item uniformity in terms of size.
\end{itemize}

For reference, we call these conditions the ‘main clustering conditions’.
The designed recursive clustering algorithm automatically chooses the optimal clusters' number at each step.
The challenge here is to find the best measure to cluster documents in a natural and fluent way.
We also clustered sentences based on three measures for representing the summaries inside the hierarchy.
This includes (i) topic-based clustering, (ii) time-aware (temporal) clustering, (iii) location-aware clustering, and (iv) hybrid time/location/topic clustering.
Each of these models is explained in Sec.~\ref{time} to ~\ref{hybrid}, including the summarisation process based on these criteria.

\subsection{Hierarchical Time-aware Clustering}
\label{time}
This model’s objective is to cluster related sentences in different timelines in b groups, meeting the ‘main clustering conditions’. 
We timestamped all sentences individually using SUTime.~\footnote{available as part of the Stanford CoreNLP pipeline: https://nlp.stanford.edu/software/sutime.html}
SUTime is a temporal tagger for recognising and normalising temporal expressions in English text; it is also used to annotate documents with temporal information. 
We used a set of rules to determine if the sentence's timestamps refer to the root verb.
The article date is used in case of no given timestamp.
We also propose a novel recursive clustering algorithm with the goal to maximise the objective function in Eq. 5.1 over all clusters (C), where C=\{$c_1,c_2,..,c_k$\ and $k=b$, using a local search algorithm.

\begin{equation}
\label{j1}
J_1(C)=\alpha_1 \sum_{c\in C} \sum_{s_i,s_j\in c} similarity(s_i,s_j) -\beta_1 \sum_{c\in C} \sum_{B_{t} \in c}  p(B_{t} )\log p(B_{t} ) +\min _{c\in C} size(c)
\end{equation}

We implemented hierarchical clustering top-down at each time, solving for Eq.~\ref{j1}.
The parameters $\alpha_1$ and $\beta_1$ control the effect of time and topic, which are determined using grid search algorithm over a evaluation set.
The first term in Eq.~\ref{j1} is the pairwise similarity of sentences in each cluster, indicating the first condition (intra-cluster distance). For this purpose, we used cosine similarity~\cite{falke2019automatic}.
The second term in Eq.~\ref{j1} is the inter-cluster distance.
After acquiring the timestamp $(t)$, we find the boost-time ($B_{t}$) of articles, defined as the time at which the published article number’s difference in a day ($\lambda(Day)$) and its previous day ($\lambda(PDay)$) is at maximum~\cite{christensen2014hierarchical}. 
Thus, consider Eq.~\ref{ch5_j2}.

\begin{equation}
\label{ch5_j2}
B_{t}= max \{\lambda(Day) - \lambda(PDay)\}, \forall Day,PDay \in {t},
\end{equation}
where $PDay$ is one day before $Day$ for the entire timestamp ($t$) in which we are searching.
We defined the number of boost times as the number of clusters, equivalent to the budget size ($b$).
Accordingly, we assigned all sentences to each boost time, as in Eq.~\ref{ch5_eq3}.

\begin{equation} 
\label{ch5_eq3}
    \begin{split}
        TimeLabel(s)= B_{t}  \hspace{0.5cm} \text{if}\ B_{t-1}<SentenceTime(s)<B_{t}  \\
        \forall s \in S,
    \end{split}
\end{equation}
where $S$ is the set of all the available sentences, and $SentenceTime(s)$ returns the time associated with sentence $s$.
The second term in Eq.~\ref{j1} is to minimise the entropy of time labels in each cluster to meet the inter-cluster distance condition.
Therefore, $p(B_{t})$ is the probability of label$B_{t}$ comparing to other time labels in each cluster.
Finally, the third term in Eq.~\ref{j1} is the uniformity condition to prevent generating clusters with a small number of samples.
The first and second values are between 0 and 1.
The third value is the normalised size of a cluster with minimum size divided by other clusters’ average number size, to avoid bias.

\subsection{Hierarchical Location-aware Clustering}
\label{location}
We considered location as an important feature to create the hierarchy.
We used the location extracted by entity recognition using the Stanford NLP service.
If there was no information for the location, we categorised this as ‘other’.
We next applied the same procedure for the location as for the time and performed top-down hierarchical clustering at each point solving for Eq.~\ref{j2}.
The two parameters $\alpha_2$ and $\beta_2$ control the effect of location and topic, which are calculated using grid search over a development set.
The goal is to find the best organisation for hierarchical clusters by maximising the following objective functions in Eq.~\ref{j2}.

\begin{equation}
\label{j2}
J_2(C)=\alpha_2 \sum_{c\in C} \sum_{s_i,s_j\in c} similarity(s_i,s_j) -\beta_2 \sum_{c\in C} \sum_{B_l\in c}  p(B_l)\log p(B_l) +\min _{c\in C} size(c)
\end{equation}

The definition of all of the terms in Eq.~\ref{j2} is the same as Eq.~\ref{j1}.
We defined location-boost labels ($B_l$) where the number of articles in a location is the maximum, similar to Eq.~\ref{ch5_eq3}.
In practise, labelling location is not as simple as time since various labels lead to many clusters. 
Therefore, we defined two levels of labelling\textemdash country and city\textemdash and used a local search to find the best combination for choosing the boost location to maximise the inter-cluster distance.

\subsection{Hierarchical Topic-based and Hybrid Time–Location Clustering}
\label{hybrid}
The best splitting procedure is one that generates a naturally flowing summary.
In real-world scenarios, sometimes the article’s inflation can be a combination of time, location and/or topic.
As such, we defined another objective function based on only the topic.
The objective function is based on the cosine similarity of sentences (first term in Eq.~\ref{j1}), defined as $J_3$.
Therefore, instead of forcing the summaries to be split based on location or time, we made a combination of the two and selected the maximum value at each level (Eq.~\ref{jf}).

\begin{equation}
\label{jf}
max \{J_1(C),J_2(C),J_3(C)\}
\end{equation}

\subsection{Summarising Over the Hierarchies}
At each level of the hierarchy, we need to define the representative sentences for each cluster using the hierarchy and clusters’ structure.
Therefore, the problem transforms and we must select the best sentence as the representative of each cluster.
We defined three main measures to select representative sentences, including being indicative, redundancy and smoothness (fluency).

The variable ‘being indicative’ (I) is based on two aspects: generality and salience (Eq.~\ref{ch5_ind}).

\begin{equation}
    \label{ch5_ind}
    I(s) = G(s)+S(s)
\end{equation}

The first measure, generality, indicates if the selected sentence is general enough to represent all sentences in a cluster.
To assess the generality of a sentence ($s_k$), we generated a similarity graph for all sentences in the cluster, such as $c_j$.
We then calculated the normalised sum of similarity of all neighbours of a sentence as the value of generality of that sentence (Eq.~\ref{ch5_eq7}).

\begin{equation}
\label{ch5_eq7}
    G(s_k)=\frac{1}{size(c_j)}\Big(\sum_{\substack{i=1 \\ i\neq k}}^{size(c_j)} Similarity(s_i, s_k)\Big), \forall s_i \in S_{c_j}, \forall c_j \in C,
\end{equation}
where the similarity measure is the cosine similarity~\cite{falke2019automatic}, $S_j$ is the set of all sentences in cluster $c_j$, $C$ is the set of all clusters, and $size(c_j)$ returns the number of sentences in a cluster $c_j$.
Another measure that should be considered when selecting the best sentence is its ‘salience’.
We estimated the importance of each feature in making summaries using ExDos and, therefore, trained a log-linear regression based on five critical types of features: these are frequency-based features (TF-IDF, RIDF, gain and word co-occurrence), word-based features (upper-case words and signature words), similarity-based features (Word2Vec and Jaccard measure), sentence-level features (position, length cut-off and size), name entities~\cite{meena2015optimal}, and the ROUGE measure as the final score to predict salience.

Next, the ‘redundancy’ ($R$) measure for a sentence is defined as the combination of a sentence’s embedding similarity with the previously selected sentences.

To maximise the ‘fluency’ ($F$) of the produced summary ($F$), we require two coherence types\textemdash the parent and child coherency, and coherency within each summary level.
For this purpose, we used G-Flow~\cite{christensen_naacl13}, a graph model used for selection and ordering, which balances coverage and coherence.
This model relies on the approximate discourse graph, where each node is a sentence and edges indicate whether a sentence coherently follows one other.
The indicators include coreference, discourse cues, de-verbal nouns, and more. Coherence is defined as the sum of the edge weights between successive summary sentences.
The coherence of a summary is the sum of all sentences’ coherence in a summary.

Next is ‘optimisation’.
Combining these variables unifies our objective function into a single objective function, defined in Eq.~\ref{ch5_opt}.

\begin{equation}
\label{ch5_opt}
     \textbf{maximize}: Score(S)\triangleq I(S)+\gamma F(S)-\phi R(S)
\end{equation}
where $I(S)$ represents the indicative measure of the summary, which summarises the generality and salience of the selected parents.
$R$ is the redundancy measure, $F(S)$ is the fluency measure, and $\gamma$ and $\phi$ are the control parameters for the effect of fluency and redundancy, which are calculated using grid search over a development set.
We approximated a solution using the hill-climbing algorithm with a random start over the space of hierarchical summaries.
According to each step’s dependency on the previous one, we recursively start from the root, finding the best sentence at each level, and then move down towards the leaves. 
At each point, the search algorithm is allowed to add a new sentence, remove a sentence or replace two sentences.
While this search algorithm works well in practise, the branching factor becomes large when the budget and input document size are large.
Thus, we also set the initial summary for random restarts such that the highest indicative sentences are selected first.
The other sentences are subsequently added based on their overall defined score.
When no other sentence can be added to the summary according to budget size, the algorithm is terminated.

\section{FNARS}
\label{FNARS}
A structured summary has the advantage of being used as an overview of a collection of documents.
It also facilities using the summaries for further analysis and processes. 
This model substitutes a hierarchical concept map as a structured presentation style for summarisation.
A concept map is a labelled graph, where nodes present concepts and edges are the relations among nodes~\cite{falke2019automatic}.

\subsection{Problem Definition}
The input is a set of related documents $D=\{D_{1},D_{2}, ... ,D_{n}\}$ and the output is a hierarchical concept map showing the general topics on higher levels and specific ones on lower levels, satisfying a specified size limit, $b$. 
This model also follows the general structure of NARS defined in Sec.~\ref{NARS}, based on the hierarchical clustering of concepts and making summaries over hierarchies.

\subsection{Extracting Concepts and Relations}
Concepts and relations need to be extracted before undertaking any other process. Concept and relation extraction aims to identify spans in a set of documents used as labels for concepts and relations in the concept map.
To this end, we relied on open information extraction~\cite{etzioni2008open}, an approach that extracts binary propositions from the text.
Using the sentence, ‘the Pharmaceutical Benefits Scheme subsidises cancer treatments’, the output of Open IE is  ‘the Pharmaceutical Benefits Scheme $\xrightarrow[]{subsidises}$ cancer treatments’.
‘The Pharmaceutical
Benefits Scheme’ and ‘cancer treatments’ are two concepts, and ‘subsidises’ is the relation between them.
Any extracted concept that does not contain at least one noun token or is longer than five tokens is omitted.

\subsection{Hierarchical Clustering of Concepts}
We describe hierarchical clustering as output that serves as input in the summarisation process.
We employed a recursive clustering algorithm to define the summary structure.
Clusters in the hierarchical structure represent a concept set, making sense to summarise together.
The algorithm first requires finding the semantic similarity of two concepts $c_1$ and $c_2$.
We used both semantic and lexical similarity as the features~\cite{falke2019automatic}, including the normalised Levenshtein distance, the Jaccard coefficient between stemmed content words, semantic similarity based on latent semantic analysis~\cite{deerwester1990indexing}, and WordNet~\cite{miller1990introduction}.
Then, we modelled the similarity as a binary classification using logistic regression such that a positive classification, $y= 1$, means that mentions are coreferent (Eq.~\ref{ch_eq9}).

\begin{equation}
\label{ch_eq9}
    P(y= 1|c_1,c_2,\theta) = Sigmoid (\theta^T\delta(c_1,c_2)),
\end{equation}
where $\delta(c_1,c_2)$ are the features, $\theta$ denotes the learnt parameters, and the sigmoid function is defined using Eq.~\ref{ch_eq10}.

\begin{equation}
\label{ch_eq10}
    S_{\theta}(z)=(\frac{1}{1+{e}^{\theta(1-z)}})
\end{equation}

After evaluating the similarity of two concepts based on different similarity measures, we need to partition similar concepts.
The goal is to hierarchically cluster similar concepts utilising the similarity probability of two concepts.
We used an ILP function to find an optimised partitioning schema that maximally agrees with the pairwise classifications~\cite{barzilay2006aggregation}.
Let $x_{p}\in \{0,1\}$ be a binary value representing the coreference of mentions $c_1,c_2$ being in the same cluster. 
The goal is to optimise the cross-entropy loss function in Eq.~\ref{2max} using a greedy local search to partition similar concepts at each level of the hierarchy.

\begin{equation}
    \sum_{\substack{p \in C^2}} c(p)x_{p}+ (1-c(p)) (1-x_{p})
\label{2max}
\end{equation}

\subsection{Summarising Over Hierarchies}
\begin{figure*}[t]
    \centerline{\includegraphics[width=\textwidth]{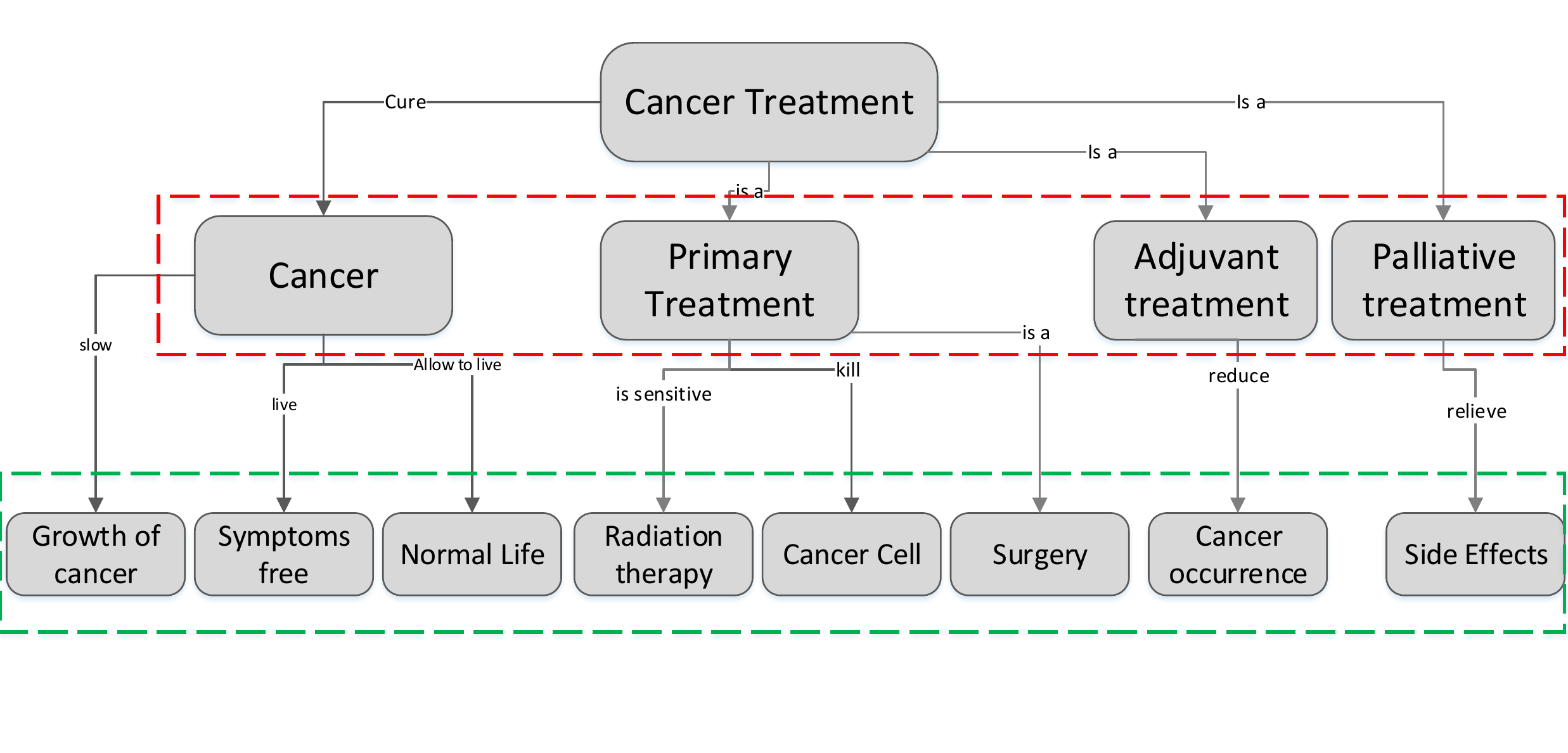}}
    \caption{Example of a hierarchical concept map: two levels of hierarchy are coloured, and each node’s children is shown to users if they select a parent node.}
    \label{concept-map}
\end{figure*}

After partitioning concepts at different hierarchical levels, we can now construct the hierarchical concept map, a graph where $G= (C, R)$ (the nodes representing concepts $C$).
An edge with label $r$ exists for every proposition $(c_1,r,c_2)$ between the nodes.
For each concept, $c_i$, we selected the most frequent and shortest mention as its label to choose the most generic and representative label.
Fig.~\ref{concept-map} is an example of the proposed hierarchical concept map (FNARS).

\section{Experiments and Evaluation}
\label{Eval}
This section presents the experimental set-up for implementing and assessing our summarisation model.
The three variants of ROUGE (ROUGE-1, ROUGE-2 and ROUGE-L) were used.
ROUGE-1 and ROUGE-2 evaluate informativeness, and ROUGE-L (LCS) evaluates fluency.
We used the limited-length ROUGE recall-only evaluation (75 words) to compare DUC2002 to avoid bias, and the full-length F1 score to evaluate the CNN/Daily Mail dataset.

Since our goal is to improve the application of summaries for users, the effect of hierarchical summaries cannot be measured using ROUGE.
Therefore, we also conducted a human experiment to evaluate the model.
For this purpose, we analysed four aspects of user requirements and designed a series of micro-tasks for each experiment, covering information coverage, knowledge extraction and effectiveness (speed), and users’ preference.
We selected not recently published articles to avoid bias in understanding the topics.
Thirty-five MTurk participants attended the task, with no specific prior background in summarisation.
To ensure the human subjects understood the study’s objective, the workers were asked to complete a qualification task, requiring them to write a summary of a news article.
The results that did not have meaning or structure were labelled as spam and removed manually from the data.
For example, in the qualification tasks, we asked users to write a summary explaining the main parts of a document.
Some results could not pass the qualification task. Another example is the short response time, proving that the answers are random or not carefully provided in advance.
We analyse four evaluation aspects including (i) information coverage (how much information the summary covers), (ii) knowledge extraction (how much users can learn from summaries), (iii) effectiveness (the users’ learning speed) and (iv) user preference (the users’ preference compared to other approaches).

\begin{table}[t]
    \centering
    \caption{ROUGE Score (\%) Comparison on DUC2002 Dataset}\label{2tab:duc2002}
    \begin{tabular}{|l|c|c|c|}
        \hline
        Model & ROUGE-1 & ROUGE-2 & ROUGE-L \\
        \hline
        Lead-3 & 43.6 & 21.0 & 40.2 \\ 
         \hline
        ILP & 45.4 & 21.3 & 40.3 \\ 
         \hline
        TGRAPH~\protect\footnotemark & 48.1 & 24.3 &  N/A \\
         \hline
        URANK & 48.5 & 21.5 &  N/A \\
         \hline
        NN-SE & 47.4 & 23.0 & 43.5\\
         \hline
        \begin{tabular}{@{}c@{}}SummaRuNNer\end{tabular} & 46.6 & 23.1 &43.03 \\
         \hline
        HSSAS & 52.1 & 24.5 & 48.8\\
         \hline
          ExDos & 52.5& 18.7 & 37.6 \\
         \hline
        \textbf{SNARS} & 52.9 & 24.8 & 48.9\\
        \hline
         \textbf{FNARS} & 48.3  & 23.8 & 47.3\\
        \hline
    \end{tabular}
\end{table}
\footnotetext{ROUGE-L results for TGRAPH and URANK are not reported.}

\begin{table}[t]
\centering
\caption{Score Comparison on CNN/Daily Mail Using F1 Variant of ROUGE}
\label{2tab:cnn}
    \begin{tabular}{|l|c|c|c|}
        \hline
        Model & ROUGE-1 & ROUGE-2 & ROUGE-L \\
        \hline
        Lead-3 & 39.2 & 15.7 & 35.5  \\
         \hline
        NN-SE & 35.4 & 13.3 & 32.6\\
         \hline
        SummaRuNNer & 39.9 & 16.3 & 35.1 \\
         \hline
        HSSAS & 42.3 &  17.8 & 37.6 \\
         \hline
        BanditSum & 41.5 & 18.7 & 37.6 \\
         \hline
         ExDos & 42.1 & 18.9 & 37.7 \\
         \hline
         Transformer & 40.9 & 18.2 & 37.2 \\
         \hline
         BERTSUM+Transformer & 43.2 & 20.2 & 39.6 \\
         \hline
        \textbf{SNARS} & 42.9 & 19.1 & 37.8\\
        \hline
        \textbf{FNARS} & 40.1 & 18.6 &  37.1\\
        \hline
    \end{tabular}
\end{table}

\subsubsection{Information Coverage}
\label{infoCov}
To automatically evaluate SNARS using traditional state-of-the-art approaches, consider the summaries as the first level of the hierarchy.
Besides, considering all hierarchical levels leads to a high ROUGE value that generates an unfair comparison while exceeding the summary limit size.

To evaluate FNARS, we concatenated all concepts as a textual summary considering the limit size ($b$).
The ROUGE results are illustrated in Table~\ref{2tab:duc2002} and ~\ref{2tab:cnn}.
According to Table~\ref{2tab:duc2002} (DUC2002 dataset), SNARS outperforms most state-of-the-art approaches and competes with HSSAS and ExDos.
Results on the CNN/Daily Mail dataset follow the same trend as DUC2002.
Since we did not consider the hierarchical structure of summaries, this is a promising result that proves even the first level of both NARS techniques can compete with most state-of-the-art approaches.

The proposed approach generates a hierarchical structure that facilitates navigation rather than producing one optimised summary.
Most state-of-the-art approaches will optimise their system based on reference summaries, with ROUGE being the measure with which to evaluate the common n-grams in both reference and generated summaries.
Therefore, we used systems that were optimised based on a reference summary, and compared them with another reference summary with a better ROUGE value than our proposed approach.
Note that the score in the CNN/Daily Mail dataset is generally lower compared to DUC2002.
This is because gold-standard summaries include paraphrasing.
Evidently, FNARS did not present promising results in terms of the ROUGE measure. 
This was expected, as FNARS is an abstract model that does not provide detail.
Its advantage instead rests in its structured format, which facilitates further processing for users. 
Nonetheless, FNARS helps users to understand the topics in a dataset quickly, while simultaneously highlighting the relations between concepts.
The human evaluation of coverage aspect aims to evaluate how information is scattered throughout the hierarchy.
We asked MTurk workers to read an article on a topic and then select the three most important sentences and concepts and then three most critical secondary sentences and concepts.

We combined responses from participants according to the topic and chose the three most basic primary and secondary sentences and concepts.
We manually analysed the presence of these sentences and concepts in the first and second levels of the hierarchy using different summarisation approaches.
We also evaluated the position of sentences and concepts in the hierarchy.
Next evaluated were the SNARS and FNARS based on the recall measure, the percentage of essential sentences, and concepts mentioned at the first and second top levels.
We repeated this experiment for 30 topics and averaged the recall measures.
SNARS retrieved 92.1\% of all important sentences at the first level and 7.3\% at the second level; FNARS retrieved 63\% of all critical concepts at the first level and 26.9\% at the second level.
This experiment illustrates that even the first level of hierarchy works as a general summarisation approach containing the most critical sentences.
Besides, users are allowed to navigate the hierarchy should they desire more detail, as in FNARS.
However, since the representing unit in FNARS is a concept and a concept map, rather than a sentence hierarchy, this puts far less cognitive burden on users in terms of navigation.

\begin{figure*}[t]
    \centerline{\includegraphics[width=1.1\textwidth]{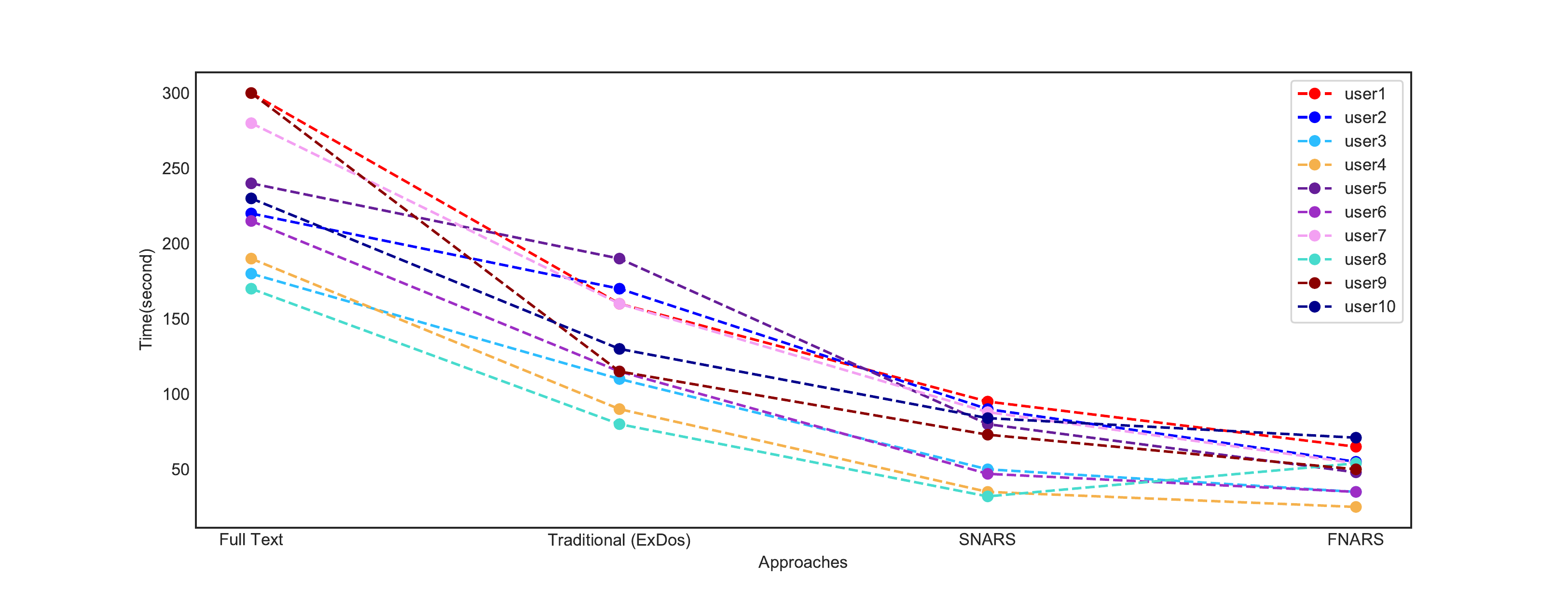}}
    \caption{Time each user took to answer predefined questions when reading different summarisation approaches.}
    \label{speed}
\end{figure*}

\subsubsection{Knowledge Extraction and Speed}
Evaluating the information users gain by allowing them the freedom to seek what they desire based on personal interest is a challenging task.
As such, we designed two experiments that contain an approximation evaluation.

In the first experiment, we gave 10 MTurk workers two minutes to read a summary generated by a traditional competitor approach (ExDos), the full text, and NARS variants.
We chose different topics for each user to avoid bias by learning from other summaries.
Then, participants were asked to write down their understanding.
We could not evaluate the result automatically using the ROUGE measure, as the concepts were all paraphrased by the users.
Therefore, an evaluator manually assessed the understating level and details based on participants’ answers without knowing which approach was used (again, to avoid bias). 
We also prepared a predefined list of answers and scored workers’ responses based on the percentage of concepts covered by their answers.
The results show that users are able to identify general information when reading any of the summaries (the scores were all above 8).
However, the MTurk workers could identify more details when SNARS was shown to them.
Conversely, users who answered by reading the FNARS could mention broader concepts but in less detail.

\begin{figure*}[t]
    \centerline{\includegraphics[width=0.8\textwidth]{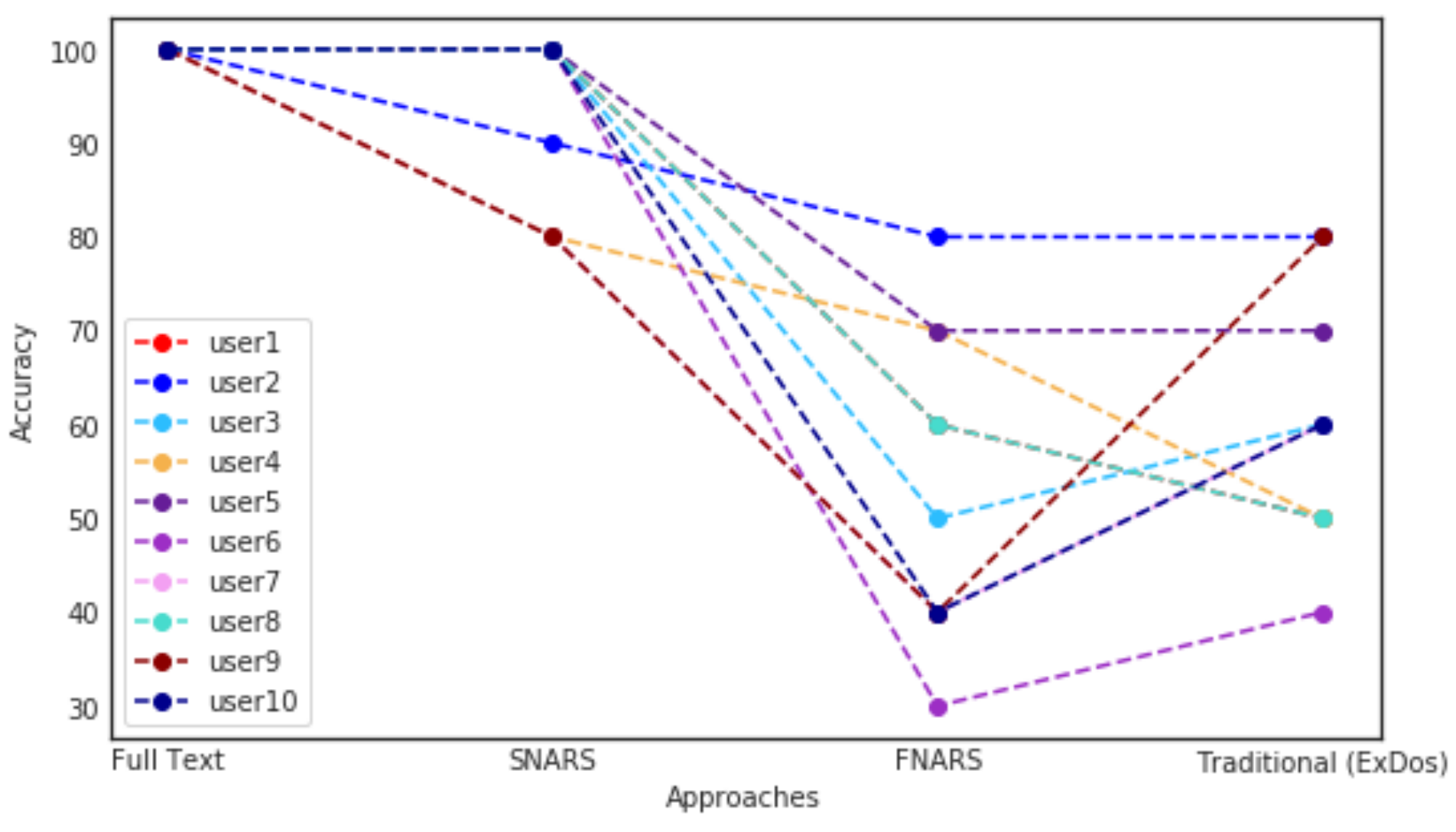}}
    \caption{Accuracy of users’ answers when reading summaries generated using different approaches.}
    \label{correctness}
\end{figure*}

In the second task, 10 workers were given a set of predefined questions and asked to respond to a specific summary.
The questions were selected to cover different aspects (e.g., either more detailed or more general).
Each question came with multiple answers from which to select.
Participants’ speed and the accuracy of their answers were recorded.
Fig.~\ref{speed} and ~\ref{correctness} show that reading a full text takes time but helps users answer questions correctly.
In contrast, traditional approaches take less time but yield less accurate results.
For example, answering using FNARS took the least amount of time, since few details are provided, but discouraged correct answers, as users were forced to respond using inference. 
In contrast, answers based on SNARS demonstrated reasonable results in terms of both time and accuracy.

\begin{figure*}
    \centerline{\includegraphics[width=1.1\textwidth]{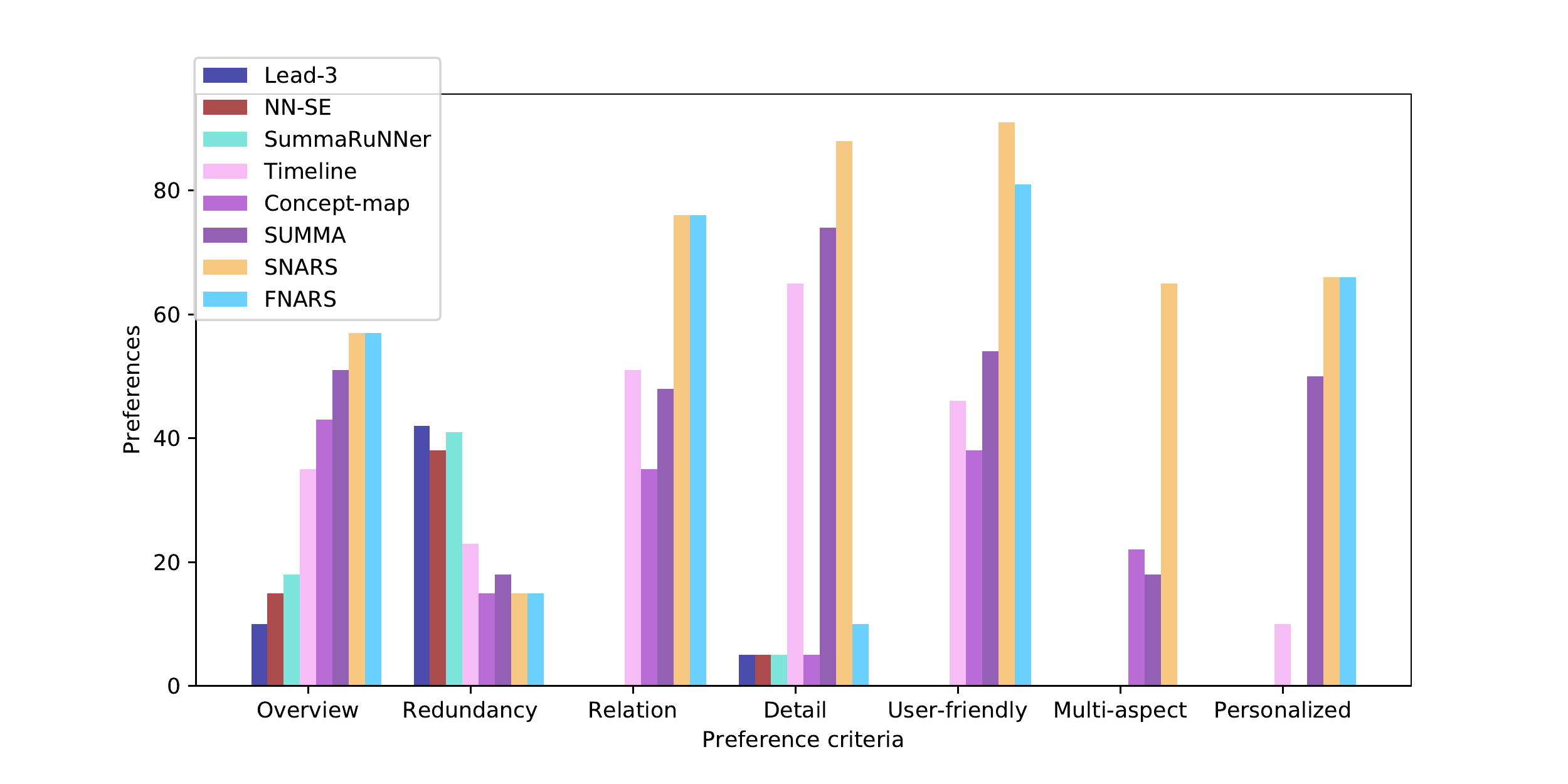}}
    \caption{User preferences based on different aspects mentioned in Table~\ref{tab:humanEvalutation}(tie is allowed).}
    \label{barchart}
\end{figure*}

\begin{table}
\centering
\caption{Comparing Previous Approaches Based on Different Features}
\vspace{1mm}
\label{2tab:humanEvalutation}
\begin{threeparttable}
    \begin{tabular}{|l|c|c|c|c|c|c|c|}
        \hline
        &F1 &F2 &F3 &F4 &F5 &F6 &F7\\
        \hline
        Original Text           &- &- &- &+ &- &- &-\\
        Traditional             &+ &+ &- &- &- &- &-\\
        Timeline                &* &+ &* &* &* &- &-\\
        Document Thread          &* &+ &- &- &- &- &-\\
        Concept Map              &* &+ &+ &- &+ &- &-\\
        SUMMA                   &+ &+ &+ &+ &+ &- &+\\
        NARS                   &+ &+ &+ &+ &+ &+ &+\\
        \hline
    \end{tabular}
    \begin{tablenotes}
      \footnotesize
      \item F1:Overview; F2: Redundancy; F3: Relation; F4: Detail; F5: User friendly; F6: Multifaceted; F7: Personalised. (The notion $+$ shows a fully positive indicator of the property; $-$ is a fully negative indicator of the property; $*$ is a partial indicator of the property).
    \end{tablenotes}
   \end{threeparttable}
\end{table}

\subsubsection{User Preference}
\label{UPref}
To present how the proposed approach could improve the drawbacks of previous approaches, we considered the following seven properties:
\begin{itemize}
    \item Overview\textemdash whether a summary provides an overview of a document.
    \item Redundancy\textemdash if the summary removes any redundant parts.
    \item Relation\textemdash the relation (if any) between factors in documents.
    \item Details\textemdash whether details can be accessed in the produced summary.
    \item User-friendliness\textemdash if the model is easy to use.
    \item Multifaceted\textemdash if the hierarchy is based on one feature, such as time, or multiple features.
    \item Personalisation\textemdash if the summary can be tailored to individual users. 
\end{itemize}

We compared the state-of-the-art categories based on the criteria in Table~\ref{2tab:humanEvalutation}.
These approaches include traditional approaches, timelines, document thread approaches, concept map summarisation, and SUMMA.
The proposed approach—a hierarchical personalised summarisation\textemdash has all the key features of a proper summary (defined in Table~\ref{2tab:humanEvalutation}). 
To evaluate how hierarchical summarisation can help users obtain their required information, we selected news articles covering various topics. 
We then asked 10 MTurk workers to rank their preferred summary among six competitors with various structures, considering the different aspects mentioned in Table~\ref{2tab:humanEvalutation} (with ties permitted). 
The results (percentage) are averaged among 10 workers and are illustrated in Fig.~\ref{barchart}.
Overall, SNARS achieves better results across the criteria in all metrics.

\section{Summary}
\label{Con}
This chapter proposed NARS, a novel personalised interactive hierarchical summarisation approach that enables users to explore whatever information they desire with minimal reading required.
NARS is in contrast to a generic summary and is unique for all users.
Users are able to obtain information based on their individual needs and interests by navigating the personalised hierarchy. 
Two NARS variants were proposed, the SNARS and the FNARS.
FNARS provides a more concise overview of information, while SNARS provides greater detail.
Conversely, FNARS is a fully structured model and can be used for further analysis.
The proposed approaches help users with general knowledge about a topic to explore a wide range of information.
As such, we evaluated our approach using both automatic and human evaluation, considering four aspects: information coverage, knowledge extraction, effectiveness and user preference.
The results prove the use of the proposed approach as a personalised summarisation technique.

%% file: ch_6/PersonalizedSummarization.tex
\chapter{Towards Personalized Document Summarisation}
\label{ch_6}
Making a user-specific summary is a challenging task.
In a personalised approach, the system needs to know about users’ background knowledge or interests. 
When we do not have access to this prior knowledge base, the system requires interaction to acquire feedback for modelling user interests.
Along this line, we provide human-in-the-loop approaches to create a personalised summary that understands individual needs better.
Further, it eliminates the need for reference summaries, which is a challenging issue for summarisation tasks. The goal of this chapter is to answer the following questions:
\begin{itemize}
    \item What structures are required to help users seek their desired information?
    \item How can we use human feedback so that summarisation approaches can adapt to users’ needs?
    \item Can we simulate users’ behaviour to predict their ideal summary?
    \item How can we eliminate the need for reference summaries to reduce the summarisation cost?
    \item Can user preferences over concepts provide personalised summaries that reflect users’ interests with less cognitive load?
    \item Can domain expert knowledge be embedded in the learning process?
    \item How can user preferences and domain expert experiences be combined to automatically generate the desired summaries?
\end{itemize}

We provide three solutions to answer these questions. 
We propose human-in-the-loop summarisation approaches to generate personalised summaries that can capture users’ interests and needs.
First, we propose a novel optimisation algorithm that directly reflects users’ interest in making extractive summaries, called ‘adaptive summaries’.
Second, we propose a preference-based interactive summarisation algorithm that extracts users’ interests and generates user-adapted results.
The proposed method, SumRecom, learns to predict users’ preferences by utilising their feedback and creating a behavioural model following an RL approach.
Predicting users’ desired structured summary is another challenge addressed in this chapter, called ‘summation’.

\section{Adaptive Summaries}
Adaptive summaries are extractive concept-based summarisation models that interact with users to generate content-specific summaries instead of single inflexible summaries. 
The system learns gradually from information provided by users while interacting with them in an iterative loop. 
We allowed even novice users to interactively explore, manipulate and analyse sizeable, unstructured text document collections to integrate their user-specific notion of importance.
Our model also employs an ILP optimisation function to maximise user-desired content selection.
An overview of the proposed approach is illustrated in Fig.~\ref{overview}. 

\begin{figure*}[t]
    \centerline{\includegraphics[width=\textwidth]{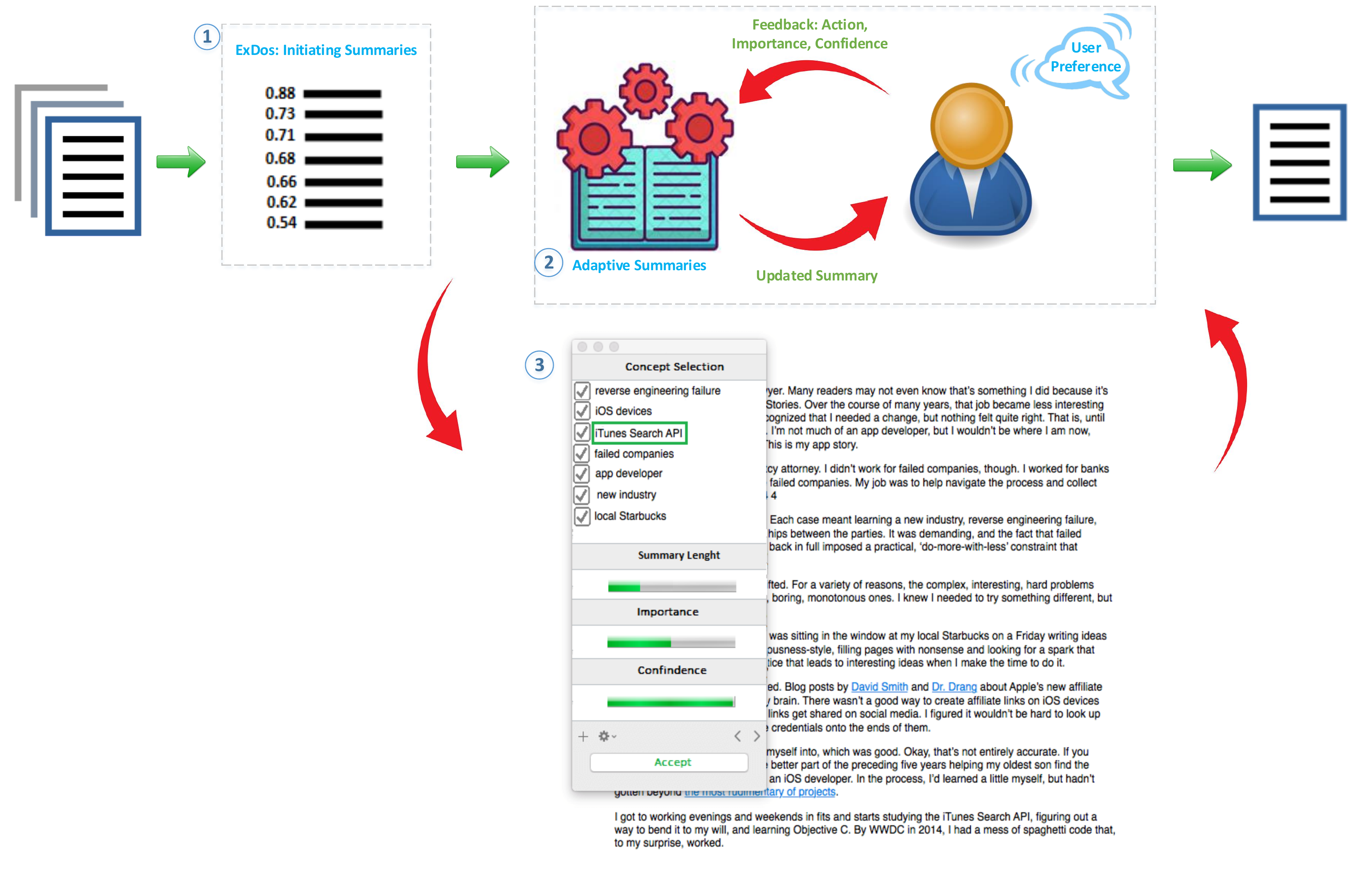}}
    \caption{An overview of the proposed approach (Adaptive Summaries). 1) Summaries are initiated with ExDos. 2) Users integrate their preferences in making summaries by giving feedback in an iterative loop. 3) An example of user interaction.}
    \label{overview}
\end{figure*}

The input is a set of documents, and the output is a human-readable summary consisting of a group of sentences set according to the user’s preference.
As mentioned, users are allowed to provide their feedback in an iterative loop and can also choose which concepts they deem important and that degree of importance. 
Moreover, they can define the confidence level in their choices and can even select which concepts not to include in the final summary.
The proposed approach learns to select sentences that maximise the summary score according to user feedback.
To guarantee interactive speed and keep users engaged, we propose a heuristic approach for selecting users’ queries.
Sec.~\ref{ch6_dproblem} formally defines the summarisation tasks specified in this model.

\subsection{Problem Definition}
\label{ch6_dproblem}
The input is a set of documents $D=\{D_{1},D_{2}, ... ,D_{N}\}$, and each document consists of a sequence of sentences $S=[s_1,s_2,$$...$$,s_n]$. 
Each sentence $s_i$ is a set of concepts $\{c_1,c_2, ..,c_k\}$, where a concept can be a word (unigram) or a sequence of words.
This framework optimises the summarisation outcome for a specific user.
Therefore, the user interacts with the system, and their feedback is used to make the summaries.
Feedback comes in three forms, characterised as:
\begin{itemize}
    \item action, $A$, or accepting (A=1) or rejecting (A=-1) the value of extracted concepts
    \item concept weight, $W$, or corresponding to a concept’s importance according to the user’s opinion
    \item confidence level, $conf$, representing one’s confidence in choosing each action.
\end{itemize}
The output is a set of sentences in $S$  according to the budget limit ($b$) defined by the user.

\subsection{Methodology}
The goal of adaptive summaries is to incorporate user preferences to generate a summary.
Therefore, a continuous objective function is defined for analytically optimising the user preference.
In the first iteration, a summary is generated using ExDos, which ranks sentences based on a general notion of importance using dynamic local feature weighting.
It also demonstrates sentences in groups based on their similarity to help users select content. The user can then choose an action $A$,, denoting a concept, where the values can either be accepted (A=1) or rejected (A=–1).
Next, for each concept, users can define a weight,$W$, corresponding to a concept’s importance based on personal opinion.
The user then defines the level of confidence, $conf$, for the chosen action. 
When the action is accepted, this weight represents the importance of the concept, and when the action is rejected, the weights represent the value of no relation.
The logic behind this is that not all concepts have an equal level of importance.
For example, when a user searches for the specific symptoms of an illness, a headache may not be as important as sneezing from one perspective to another.
Conversely, a fever may not be as unrelated as acne. 
The overall objective function, which is an ILP, is defined in Eq.~\ref{ch6_eq1}.

\begin{equation}
\label{ch6_eq1}
\begin{gathered}
    maximize \sum_{s_i\in D} \sum_{c_j\in s_i} A \times conf(A) \times W_{c_j}\\  
     s.t.  \hspace{1cm} \sum_{s\in Summary} length(s) <b,
\end{gathered}
\end{equation}
where $A$ is the action, $c_j$ is the concept in a given sentence ($s_i$), D denotes the source documents,
$W_{c_j}$ is the corresponding user preference weight for the concept $c_j$, and $b$ is the summary length given by the user.
The objective function in Eq.~\ref{ch6_eq1} maximises the occurrence of concepts with maximum weights and maximum confidence level.
The following is an in-depth description of the proposed approach:
\begin{itemize}
    \item To accelerate the process of making a summary, in the first iteration, the sentences are ranked by ExDos. Then, weights are updated based on user feedback. To prevent users from being overwhelmed, similar sentences (with common concepts) are grouped and shown to the user simultaneously.
    \item If the weight of a concept is updated in an iteration, the weight is also updated for every occurrence of that concept.
    \item If the user rejects a sentence ($A_{s_i}=-1$), then the weight of the sentence is set to 0 ($W_{s_i}=0$).
    However, the system does not update the weights of concepts included in the sentence, as there may be different reasons for its rejection, such as redundancy or lack of importance.
    \item A concept is only selected if it is present in at least one of the selected sentences.
    \item The number of sentences is a user parameter defined in each iteration, and the confidence in feedback is set to 1 by default.
    \item If there are no more concepts to query, the process is terminated.
\end{itemize}
The pseudo code of the proposed algorithm is reported in Algorithm~\ref{alg1_ch6}.

\begin{algorithm}[t]
\begin{algorithmic}[1]
\caption{Adaptive Summaries}\label{alg1_ch6}
\State \textbf{Input}: Document cluster
\State \textbf{Output}: Optimal summary generated by user ($S$)
\State $Ranked Sentences\gets ExDos(D)$
    \State While user is not satisfied
        \State \quad $Concepts \gets ExtractNewConcepts (Ranked Sentences)$
        \State \quad  if $Concepts \neq\emptyset$
            \State \quad \quad {Ask user for action (A), importance(W), and confidence (Conf)}
            \State  \quad \quad {Select sentences to maximize Eq.~\ref{ch6_eq1}}
\State \textbf{return} Summary(S) 
\end{algorithmic}
\end{algorithm}

\subsection{Experiment}
This section presents the experimental set-up for implementing and assessing the summarisation model.
In traditional approaches, to evaluate a summarisation system, the mean ROUGE scores across clusters are averaged.
Adaptive summaries are evaluated using mean ROUGE scores across clusters per standard summary.

It is worth mentioning that this approach aims at facilitating the creation of summaries for individual users, not for improving the general accuracy of summaries. 
Since this approach is interactive, it requires humans to interact with the system for a user study-based evaluation.
However, collecting data for different settings from different humans is too expensive. 
Thus, we simulated users’ behaviour by generating feedback using two variations of the proposed approach.

\begin{table}[b!]
\centering
\caption{ROUGE score comparison on CNN/DailyMail using F1 variant of ROUGE.}
\label{tab:cnn}
    \begin{tabular}{|l|c|c|c|}
        \hline
        Model & Rouge-1 Score & Rouge-2 Score & Rouge-L Score \\
        \hline
        LEAD-3 & 39.2 & 15.7 & 35.5  \\
         \hline
        NN-SE & 35.4 & 13.3 & 32.6\\
         \hline
        SummaRuNNer & 39.9 & 16.3 & 35.1 \\
         \hline
        HSSAS & 42.3 &  17.8 & 37.6 \\
         \hline
        BANDITSUM & 41.5 & 18.7 & 37.6 \\
         \hline
        \textbf{Adaptive dictionary} & 42.9 & 20.1  & 38.2 \\
        \hline
        \textbf{Adaptive reference}  &  41.4 & 19.7 &  32.1\\
        \hline
    \end{tabular}
\end{table}

\begin{table}[b!]
    \centering
    \caption{ROUGE score (\%) comparison on DUC-2002 dataset.}\label{tab:duc2002}
    \begin{tabular}{|l|c|c|c|}
        \hline
             Model & Rouge-1 Score & Rouge-2 Score  & Rouge-L Score \\
        \hline
              LEAD-3 & 43.6 & 21.0 &N/A \\ 
         \hline
             NN-SE & 47.4 & 23.0 & N/A\\
         \hline
            \begin{tabular}{@{}c@{}}SummaRuNNer\end{tabular} & 46.6 & 23.1 & N/A \\
         \hline
            HSSAS & 52.1 & 24.5 & N/A\\
         \hline
            Upper Bound & 47.4 & 21.6 &  18.7\\
         \hline
            Avinesh-Al & 44.8 & 18.8 &  16.8\\
         \hline
            Avinesh-Joint & 44.4 & 18.2 &  16.5\\
         \hline
            \textbf{Adaptive dictionary} & 50.4 & 22.1 & 18.4 \\
         \hline
            \textbf{Adaptive reference}  & 46.5  & 20.1 & 18.8\\
        \hline
    \end{tabular}
\end{table}
\begin{figure*}[h]
    \centerline{\includegraphics[width=1.2\textwidth]{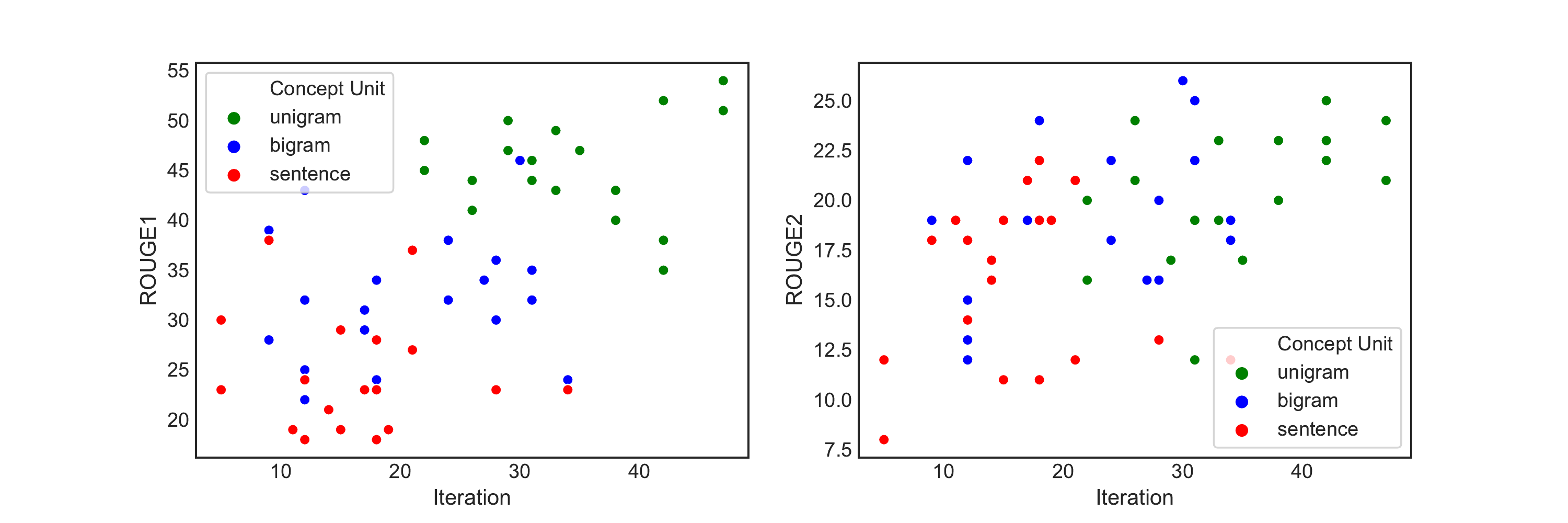}}
    \caption{Left graph shows the ROUGE-1 based on iteration number, and the right graph shows the ROUGE-2 based on iteration number. The green samples represent the permitted concept unit in unigrams, blue denotes bigrams and red represents sentences.}
    \label{ada_conceptunit}
\end{figure*}

\begin{figure*}
    \centerline{\includegraphics[width=1.2\textwidth]{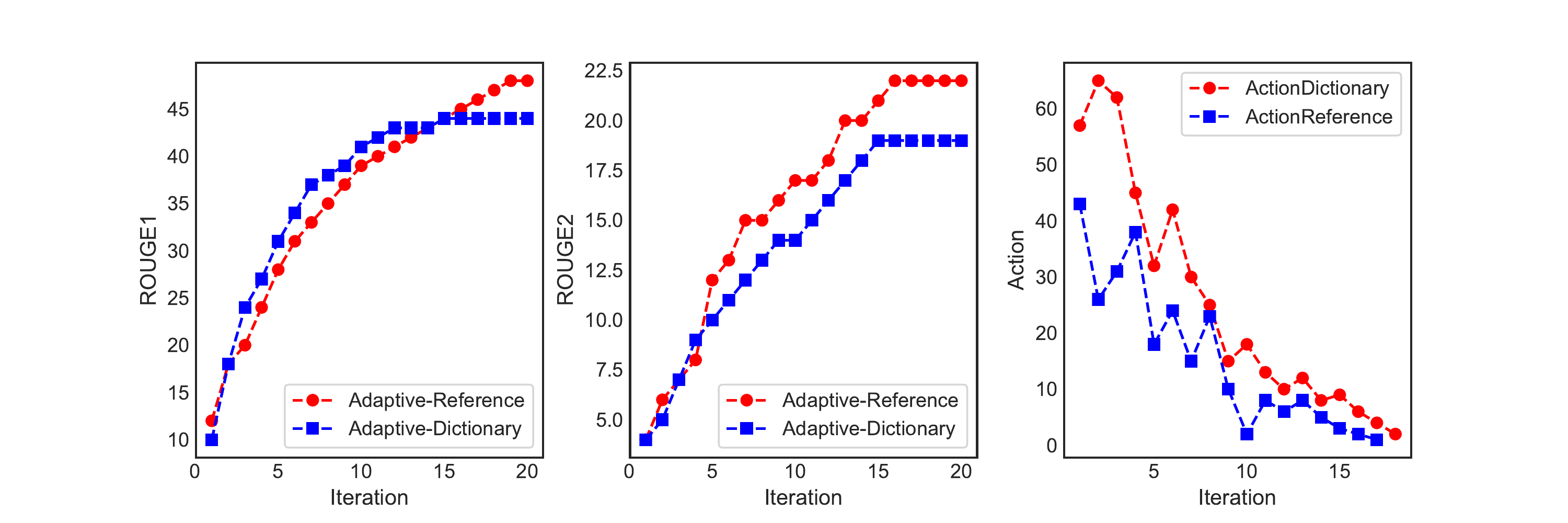}}
    \caption{(A) Number of iterations and ROUGE-1 for DUC2002. (B) Number of iterations versus ROUGE-2 values. (C) Number of actions versus iterations for DUC2002.}
    \label{iterations}
\end{figure*}

In the first approach (AdaptiveDictionary), we defined a dictionary for 10 clusters of topics including the essential concepts and weights, with defined actions for each concept.
In the second one (AdaptiveReference), the reference summaries are considered the users’ feedback. 
The concepts are essential if they are presented in the reference summary.
Therefore, we assigned the maximum weight for the presented concepts. We compared our approach with both traditional and personalised approaches, the results of which are reported in Table~\ref{tab:cnn}  and ~\ref{tab:duc2002} for both datasets.
As discovered, the proposed approach nearly surpasses the competitors for both datasets.

The ROUGE analysis with real users does not show any pattern of increasing or decreasing.
However, this is expected since the approach aims to optimise the ouput summary for individual users and is not a gold-standard summary.

To compare each concept’s unit’s effect, we evaluated our approach based on three unit measures: unigrams, bigrams and sentences.
Although our model reaches the upper bound when using unigram-based feedback, it requires significantly more iterations and feedback to converge, as shown in Fig.~\ref{ada_conceptunit}.
We further analysed the speed (iterations) and the accuracy (ROUGE-1 and ROUGE-2) for different concepts’ units for DUC2002. The CNN/Daily Mail dataset follows the same trend.
The graphs show that when the permitted selection unit is a unigram, the ROUGE-1 score is higher.
However, it takes more iterations to converge.
For ROUGE-2, both the bigram and unigram have higher scores; however, the former converges sooner.

Another experiment considered the ROUGE scores versus the number of iterations.
Fig.~\ref{iterations}(A) and ~\ref{iterations}(B) show the results for the DUC2002 dataset for two versions of adaptive summary, using a dictionary and a reference summary as feedback, respectively.
Fig.~\ref{iterations}(B) represents the models evaluated based on the action number (A) taken by the users to converge to the upper bound within 10 iterations. 
While adaptive summaries incorporate user feedback to generate summaries, they require interaction.
Therefore, there is a need to predict users’ behaviours.

\section{SumRecom}
Making a user-specific summary involves (i) acquiring relevant information about a user, (ii) aggregating and integrating that information into a user model, and (iii) generating the personalised summary.
As such, we incorporated human-in-the-loop systems and created a personalised summary that can predict users’ needs based on samples of their feedback.
The rationale behind this was to evaluate the quality of a summary based on a domain expert’s knowledge given users’ feedback, and to keep humans in the loop through interaction.
To reduce users’ cognitive burden when providing feedback, we considered two aspects. 
First, feedback should be given based on preference, and second, users’ preferences should be formed as concepts and not complete summaries.
Moreover, users are allowed to define the detailed properties of the produced summaries, thus, helping to reduce the search space by leveraging their feedback.

The proposed method, SumRecom, is a preference-based interactive summarisation approach that extracts users’ interests to generate user-adapted results.
SumRecom predicts users’ desired summaries by incrementally adapting the underlying model through interaction. 
The proposed approach has two steps involving (i) the user preference extractor and (ii) the summariser.
Our model employs active learning and preference learning to extract users’ preference in selecting content.
SumRecom also utilises ILP to maximise user-desired content selection based on the given feedback.
It then proposes an IRL algorithm using domain expert knowledge for evaluating the quality of summaries based on the given feedback.
The learnt reward function is used to learn the optimal policy to produce the desired summary using RL. 
A general overview of the algorithm is depicted in Fig.~\ref{overview}.
Before explaining the proposed method, we observe preference-based and reinforcement-based approaches used for summarisation.

\begin{figure*}
    \centerline{\includegraphics[width=\textwidth]{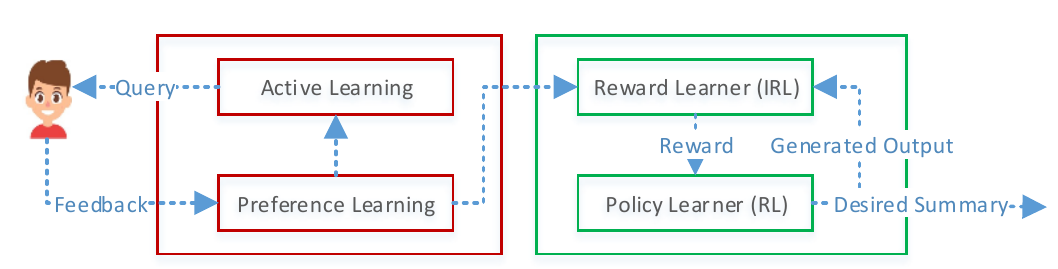}}
    \caption{Overview of the SumRecom approach. Active and preference-based learning are used to extract users’ preferences. The learnt preference ranked function is used to produce the desired summary using IRL for learning the reward. An RL algorithm is proposed for learning the optimal policy.}
    \label{overview}
\end{figure*}

\subsection{Preference-based and Reinforcement-based Approaches}
There is increasing research interest on using preference-based feedback and RL algorithms in summarisation.
For example, one approach is to learn a sentence ranker function based on human preferences on sentence pairs~\cite{zopf2018estimating}.
The ranker function is then used to assess the summaries' quality by calculating the number of high-ranked sentences inserted in the summary.
RL-based approaches are employed for extractive and abstractive summarisation recently~\cite{ryang2012framework,pasunuru2018multi,paulus2018deep}.

Most existing RL-based document summarisation systems employ heuristic methods to determine the reward function and, therefore, are not dependant on reference summaries~\cite{ryang2012framework,rioux2014fear}.
Other approaches use different ROUGE measure variants as the reward function and, therefore, require reference summaries for estimating the reward value~\cite{pasunuru2018multi,paulus2018deep,kryscinski2018improving}.
However, neither ROUGE nor the heuristics-based methods can determine users’ preferences as the reward~\cite{chaganty2018price}. 
Therefore, using these imprecise reward models can critically deceive the RL-based summariser.
The reward quality is the biggest bottleneck for RL-based summarisation~\cite{gao2019reward}.

\begin{figure*}[t]
    \includegraphics[width=\textwidth]{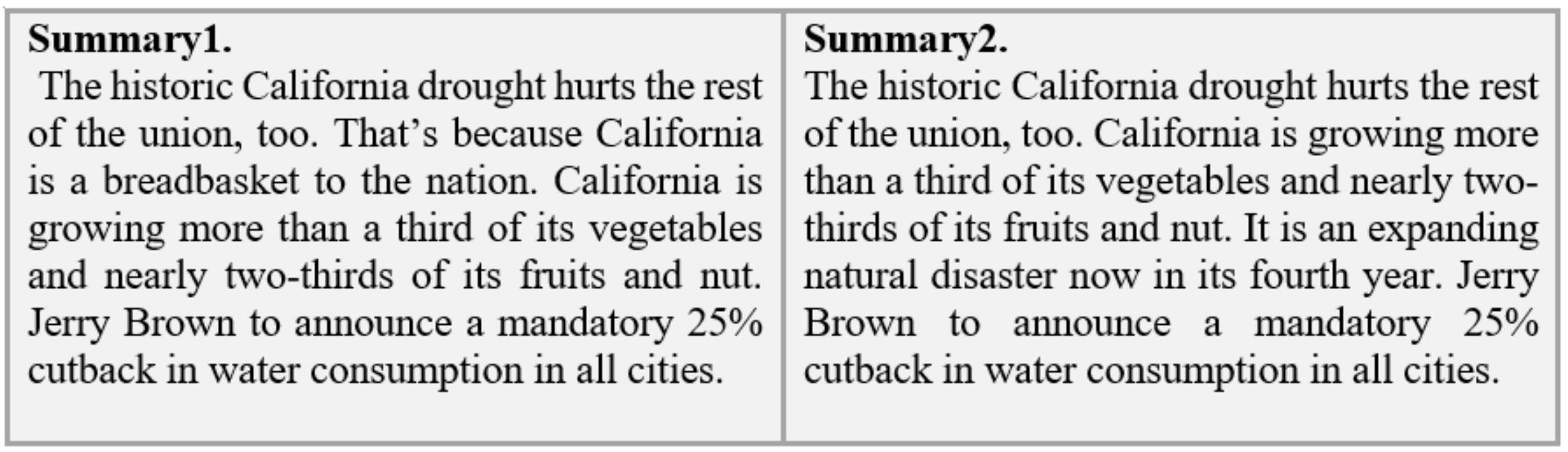}
    \caption{Comparing two summaries that put significant cognitive burden on users.}
    \label{comparing}
\end{figure*}

Both preference-based and RL algorithms have been used in summarisation simultaneously. 
The first approach is SPPI~\cite{sokolov2016stochastic,kreutzer2017bandit}, a policy-gradient RL algorithm that rewards are made based on the given preference-based feedback.
One drawback with SPPI is that it suffers heavily from the high sample complexity problem.
However, the complexity of SPPI is a severe problem.
Another recent preference-based RL approach is APRIL~\cite{gao2018april}, which has two stages.
First, the users rank candidate summaries, and then a neural RL agent is used to find the optimal summary.
However, preferring one summary to another in both approaches puts considerable burden on users.
It is worth mentioning that summarisation aims to provide users with a summary that reduces the need to read multiple documents.
Although asking users to favour a summary to another in multiple rounds among a summary space that includes all randomly possible combinations of sentences puts additional cognitive load on them, this still outweighs the demand of reading. 
Fig.~\ref{comparing} represents an example of this comparison, where the challenge to read summaries increases as the summary length increases.

\subsection{Methodology}
\label{proposed}
\begin{figure*}[t]
    \centerline{\includegraphics[width=\textwidth]{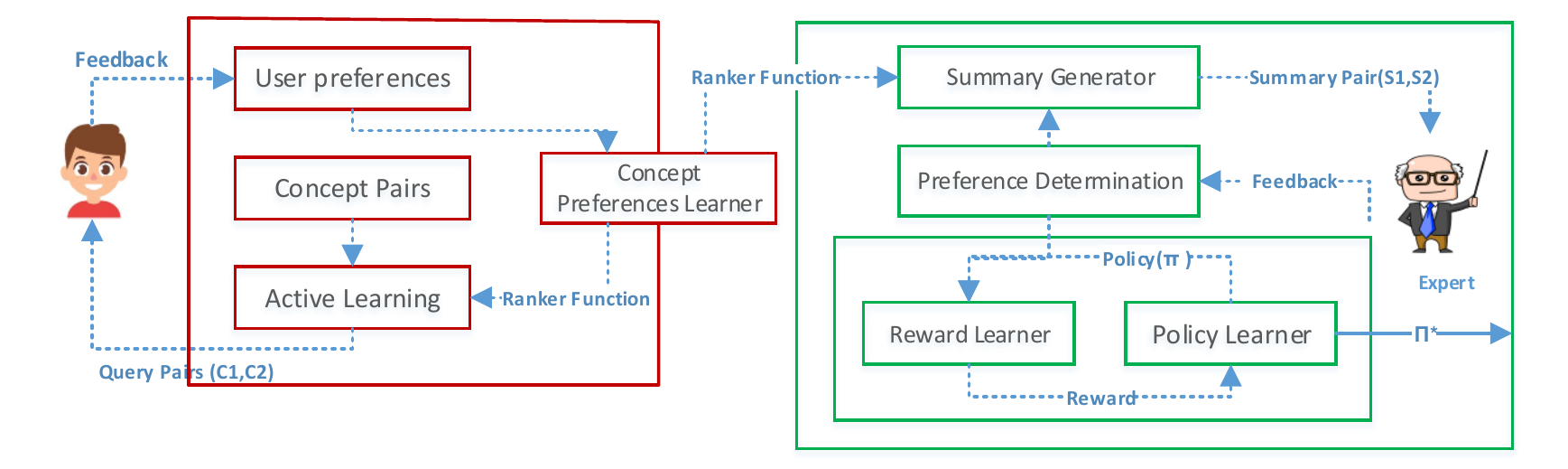}}
    \caption{\textit{SumRecom} approach in more detail: 1) The left side is the user preference extractor using active preference learning over concepts, and 2) the right side is the summarizer including reward learning (IRL) and policy learning (RL).}
    \label{DetailedAlg}
\end{figure*}

One of the ultimate goals of machine learning is to provide predictability. 
Part of this concerns personalisation, which is fundamental to constructing tailored summaries.
Hence, we propose a human-in-the-loop approach to better capture users’ needs.
SumRecom considers the summarisation problem a recommender system, where the goal is to suggest a personalised summary to a user based on their preferences.
This novel framework has two components—(i) the user preference extractor and (ii) the summariser.
The user preference extractor is responsible for querying the user and potentially receiving their feedback using active preference learning; the summariser aims to learn how to create summaries that are tailored to users’ needs based purely on their feedback~\cite{abbeel2004apprenticeship}.
The process is depicted in Fig.~\ref{DetailedAlg} and the overall algorithm is reported in Algorithm~\ref{ch6_alg1}.

\subsubsection{User Preference Extractor}
Understanding users’ interests is the first step towards making personalised summaries.
Individual interests can be extracted implicitly based on personal profiles, browsing history, likes or dislikes, or by retweeting content on social media~\cite{alhindi2015profile}.
Consequently, interaction is gauged to predict users’ perspectives in a variety of circumstances based on the feedback they have provided in the past.
This feedback can be in any form, such as a mouse click or an edit to a post on social media.
Further, experiments suggest that preference-based interactive approaches put less cognitive burden on human subjects compared to asking for absolute ratings or labels, as in a binary decision~\cite{zopf2018estimating,kingsley2010preference}.
For example, asking users to compare two concepts, such as ‘cancer treatment’ and ‘cancer symptoms’, compared to scoring each of these concepts requires less cognitive workload.
Conversely, it is equally challenging for users to decide the value of a summary using a scoring scheme. 
Therefore, to reduce cognitive load, queries are in the form of concept selection, where feedback denotes user preference.

Concept selection aims to find the critical information among the source documents, as humans can quickly assess the importance of concepts given a topic.
Since the notion of importance is specific to a particular topic or user, we queried users’ preference over concepts, as it is easier to prefer one concept to another rather than select important concepts.
However, to collect sufficient data for a meaningful conclusion, we required users to interact with the system in many rounds to simulate ideal user feedback data. 
Therefore, active learning was used to reduce the number of interaction rounds.
To recap, we used active preference learning in an interaction loop to maximise the information gained from small preferences samples, hence, reducing the complexity.

\begin{algorithm}
\begin{algorithmic}[1]
\State \textbf{Input}: Document cluster d
\State \textbf{Output}: Optimal summary for a user
    \caption{SumRecom algorithms.}
    \label{ch6_alg1}
    \State \quad $\textit{Concepts} \gets \textit{Concept extraction (d)}$
    \State \emph{\textbf{Modeling User Preference}}
    \State \quad $\textit{Query pairs} \gets \textit{Active Learner (concepts)}$
    \State \quad $\textit{User Preferences} \gets \textit{Query pairs (user)}$
    \State \quad $\textit{Ranker function} \gets \textit{Preference learner (user preferences)}$
    \State \emph{\textbf{The summariser}}
    \State \quad $\textit{Summaries} \gets \textit{Summary generator (ranking function)}$
    \State \quad $\textit{Summary ranker} \gets \textit{Reward learner (summaries)}$
    \State \quad $\textit{Optimal Policy} \gets \textit{Policy learner (summary ranker)}$
    \end{algorithmic}
\end{algorithm}

\textbf{\textit{Preference learning.}} Preference learning is a classification method that learns how to rank instances based on observed preference information.
It learns based on a set of pairwise preferred items and by obtaining the total ranking of objects~\cite{furnkranz2010preference}.

To formally define the preference learning in the proposed algorithm, let $D$ be the input space and $d$ a cluster of documents.
Also define $C(d)$ as all the extracted concepts from document cluster $x$. Therefore, we have the concept space $C(d)=\{c_1,c_2,..,c_N\}$ with $N$ concepts.
The goal is to query users’ pairwise preference among a set of concepts $\{p(c_{11},c_{21}),p(c_{12},c_{22}),...,p(c_{1n},c_{2n})\}$, where $p(c_{1i},c_{2i})$ is a preference instance shown to users in $i-th$ round, and Eq.~\ref{ch6_pre} is as such:
\begin{equation}
  p(c_{1i},c_{2i})=\begin{cases}
    1, & \text{if $c_{1i}>c_{2i}$}\\
    0, & \text{otherwise},
  \end{cases}
  \label{ch6_pre}
\end{equation}
where $>$  indicates the preference of $c_{1i}$ over $c_{2i}$.
Then, preference learning aims to predict the overall ranking of concepts.
The goal is to find a mapping function to transform data to real numbers, called utility function utility function $U$ such that $c_i > c_j \xrightarrow{} U(c_i) > U (c_j)$ , where U is a function $U: C\xrightarrow{} \mathbbm{R}$.
In this problem, the ground-truth utility function ($U$) measures each concept’s importance based on users’ attitudes ( no two items in $C(d)$ have the equal $U$ value).
Finding the utility function is a regression learning problem that is well studied in machine learning.
In this problem, the ranking function ($R$) measures the importance of each concept based on users’ attitude towards other concepts, defined in Eq.~\ref{ch6_pr2}.
\begin{equation}
\label{ch6_pr2}
    R(c_i)=\sum \mathbbm{1} \{U(c_i)>U(c_j)\} , \forall c_j \in {C(d)},
\end{equation}
where $\mathbbm{1}$ is the indicator function. 
Therefore, $R$ gives the rank of $c_i$ among all extracted concepts in $d$ ($C_d$).

The Bradley–Terry model~\cite{szummer2011semi} is a probability model widely used in preference learning.
Given a pair of individuals $c_i$ and $c_j$ drawn from some population, it estimates the probability that the pairwise comparison $c_i > c_j$ turns out true.
Having $n$ observed preference items, the model approximates the ranking $R$ by computing the maximum likelihood estimate using Eq.~\ref{ch6_pr3}.
\begin{equation}
\label{ch6_pr3}
\begin{split}
J_x(w)= & \sum_{i \in n}[p(c_{1i},c_{2i})log H(c_{1i},c_{2i};w)+\\
 &  [p(c_{2i},c_{1i})log H(c_{2i},c_{1i};w)))],
\end{split}
\end{equation}
where $H(c)$ is the logistic function defined in Eq.~\ref{ch6_pr4}.
\begin{equation}
\label{ch6_pr4}
 H(c_{i},c_{j};w)= \frac{1}{1+exp [U^{*}{(c_j;w)}-U^{*}{(c_i;w)}]}
\end{equation}

$U^{*}$ is the approximation of U parameterised by $w$, which can be learnt through different function approximation techniques.
A linear regression model is designed for this purpose, defined in Eq.~\ref{ch6_eq3}.
\begin{equation}
\label{ch6_eq3}
U(c;w)=w^{T}\phi(c),
\end{equation}
where $\phi(c)$ is the representation feature vector of concept $c$.
For any  $c_i,c_j \in C$, the ranker prefers $c_i$ over $c_j$ if $w^{T}\phi(c_i)> w^{T}\phi(c_j)$.
By maximising the $J_x(w)$ in Eq.~\ref{ch6_pr3} we have Eq.~\ref{ch6_eq31}

\begin{equation}
\label{ch6_eq31}
w^{*} = arg max_w J_x(w)
\end{equation}

The resulting $w^{*}$ using stochastic gradient ascent optimisation will be used to estimate $U^{*}$, which, can be used to produce the approximated ranking function $R^{*}: C \xrightarrow[]{} \mathbbm{R}$.
Maximisation of this objective function will assure those high probabilities are assigned to pairs with low loss.
SumRecom learns a ranking over concepts and uses the ranking to guide the summariser.

\textbf{\textit{Active Learning.}}
\label{sec:activelearn}
To emphasise the need for active learning, consider a scenario in which we have$M$
sentences to summarise, and each sentence has four unique concepts on average.
As a result, the number of unique concepts is  $4\times M$.
Therefore, the number of pairwise preferences to query the user to have a complete comparison in this setting is equal to $\binom {4M}2=\frac{4M!}{2!(4M-2)!}$.
As an example, if $M=100$, this number is equal to $79,800$, which is impossible.
Therefore, the aim of active learning is to find the minimum subset of best samples in our problem, as well as the best pairs, to query the user and gain the most information.
Thus, the number of examples with which to query users is much lower than the number required in regular supervised learning.
There are different strategies to find the minimum subset of best samples. Examples include~\cite{settles2009active}:

\begin{itemize}
\label{strategies}
    \item balance exploration and exploitation: The exploration and exploitation of the data space representation are used to choose samples. In this strategy, the active learning problem is modelled as a CB problem.
    \item expected model change: The policy behind this model is to select the samples that would most change the current model.
    \item expected error or variance reduction: This strategy selects samples that would most reduce the model’s generalisation error or variance.
    \item uncertainty sampling: The idea is to select samples for which the current model is least certain of the correct output.
    \item conformal predictors: This method works based on the similarity of a sample with previous queried samples.
    \item query by committee: Here, different models are trained. The samples with which most models disagree, called the ‘committee’, have the potential to be queried.
\end{itemize}

\begin{algorithm}[t]
\begin{algorithmic}[1]
    \caption{Modeling User Preference.}\label{alg2}
    \State \textbf{Input}: Concepts, learning rate ($\gamma_1$), query budget $t$
    \State \textbf{Output}: Concept ranker function (R)
    \State While {i=0,...,t}
    \State \quad ${(c_{1i},c_{2i}) \gets \textit{Select a pair based on Eq.~\ref{max}}}$
    \State  \quad $p(c_{1i},c_{2i}) \gets \textit{Query user for the feedback}$
    \State  \quad $\textit{$w_{i+1}=w_i+\gamma_1 \frac{\delta J_x(w)}{w}$ based on Eq.~\ref{ch6_pr3}}$
    \State \textbf{return} $\textit{ranker function (R)}$
    \end{algorithmic}
\end{algorithm}

As a solution, we propose a heuristic approach (presented in Algorithm~\ref{alg2}) for selecting query sample pairs.
The approach aims to select the most diverse concepts to compare at first and gradually move to similar ones to reduce the search space.
For this purpose, we partitioned the concepts into clusters based on different similarity measures.

We used semantic and lexical similarity as the features.
A similar measure was proposed by Falke et al.~\cite{falke2019automatic} for grouping similar concepts in the process of making a concept map.
These features include normalised Levenshtein distance, Jaccard coefficient between stemmed content words, semantic similarity based on latent semantic analysis~\cite{deerwester1990indexing}, WordNet~\cite{miller1990introduction}, and word embedding~\cite{mikolov2013distributed}.
Then, we modelled the similarity as a binary classification using logistic regression such that a positive classification, $y= 1$, denotes coreferent concepts, as in Eq.~\ref{ch6_cor}.

\begin{equation}
\label{ch6_cor}
    P(y= 1|c_1,c_2,\theta) = Sigmoid (\theta^T\delta(c_1,c_2)),
\end{equation}
where $\delta(c_1,c_2)$ are the features, $\theta$ the learnt parameters, and the sigmoid function is defined in Eq.~\ref{ch6_delta}.

\begin{equation}
\label{ch6_delta}
S_{\theta}(z)=(\frac{1}{1+{e}^{\theta(1-z)}})
\end{equation}

Based on the similarity of the two concepts, we used an ILP function to find an optimised partitioning schema that maximally matches with the pairwise classifications and is transitive due to the constraints~\cite{barzilay2006aggregation}.
Let $x_{ij}\in \{0,1\}$ be a binary value representing the coreference of concepts $(c_i,c_j)$ and $p(c_i,c_j)$ denotes their coreference probabilities.
The goal is to optimise the objective function in Eq.~\ref{max} using a greedy local search to partition similar concepts.

\begin{equation}
\begin{split}
&\sum_{\substack{c_i,c_j \in C^2}} p(c_i,c_j)x_{ij}+ (1-p(c_i,c_j)) (1-x_{ij})\\
& s.t. \hspace{2mm} x_{ik} \geq x_{ij}+x_{jk}-1 \hspace{2mm} \forall i,j,k \in [1,..,|C|]\hspace{2mm} and \hspace{2mm} i \neq j \neq k
\end{split}
\label{max}
\end{equation}

After partitioning similar concepts in each iteration (the number of iterations is equal to the query budget), we selected two concepts in different partitions.
These concepts are chosen based on the trained similarity measure and by minimising the similarity of concepts chosen and previously queried pairs to gain maximum information.

\subsubsection{The Summariser}
The user preference extractor’s output is the ranking function that estimates each concept’s importance based on user feedback.
The summariser is responsible for making the desired summaries based on the given preferences. 
The summariser consists of three phases—(i) a summary generator; (ii) an IRL to learn how to evaluate generated summaries based on an expert’s evaluation history; and iii) an RL to learn how to generate the desired summary for the user.

\textbf{\textit{The summary generator.}}
Learning the importance of concepts for a user, $R$ function, we constructed summaries that are likely suited to users’ desire to reduce the summary search space. 
Let $C(d)$ be the set of concepts in the input documents $d$, $p_{c_i}$ is the existence of the concept $c_i$ in the output to which ${c_i}$ belongs in this sentence, $w_i$ is the concept’s weight (importance), $l_j$ denotes the sentence length $j$, $p_{s_j}$ is the existence of sentence $j$ in the output, and $L$ is the summary length constraint defined by the user.
Based on these definitions, we formulated the following optimisation function (Eq.~\ref{ch6_sumgen}) using ILP, which selects sentences with important concepts based on user feedback:

\begin{equation} 
\label{ch6_sumgen}
\begin{split}
  &  max \sum_{i} w_ip_{c_i}\hspace{2mm} where \hspace{2mm}  \forall i\in [1,..,|C|] \hspace{2mm} and \hspace{2mm} \forall c_i \in s_j \sum_{j} l_jp_{s_j}<L 
\end{split}
\end{equation}

Weights are based on the R function learnt in the previous part. 
Then, a summary pool is made using Eq.~\ref{ch6_sumgen}.
To generate a diverse summary pool, among top score summaries (according to Eq.~\ref{ch6_sumgen}), we selected summaries with no redundancy.
This is defined as the similarity of sentences within a summary without considering a user’s mentioned concepts divided by the summary length. 
Document summarisation is then formulated as a sequential decision-making problem, solved by a proposed RL algorithm.

\textbf{\textit{Problem Definition.}}
We formulated summarisation as a discrete optimisation problem inspired by the APRIL approach~\cite{gao2018april}. 
Let $Y_d$ indicate the set of all extractive summaries
for the document cluster $d$, where $y_d \in Y_d$ is a potential summary for document cluster $d$.
The summarisation task is to transform each input $d$ to the best summary in $Y_d$ for the learnt preference ranking function. 
Extractive MDS is defined as a sequential decision-making problem that sequentially chooses sentences and adds them to a draft summary.
Therefore, it can be defined as an episodic Markov decision process (MDP) problem.

An episodic MDP is a tuple $(S,A,P,R,T)$, where $S$ is the set of states, $A$ is the set of actions, $P:S\times A \times S \xrightarrow{} \mathbbm{R}$ is the transition function, $R(S,A)$ is the reward for performing an action ($A$) in a state (S), and $T$ is the set of terminal states.
In the extractive MDS context, a state is a draft summary and A includes two types of actions\textemdash adding a new sentence to the current draft summary or terminating the summary construction process~\cite{gao2019reward}. 
The reward function $R$ returns the reward score once the action is terminated; otherwise, it returns $0$ because the summary is still under development and, hence, cannot be evaluated.
A policy $\pi(S,A): S \times A \xrightarrow{} R$ in an MDP defines the selection of actions in state S.

Episodic MDP for modelling document summarisation has two components: (i) reward, or what is defined as a good summary; and (ii) policy, or how to select sentences (actions) to maximise the rewards.
State-of-the-art summarisation approaches are divided into two categories: cross-input paradigms and input-specific paradigms~\cite{gao2019reward}.
The former employs RL algorithms such that the agent interacts with a ground-truth reward oracle over multiple episodes to learn a policy that maximises the accumulated reward in the episode.
The learnt policy is then applied to unseen data at test time (in this problem to generate summaries using new data).
However, learning such a cross-input policy requires significant data, time, and tuning parameters because of the broad search spaces and delayed rewards.
Conversely,  a more efficient alternative to learning a policy is to learn input-specific RL policies which do not require parallel data or reward oracles.
However, these policies depend on handcrafted rewards, so they are challenging to create to fit all inputs~\cite{gao2019reward}.

SumRecom takes advantage of two categories of cross-input and input-specific RL.
First, it learns a cross-input reward at the training time and then employs the learnt reward to train an input-specific policy at test time.

\textbf{\textit{The reward learner.}} SumRecom is inspired by IRL~\cite{abbeel2004apprenticeship}, where instead of the policy, it first learns the reward utilising a domain expert to find optimal trajectories. 
The demonstrator is a domain expert who can evaluate two summaries based on user feedback.
The domain expert  can be another RL agent trained to become an advisor for other learner agents, or a human.
To learn the ground-truth reward, numeric scores indicating the summaries' quality and preferences over summary pairs are used~\cite{kreutzer2018reliability}.
In practice, leveraging preference learning reduces the cognitive load and, consequently, the inevitable noise in evaluating summaries.
SumRecom learns from preference-based (pairwise) feedback that ranks preferences over summary pairs.
Then, the summaries are queried to the demonstrator for evaluation.

For point-based oracles, we drew $L$ sample outputs from the summary pool without replacement.
We used ExDos to evaluate the summaries based on three measures: coverage, salience and redundancy.
In each iteration, unique summaries compared with previously selected samples were chosen for query. 
The same approach was also selected for preference-based summaries. 
We then queried their score values ($V$) from the oracle and used a regression algorithm to minimise the average mean squared error between $V$ and the approximate value $V^*$ , where the loss function is defined in Eq.~\ref{ch6_los}.

\newcommand{\Lagr}{\mathcal{L}}
\begin{equation}
\label{ch6_los}
    \Lagr^{MSE}= \frac{1}{L}\sum_{i=1}^{L}(V^*-V)^2
\end{equation}

In pairwise oracles, we denote the collected preferences by $P_s=\{p(y_{11},y_{12}),...,p(y_{1l},y_{2l})\}$,
where $y$ denotes the summary and $l$ sample pairs are queried. 
Then, the procedure is the same as in Eq.~\ref{ch6_pr3} using the cross-entropy loss function, defined in Eq.~\ref{losspresum} and Eq.~\ref{ch6_eq211}.

\begin{equation}
\label{losspresum}
\begin{split}
  \Lagr^{CE}= & -\sum_{i \in l}[p(y_{1i},y_{2i})log H(y_{1i},y_{2i};w)+\\
 &  [p(y_{2i},y_{1i})log H(y_{2i},y_{1i};w)))],
\end{split}
\end{equation}
where
\begin{equation}
\label{ch6_eq211}
 H(y_{1i},y_{2i};w)= \frac{1}{1+exp [V^{*}{(y_j;w)}-V^{*}{(y_i;w)}]}
\end{equation}

The output is the ranked function, $V$, which demonstrates each summary’s reward compared to the others.
Since such a demonstrator is hardly available in practise, SumRecom leverages an approximate function to learn the reward.

\begin{algorithm}[t]
    \begin{algorithmic}[1]
    \caption{Summariser}\label{alg3}
    \State \textbf{Input}: Concept ranker (R), learning rate ($\gamma_2$) 
    \State \textbf{Output}: Optimal summary for a user (S)
    \State $\textit{Summaries} \gets \textit{Generating summaries using Eq.~\ref{ch6_sumgen}.}$
    \State While{ i=0,...,t}
    \State \quad $\textit{$(y_{1i},y_{2i})$}  \gets \textit{Select a pair based on Eq.~\ref{losspresum} from summaries}$
    \State \quad $\textit{$p(y_{1i},y_{2i})$} \gets \textit{Query user for the feedback}$
    \State \quad  $\textit{$w_{t+1}=w_t-\gamma_2 \frac{ \Lagr^{CE}(w)}{\delta w}$ based on Eq.~\ref{ch6_los} or Eq.\ref{losspresum}.}$
    \State \quad  $\pi^* = arg max R^{RL}(\pi|x) = arg max \sum_{y in Y}\pi(y)V(y)$
    \end{algorithmic}
\end{algorithm}

\textbf{\textit{The policy learner.}}
The goal of policy learning is to search for optimal solutions in MDP environments.
We modelled the summarisation problem as an episodic MDP, meaning that each action’s reward is equal to 0 if the state is not terminated. 
At each step, the agent can perform either of the two actions\textemdash add another sentence to the summary or terminate it.
The immediate reward function $R(S,A)$ assigns the reward if $S$ is the terminate state.
The reward in SumRecom is the learnt expert’s reward, $V$.
A policy $\pi$ defines the strategy to add sentences to the draft summary to build the summary for the user.
SumRecom then defines  $\pi$  as the probability of choosing a summary of $y$ among all summaries $Y$, denoted as $\pi(y)$. 
Therefore, the optimal policy, $\pi^*$, is the function that finds the desired summary for a given input based on users’ feedback.
The expected reward of applying policy $\pi$ is defined in Eq.~\ref{ch_15}.

\begin{equation}
\label{ch_15}
    R^{RL}(\pi|d)= \mathbbm{E}_{y \in Y}R(y)= \sum_{y\in Y} \pi(y)R(y),
\end{equation}
where $R(y)$ is the reward for selecting summary $y$ in document cluster d. 
In our problem, the reward is the ranker approximated by the domain expert, $V$.
Therefore, the accumulated reward to be maximised in our problem is equal to Eq.~\ref{ch_16}.

\begin{equation}
\label{ch_16}
    R^{RL}(\pi|d)= \sum_{y\in Y} \pi(y)V(y)
\end{equation}

The MDP aims to find the optimal policy $\pi^*$ that has the highest predicted reward (Eq.~\ref{ch6_17}).

\begin{equation}
\label{ch6_17}
\pi^* = arg max R^{RL}(\pi|d) = arg max \sum_{y \in Y}\pi(y) V(y)
\end{equation}
We used the linear temporal difference algorithm to obtain $\pi^*$.
The summariser algorithm is explained in Algorithm~\ref{alg3}.

\subsection{Experiment}
\label{Evaluation}
This section presents the experimental set-up for implementing and assessing the proposed model.
First, we explain the evaluation settings and then present the results for analysis.
SumRecom is evaluated from different evaluation aspects, including:
\begin{itemize}
    \item the effect of different features in approximating preference learning algorithms used in the model, including concept and summary preferences
    \item the use of different strategies for active learning
    \item the role of the query budget in both concepts and summary preferences
    \item the quality of produced summaries
    \item a human study to evaluate SumRecom from users’ perspective
    \item the effect of different values for parameters
    \item an ablation study to evaluate different components of the proposed framework.
\end{itemize}

\subsubsection{Feature Analysis} 
\begin{figure}[t]
\centering
    \includegraphics[width=0.5\linewidth]{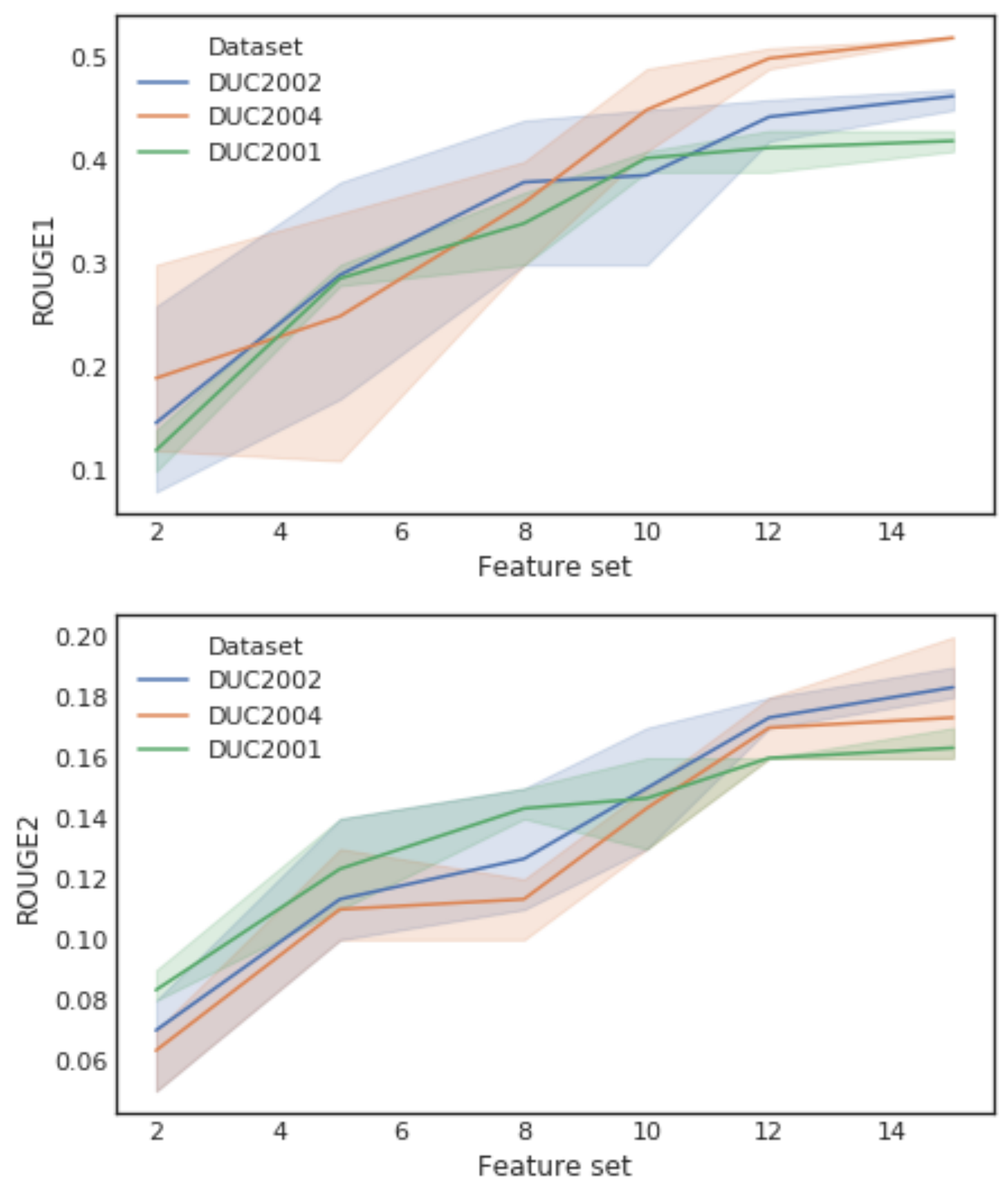}
    \caption{Features for estimating the ranker function for preferred concepts.}
    \label{2featureAnalysis}
\end{figure}

Before evaluating the effect of concept preference in summarisation, it is important to explain the ground-truth concept ranker function ($U$) and the approximate function ($U^*$).
The ground-truth concept ranker function ($U$) indicates the importance of each concept.
We defined a predefined list of preferences over concepts and the ground-truth concept ranker value for 10 clusters to simulate users’ preferences.
To estimate the approximate function ($U^*$), we defined a linear model $U^*(c)=W^T\phi(c)$ where $\phi$ are the features.
To this end, a set of features, whose importance was validated using ExDos (including surface-level and linguistic-level features), was used.
Surface-level features include frequency-based features (TF-IDF, RIDF, gain, and word co-occurrence), word-based features (upper-case words and signature words), similarity-based features (Word2Vec and Jaccard measure), sentence-level features (position, length cut-off and length), and named entities. 
Linguistic features are generated based on a semantic graph, including the average weights of connected edges, the merge status of a sentence as a binary feature, the number of concepts merged with a concept, and the number of concepts connected to that concept.
We defined different combinations of features, $\{2,5,8,10,12,15\}$, starting from the most critical feature based on the importance estimated using ExDos.
We repeated the experiments for 10 cluster documents.
The results of ROUGE-1 and ROUGE-2 are reported in Fig.~\ref{2featureAnalysis}.

The results show that the model performs better after adding more features; however, the last set of features did not significantly affect ROUGE values. 
It is worth mentioning that adding more features also increases complexity.
To simulate domain expert knowledge for evaluating summaries based on the available feedback (reward), we modelled the ground-truth summary reward (V) based on the three measures (ROUGE-1, ROUGE-2 and the redundancy) defined in Eq.~\ref{ch6_reward}.

\begin{equation}
\label{ch6_reward}
    V=\alpha ROUGE1+ \beta ROUGE2 - \gamma Redundancy
\end{equation}

ROUGE-1 and ROUGE-2 are extensively used for human evaluation of summary quality.
Redundancy is defined as the similarity of sentences within a summary without considering the user’s mentioned concepts, divided by the summary length.
To approximate the ground-truth reward function, we employed a linear function as $V^*=w^T\lambda(y)$, where $\lambda(y)$, which denotes both features and concepts in addition to ROUGE-1 and ROUGE-2 as the complete feature set.
The results follow the same trend in Fig.~\ref{2featureAnalysis}, except that adding ROUGE-1 and ROUGE-2 as the feature set improves the performance of the final summaries by an average of $.13$ times.

\subsubsection{Active Learning Strategy Analysis}
\begin{figure}[t]
\centering
    \includegraphics[width=0.755\linewidth]{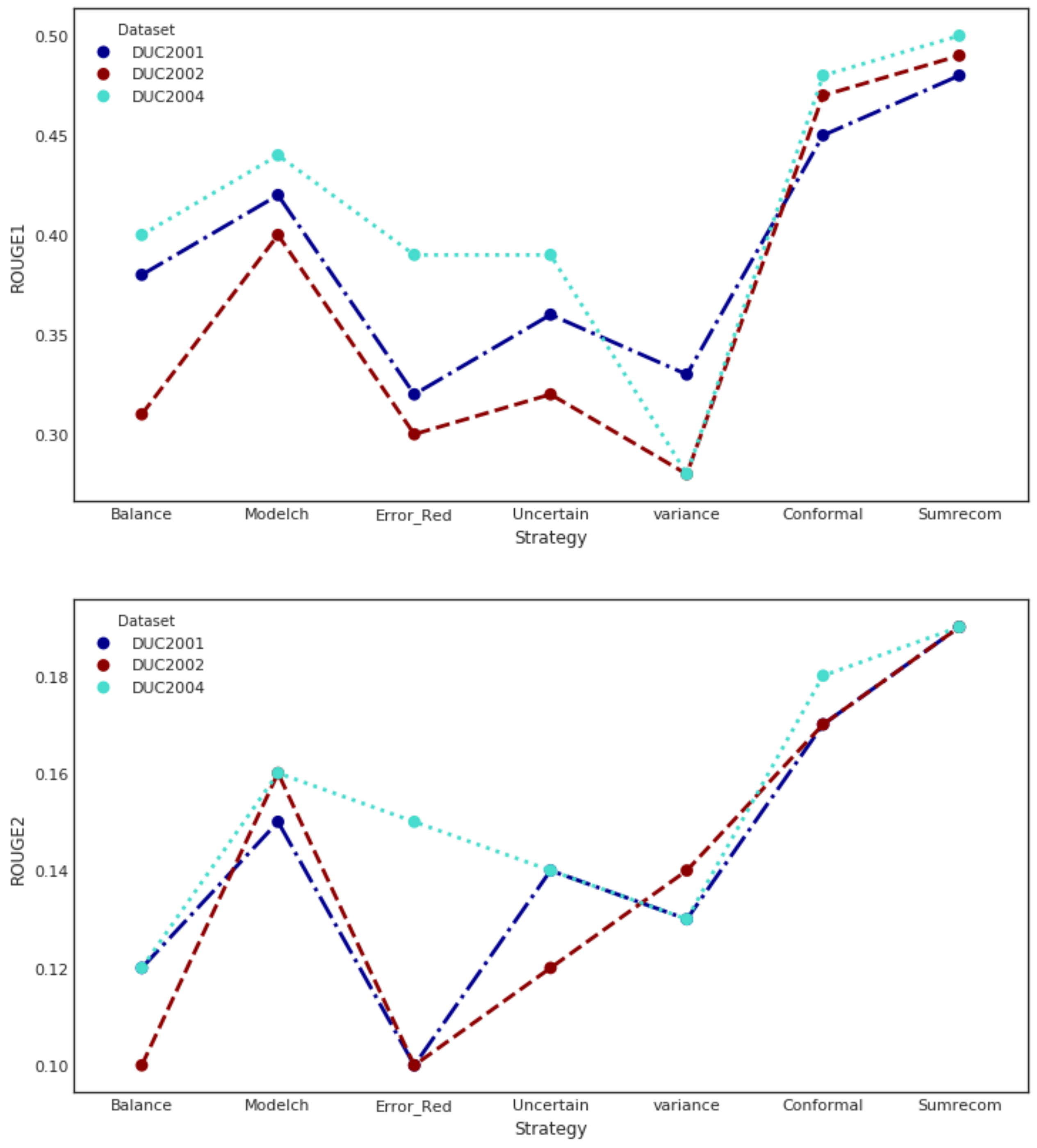}
    \caption{Comparing different strategies used in active learning.}
    \label{fig:strategies}
\end{figure}

To evaluate the effect of the proposed heuristic approach for active learning, we compared SumRecom with six different strategies explained in Sec.~\ref{sec:activelearn}.
ROUGE-1 and ROUGE-2 results for each strategy are reported in Fig.~\ref{fig:strategies}, proving the supremacy of the proposed heuristic approach for our problem.
As shown, the conformal approach performs similar to the proposed heuristic function.
Besides, the change model is superior in both cases, proving that selecting the most diverse concepts results in better summaries.

\subsubsection{Query Budget Analysis}
\begin{figure}
\centering
    \includegraphics[width=0.85\linewidth]{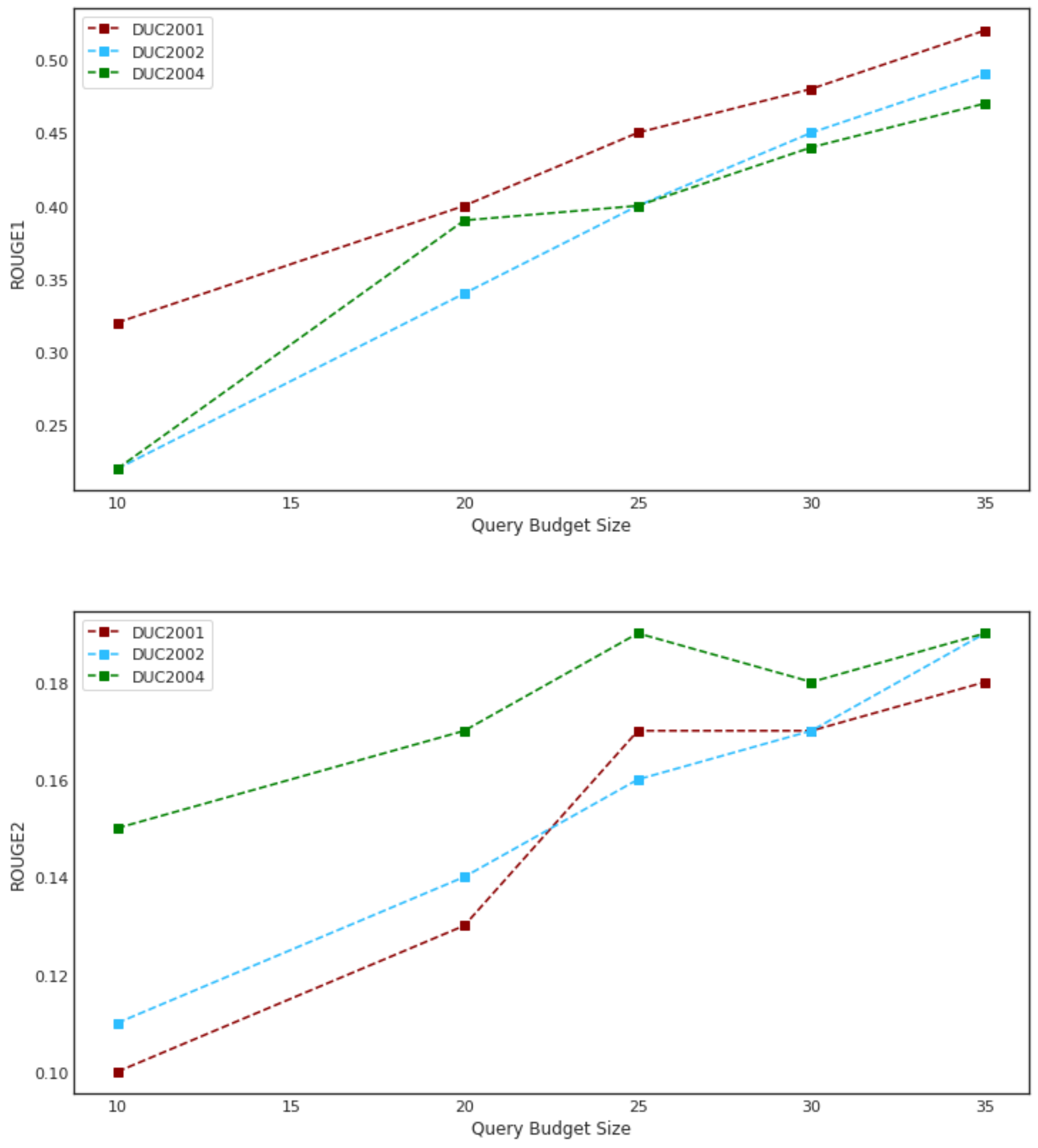}
    \caption{effect of query budget size in the quality of summaries.}
    \label{querybudget}
\end{figure}

We also measured the effectiveness of users’ query budget size in the process.
We chose the query size among the selection of $\{10,15,20,25,30,35\}$, demonstrating each user’s respective amount of feedback.
The results are reported in Fig.~\ref{querybudget}.
As expected, increasing the feedback number concurrently increases the ROUGE score significantly.
However, the difference rate simultaneously decreases in the process.

\subsubsection{Summary Evaluation}
To evaluate the coverage aspect of summaries generated by SumRecom, we used the reference summaries so that the mentioned concepts that exist in reference summaries receive the maximum score by the ranked function.
Here, the reward function is the ROUGE score after comparing summary references. We compared SumRecom to state-of-the-art strength competitors including APRIL and SPPI on three benchmark datasets reported in Table~\ref{2tab:data1},~\ref{2tab:data2} and ~\ref{2tab:data3}.
The results show the supremacy of SumRecom.
Indeed, the main goal of this approach is to help users create their desired summary while assuming little cognitive load. 
Thereafter, we conducted a study to evaluate cognitive load within this context.

\begin{table}
\centering
  \caption{Comparing SumRecom on DUC2001 Dataset}
  \vspace{1mm}
  \label{2tab:data1}
  \begin{tabular}{|l|c|c|c|}
    \hline
    Model &ROUGE-1 &ROUGE-2 &ROUGE-L\\
    \hline
    APRIL &0.325  &0.070 &0.26\\
    \hline
    SPPI  &0.232  &0.068 &0.259\\
    \hline 
    SumRecom &0.341 &0.078 &0.28\\
 \hline
\end{tabular}
\end{table}

\begin{table}
\centering
  \caption{Comparing SumRecom on DUC2002 Dataset}
    \vspace{1mm}
  \label{2tab:data2}
  \begin{tabular}{|l|c|c|c|}
     \hline
    Model &ROUGE-1 &ROUGE-2 &ROUGE-L\\
    \hline
     APRIL &0.351 &0.078 &0.279\\
      \hline
     SPPI &0.350 &0.077 &0.278\\
      \hline
     SumRecom &0.372 &0.083 &0.333\\
   \hline
\end{tabular}
\end{table}
\begin{table}
\centering
  \caption{Comparing SumRecom on DUC2004 Dataset}
    \vspace{1mm}
  \label{2tab:data3}
  \begin{tabular}{|l|c|c|c|}
     \hline
    Model & ROUGE-1 & ROUGE-2 & ROUGE-L\\
    \hline
     APRI L&0.373 &0.093 &0.293\\
      \hline
     SPPI &0.372 &0.093 &0.293\\
      \hline
     SumRecom &0.382 &0.945 &0.301\\
   \hline
\end{tabular}
\end{table}

\subsubsection{Human Analysis} 
The goal of SumRecom is to help users generate the desired summary with low cognitive load.
Therefore, we conducted two human experiments to evaluate the model.
We hired 15 MTurk workers to perform the tasks without any specific prior background required.
Ten document clusters were randomly selected from the DUC datasets, and each participant was presented with five random documents to avoid any subject bias, with two minutes to read each article. 
To ensure the human subjects understood the study’s objective, workers were asked to complete a qualification task first.
They were required to write a summary of their understanding.
We also manually removed spam answers from the results.
Spam was defined based on the qualification task and the response time (i.e., a very short answer time is unacceptable, as it proves random and/or imprecise).

In the first experiment, participants were asked to define their preferences by comparing some concepts. 
The generated summary based on the given feedback and four other general summaries were shown, and participants were asked to choose their preferred summary.
Approximately 83\% of participants selected the generated summary produced by SumRecom.
Then, the produced approach for each summary was provided.
Participants were asked to define their satisfaction level and evaluate the produced summary based on their given feedback by assigning a rating between 0 and 10.
The average rating of summaries produced by SumRecom was $8.2$, demonstrating a reasonable level of satisfaction.

In the second experiment, to assess whether users can obtain the information they desire from a summary, participants were asked to answer a question about each topic by selecting an answer among a selection of potential responses.
The questions were defined by the authors and covered both specific and general information. Participants’ answers and their level of confidence in responding were recorded.
An evaluator then assessed their accuracy. Among the 15 workers, 86.67\% were completely confident in their answers; however, only 80\% were accurate.

\begin{table}
\centering
  \caption{Overview of the Parameters Used in Simulation Experiments}
  \label{tab:parameter}
  \begin{tabular}{|c|c|l|}
     \hline
    Parameter & Description & Value\\
    \hline
    L & User input &Summary length\\
     \hline
    $\alpha$ &0.8 &ROUGE-1 coefficient for ground truth reward in Eq.~\ref{ch6_reward}\\
     \hline
    $\beta$ &0.5  &ROUGE-2 coefficient for ground truth reward in Eq.~\ref{ch6_reward}\\
     \hline
    $\gamma$ &0.25 &redundancy coefficient for ground truth  reward in Eq.~\ref{ch6_reward}\\
     \hline
    $\gamma_1$ &0.001 &learning rate for concept preference in Eq.~\ref{ch6_pr3}\\
     \hline
    $\gamma_2$ &0.005 &learning rate for summary preference in Eq.~\ref{ch6_los}\\
     \hline
\end{tabular}
\end{table}

\begin{table}
\centering
  \caption{Comparing SumRecom, SumRecom-AC and SumRecom-PR on DUC2001 Dataset}
    \vspace{1mm}
   \label{AblPreference1}
  \begin{tabular}{|l|c|c|c|}
     \hline
    Model & ROUGE-1 & ROUGE-2 & ROUGE-L\\
    \hline
     SumRecom-AC &0.103 &0.031 &0.140\\
      \hline
     SumRecom-PR &0.112 &0.001 &0.129\\
      \hline
     SumRecom &0.341 &0.078 &0.28\\
   \hline
\end{tabular}
\end{table}
\begin{table}[t]
\centering
  \caption{Comparing SumRecom, SumRecom-AC and SumRecom-PR on DUC2002 Dataset}
    \vspace{1mm}
   \label{AblPreference2}
  \begin{tabular}{|l|c|c|c|}
     \hline
    Model & ROUGE-1 & ROUGE-2 & ROUGE-L\\
    \hline
     SumRecom-AC &0.190 &0.018 &0.132\\
      \hline
     SumRecom-PR &0.157 &0.021 &0.198\\
      \hline
     SumRecom &0.572 &0.083 &0.333\\
   \hline
\end{tabular}
\end{table}

\subsubsection{Parameter Analysis}
As in other parametric models, SumRecom has some hyper-parameters that require tuning.
A wide range of these values was tested, and the correlation between them subsequently analysed.
To reduce the need for reading and to help researchers replicate the proposed algorithm, Table~\ref{tab:parameter} summarises the parameters with the same description used in our simulation experiments.

\begin{table}
\centering
  \caption{Comparing SumRecom, SumRecom-AC and SumRecom-PR on DUC2004 Dataset}
    \vspace{1mm}
   \label{AblPreference4}
  \begin{tabular}{|l|c|c|c|}
     \hline
    Model & ROUGE-1 & ROUGE-2 & ROUGE-L\\
    \hline
     SumRecom-AC &0.111 &0.033 &0.143\\
      \hline
     SumRecom-PR &0.200 &0.021 &0.182\\
      \hline
     SumRecom &0.382 &0.945 &0.301\\
   \hline
\end{tabular}
\end{table}
\begin{table}
\centering
  \caption{Comparing SumRecom and SumRecom-GE}
    \vspace{1mm}
   \label{tge}
  \begin{tabular}{|l|l|l|l|l|l|l|}
    \hline
    \multirow{2}{*}{Dataset} &
      \multicolumn{2}{c}{SumRecom} &
      \multicolumn{2}{c|}{SumRecom-GE} \\
    & ROUGE-1  & ROUGE-2  & ROUGE-1  & ROUGE-2 \\
    \hline
    DUC2001 & 0.341 & 0.078 & 0.222 & 0.021 \\
    \hline
    DUC2002 & 0.572 & 0.083 & 0.189 & 0.047  \\
    \hline
    DUC2004 & 0.382 & 0.945 & 0.109 & 0.510  \\
    \hline
  \end{tabular}
\end{table}
\subsubsection{Ablation Study} 
To evaluate each component’s incremental contribution in our proposed framework, we ran ablation studies comparing our model ablations against each other. 
First, we evaluated the preference extractor, and, thus, removed active learning by selecting random pairs from the concept database, called SumRecom-AC.
Then, the whole preference learner component was removed, called SumRecom-PR.
The ROUGE scores of these approaches compared with the reference summaries after 10 complete runs are averaged and compared in Table~\ref{AblPreference1}, Table~\ref{AblPreference2} and ~\ref{AblPreference4}, respectively.
The results clearly show the effect of both active learning and preference learning.

In another experiment to evaluate the role of learning, the summaries generated by a summary generator were considered output, called SumRecom-GE. 
The results are reported in Table~\ref{tge}, based on the average of 10 produced summaries.

\section{Summation}
Lack of structure makes summaries challenging to process, but attempting to predict both personalised and structured summaries is a whole other task.
As such, we propose a hierarchical personalised concept-based summarisation approach called ‘Summation’, which sums up a collection of documents to a concise hierarchical concept map.
Instead of providing a short and static summary, Summation engages users by querying their preferences to learn what they desire.
An RL algorithm is then used to provide a personalised summary of unseen topic-related documents.
Summation provides a concise overview of a document collection for a specific user, structures it across document boundaries, and can be used as a navigator in document collections.

\subsection{Problem Definition}
The input is a set of documents $D=\{D_{1},D_{2}, ... ,D_{N}\}$ and each document consists of a sequence of sentences $S=[s_1,s_2,$$...$$,s_n]$. 
Each sentence $s_i$ is a set of concepts $\{c_1,c_2, ..,c_k\}$, where a concept can be a word (unigram) or a sequence of words.
The output is a personalised hierarchical concept map.
This novel framework has two components, an organiser and a summariser, explained in Sec.~\ref{sum_Organizer} and \ref{sum_summarizier}, respectively.

\subsection{Organiser (Structuring Unstructured Data)}
\label{sum_Organizer}
The first step is to structure unstructured information by making a hierarchical concept map.
A concept map is a graph with directed edges, where nodes indicate concepts and edges indicate relations.
Both concepts and relations are sequences of related words representing a semantic unit.
Consequently, the first step in creating a concept map is to identify all concepts and relations.
Here, we propose hierarchical clustering to form the hierarchical concept map. Abstract labels are created to make summaries concise and coherent.

\subsubsection{Concept and Relation Extraction}
Concepts come in different syntactic types, including nouns, proper nouns, more complex noun phrases, and verb phrases that describe activities~\cite{falke2019automatic}.
For this purpose, we used open information extraction (OIE)~\cite{etzioni2008open} through which the entities and relations are obtained directly from the text.
OIE finds binary propositions from a set of documents in the form of ($con_1$,$R$,$con_2$), which are equivalent to the desired concepts and relations. 
For example, the output for the sentence, ‘cancer treatment is underpinned by the Pharmaceutical Benefits Scheme’, is:

\textit{Cancer treatment$\xrightarrow[\text{}]{\text{is underpinned}}$ by the Pharmaceutical Benefits Scheme}

Balancing precision and recall in extracting concepts is a challenging task.
High precision means all identified spans will be defined as mentions of a concept.
Therefore, some constructions are usually missed, and this lowers the recall.
Conversely, a high recall is necessary since missed concepts can never be in a summary.
Achieving a higher recall may cause extracting many mentions and increasing false positives.
Generalisability is also essential.
The reason is that extracting a particular syntactic structure might generate only correct mentions, causing too broad mentions.
Ideally, a proper method applies to many text types. 
To avoid meaningless and long concepts, we processed the OIE results such that concepts with less than one noun token or more than five tokens are omitted.
The original nouns also replace pronouns.
If an argument is a conjunction indicating conj-dependency in the parse tree, we split them.

\subsubsection{Concept Map Construction}
Among various extracted concepts and relations, multiple expressions can refer to the same concept while not using precisely the same words; that is, they can also use synonyms or paraphrases.
However, distinguishing similar concepts to group them is challenging and subjective.
For example, adding a modifier can completely change the meaning of a concept based on the purpose of summarisation.
Consequently, grouping them may lead to propositions that are not stated in the document.
Therefore, we need to group every subset that contains mentions of a single, unique concept.
Scalability is another critical issue.
For example, pairwise comparisons of concepts cause a quadratic run-time complexity applicable only to limited-sized document sets.
The same challenges exist for relation grouping.
However, we first grouped all mentions by the concepts' pairs, and then performed relation grouping.
Therefore, this task’s scope and relevance are much smaller than when concepts are used.
Therefore, in practise, comparison-based quadratic approaches are feasible.
Moreover, as the final goal is to create a defined size summary, the summary size significantly affects the level of details in grouping concepts. 
This is because the distinction between different mentions of a concept might not be required, as it is a subjective task.
Ideally, the decision to merge must be made based on the final summary map’s propositions to define the necessary concept granularity.

We further propose hierarchical conceptual clustering using k-means with word embedding vectors to tackle this problem, as it spans a semantic space. 
Therefore, word embedding clusters give a higher semantic space, grouping semantically similar word classes under the Euclidean metric constraint defined below.
Before defining the proposed hierarchical conceptual clustering, we review word embedding schemes used in the proposed model.

\textbf{\textit{Word Embedding.}}
Word embedding is a learnt representation of text such that the same meaning words have similar representations.
Different techniques can be used to learn a word embedding from the text.
Word2Vec~\cite{mikolov2013distributed} is an example of a statistical model for learning a word embedding representation from a text corpus, utilising different architectures.
As such, we used skip-gram and bag of character n-grams in our experiments.
The skip-gram model uses the current word for predicting the surrounding words by increasing the weights of nearby context words more than other words using a neural network model.
One drawback of skip-gram is its inability to detect rare words.
In another model, authors define an embedding method by representing each word as the sum of the vector representations of its character n-grams, known as ‘bag of character n-grams’~\cite{bojanowski2017enriching}.
If the training corpus is small, character n-grams will outperform the skip-gram (of words) approach.~\footnote{We used fastText for word embedding:  https://fasttext.cc/docs/en/support.html}

\textbf{\textit{Conceptual Hierarchical Clustering.}} 
Given word (concept) embeddings learnt from a corpus, $\{v_{w_1},v_{w_2},...,v_{w_T}\}$, we propose a novel recursive clustering algorithm to form a hierarchical concept map, $H$.
This variable denotes a set of concept maps organised into a hierarchy that incrementally maintains hierarchical summaries from the most general node (root) to the most specific summary (leaves).
Within this structure, any non-leaf summary generalises the content of its children nodes. 
Hierarchical summarisation has two critical strengths in the context of large-scale summarisation.
First, the initial information under review is small and grows upon users’ request, so as not to overwhelm them.
Second, the parent-to-child links facilitate user navigation and drilling down for more details on interesting topics.
The hierarchical conceptual clustering minimises the objective function Eq.~\ref{clustering2} over all k clusters as C=\{$c_1,c_2,..,c_k$\}. 

\begin{equation}
\label{clustering2}
J = \sum_{k=1}^{K}\sum_{t=1}^{\mid T \mid} |v_{w_t}- c_k|^2 +\alpha \min _{c\in C} size(c),
\end{equation}
where $c_k$  is the randomly selected centre $k-th$ cluster.
The second term is the evenness of the clusters, added to avoid clusters with small sizes.
$\alpha$ tunes the evenness factor, which was defined by employing a grid search over a development set.
We also implemented hierarchical clustering top-down at each time, optimising Eq.~\ref{clustering2}.
After defining the clusters, we must find the concept that best represents every concept at the lower levels to ensure hierarchical abstraction. 
A concise label is the desired label for each node; however, shortening mentions can introduce propositions that are not asserted by a text.
For example, the concept labelled ‘students’ can change in meaning where the emphasis is on a few students or some students.
To this end, a centre of a cluster at each level of the hierarchy was defined as a label.
The inverse distance to the cluster centres is the membership degree or the similarity to each label.
The cluster distance for a word $w_t$ is defined as $d_{v_{w_t}}$.
Consequently, the membership of each word $w_t$ in cluster $c_k$ to its label is the inverse distance defined in Eq.~\ref{ch6_eq20}.
\begin{equation}
\label{ch6_eq20}
m_{v_{w_t}}=\frac{1}{d_{v_{w_t}}} =\frac{1}{|c_k-v_{w_t}|^2} \hspace{2mm} \forall w_t \in c_k
\end{equation}

We then fine-tuned K within the 5–50 range based on the dataset size and chose the cluster number according to gap statistic value~\cite{tibshirani2001estimating}.
The output $H$ can be directly used as a new dataset for other actions, such as browsing, querying, data mining process, or any other procedures requiring a reduced but structured version of data.
The hierarchical clustering can also be pruned at each level to represent a summarised concept map for different purposes or users.
Therefore, $H$ is fed to the summariser for pruning to generate a personalised summary. Moreover, by using preference-based learning and RL, we learn users’ preferences in making personalised summaries for unseen topic-related documents, discussed in Sec.~\ref{sum_summarizier}.

\subsection{Summariser}
\label{sum_summarizier}
The hierarchical concept map produced in the previous step is given to the summariser to make the desired summaries for users based on their given preferences.
Therefore, the summariser consists of two phases\textemdash(i) predicting user preferences and (ii) generating the desired summary.

\subsubsection{Predicting User Preference}
The first step towards creating personalised summaries is to understand users’ interests.
The same procedure in SumRecom is used for extracting users’ preferences; however, the selection of sentences is among hierarchy nodes.
$H$ is the hierarchical concept map, where at the $i-th$ level of the hierarchy there exist $m_i$ nodes defining a label $l$.
$L=\{l_{11},...,l_{nm_i}\}$ is the set of all labels, where $l_{i1}$ indicates the first node at $i-th$ level of the hierarchy and $n$ is the number of levels, and $L_i$ indicates the labels at  $i-th$ level.
We queried users with a set of pairwise concepts at the same levels,
$\{p(l_{i1},l_{i2}),p(l_{i2},l_{i3}),...,p(l_{im_i-1},p(l_{im_i})\}$, where $p(l_{i1},l_{i2})$ is defined in Eq.~\ref{ch6_eq21}.
\begin{equation}
\label{ch6_eq21}
  p(l_{i1},l_{i2})=
  \begin{cases}
    1, & \text{if $l_{i1}>l_{i2}$}\\
    0, & \text{otherwise},
  \end{cases}
\end{equation}
where $>$ indicates the preference of $l_{i1}$ over $l_{i2}$.
Preference learning aims to predict the overall ranking of concepts, which requires transforms concepts into real numbers, called utility function.
The utility function $U$ such that $ l_i > l_j \xrightarrow{} U(l_i) > U (l_j)$, where $U$ is a function $U: C\xrightarrow{} \mathbbm{R}$.
In this problem, the ground-truth utility function ($U$) measures each concept’s importance based on users’ attitudes, defined as a regression learning problem.
According to $U$, we defined the ranking function, $R$, measuring the importance of each concept towards other concepts based on users’ attitude.
This is defined in Eq.~\ref{ch6_eq22}.
\begin{equation}
\label{ch6_eq22}
    R(l_{i})=\sum \mathbbm{1} \{U(l_{i})>U(l_{j})\} , \forall l_{i} , l_{j} \in {L},
\end{equation}
where $\mathbbm{1}$ is the indicator function.
The Bradley–Terry model~\cite{szummer2011semi} is a probability model widely used in preference learning.
Given a pair of individuals $l_{i}$ and $l_{j}$ drawn from some population, the model estimates the probability that the pairwise comparison $l_{i} > l_{j}$ is true.
Having $n$  observed preference items, the model approximates the ranking function $R$ by computing the maximum likelihood estimate in Eq.~\ref{ch6_eq23}.
\begin{equation}
\label{ch6_eq23}
\begin{split}
J_x(w)= & \sum_{i \in n}[p(l_{i},l_{j})log F(l_{i},l_{j};w)+
p(l_{j},l_{i})log F(l_{j},l_{i};w)],
\end{split}
\end{equation}
where $F(l)$ is the logistic function defined in Eq.~\ref{ch6_eq24}.

\begin{equation} \label{ch6_eq24}
 F(l_{i},l_{j};w)= \frac{1}{1+exp[U^{*}{(l_j;w)}-U^{*}{(l_i;w)}]}
\end{equation}

Here, $U^{*}$ is the approximation of $U$ parameterised by $w$, which can be learnt using different function approximation techniques.
In our problem, a linear regression model was designed for this purpose, defined as $U(l;w)=w^{T}\phi(l)$, where $\phi(l)$ is the representation feature vector of the concept $l$.
For any $l_i,l_j \in L$, the ranker prefers $l_i$ over $l_j$ if $w^{T}\phi(l_i)> w^{T}\phi(l_j)$.

By maximising the $J_x(w)$ in Eq.~\ref{ch6_eq23}, $w^{*} = arg max_w J_x(w)$, the resulting $w^{*}$ using stochastic gradient ascent optimisation will be used to estimate $U^{*}$, and consequently the approximated ranking function $R^{*}: C \xrightarrow[]{} \mathbbm{R}$.
Thus, Summation learns a ranking over concepts and uses the ranking to generate personalised summaries.

\subsubsection{Generating Personalised Summaries}
The summarisation task is to transform the input $d$ to the best summary in $Y(d)$ for the learnt preference ranking function.
This problem can be defined as a sequential decision-making problem, starting from the root, sequentially selecting concepts and adding them to a draft summary.Therefore, it can be defined as an MDP problem.

An MDP is a tuple $(S,A,R,T)$, where $S$ is the set of states, $A$ is the set of actions, $R(s,a)$  is the reward for performing an action ($a$) in a state ($s$), and $T$ is the set of terminal states.
In our problem, a state is a draft summary, and $A$ includes two types of action\textemdash either adding a new concept to the current draft summary or terminating the construction process if it reaches users’ limit size.
The reward function $R$ returns an evaluation score in one of the termination states or $0$ in other states.

A policy $\pi(s,a): S \times A \xrightarrow{} R$ in an MDP defines the selection of actions in state $s$.
The goal of RL algorithms is to learn a policy that maximises the accumulated reward.
The learnt policy trained on specific users’ interests is used on unseen data at the test time (in this problem to generate summaries in new and related topic documents).

We defined the reward as the summation of all concepts’ importance included in the summary.
A policy$\pi$  defines the strategy to add concepts to the draft summary to build a user’s desired summary. 
We defined $\pi$  as the probability of choosing a summary of $y$ among all possible summaries within the limit size using different hierarchy paths, $Y(d)$, denoted as $\pi(y)$. 
The expected reward of performing policy $\pi$, where $ R(y)$ is the reward for selecting summary $y$, is defined in Eq.~\ref{ch6_eq25}.

\begin{equation}
\label{ch6_eq25}
    R^{RL}(\pi|d)= \mathbbm{E}_{y \in Y(d)}R(y)= \sum_{y\in Y(d)} \pi(y)R(y)
\end{equation}

\begin{algorithm}[t]
  \begin{algorithmic}[1]
    \caption{Summation}
    \State \textbf{Input}: Document cluster d 
    \State \textbf{Output}: Summary (H) and optimal policy
    \label{alg1}
    \State\emph{\textbf{Organiser}}
        \State \quad $\textit{Concepts and Relations} \gets \textit{Concept and relations extraction (d)}$
         \State \quad  $\textit {H} \gets \textit{Hierarchical conceptual clustering (concepts)}$
    \State\emph{\textbf{Summariser (User preference learner)}}
        \State \quad  $\textit{User preferences} \gets \textit{Query pairs (user)}$
        \State  \quad $\textit{Ranker function} \gets \textit{Preference learner (user preferences)}$
    \State\emph{\textbf{Summariser (RL learner)}}
        \State \quad $\textit{Optimal policy} \gets \textit{Policy learner ( ranker function)}$
    \State Summary (H) and optimal 
    \end{algorithmic}
\end{algorithm}

The goal of MDP is to find the optimal policy $\pi^*$ that has the highest expected reward.
Therefore, the optimal policy, $\pi^*$, is the function that finds the desired summary for a given input based on user feedback (Eq.~\ref{ch6_eq26}).

\begin{equation}
\label{ch6_eq26}
\pi^* = arg max \hspace{2mm} R^{RL}(\pi|d) = arg max \sum_{y \in Y(d)}\pi(y) R(y)
\end{equation}

We also used the linear temporal difference algorithm to obtain $\pi^*$.
The process is explained in Algorithm~\ref{alg1}.

\subsection{Experiment}
\label{Evaluation}
This section presents the experimental set-up for implementing and assessing our summarisation model.
Summation was evaluated from different evaluation aspects, first from the organiser’s output, and then concerning the hierarchical concept map (H), which can be served individually to users as the structured summarised data.
Next, we evaluated H using both human and automatic evaluation techniques to answer the following questions:

\begin{itemize}
    \item Do users prefer hierarchical concept maps to explore new and complex topics?
    \item How much do users learn from a hierarchical concept map?
    \item How coherent is the produced hierarchical concept map?
    \item How informative are summaries in the form of a hierarchical concept map?
\end{itemize}
Personalised summaries generated on test data were also evaluated from various perspectives to analyse the effect of RL and preference learning.
This helped to assess the effect of different features in approximating the proposed preference learning-based system, as well as the performance of an RL algorithm and the information coverage in terms of ROUGE.

\subsubsection{Hierarchical Concept Map Evaluation}
To answer the questions in Sec.~\ref{Evaluation}, we performed three experiments. 
First, within the same limit size as the reference summaries, we compared the summaries produced by three models\textemdash using ExDos, which is a traditional approach; using a traditional hierarchical approach~\cite{christensen2014hierarchical}; and using a structured summarisation approach~\cite{falke2019automatic} on selected documents (with ROUGE-1 and ROUGE-2 scores based on the reference summaries).
The average ROUGE-1 for Summation was 0.65 and ROUGE-2 was 0.48.
The structured approach~\cite{falke2019automatic} showed similar performance with ROUGE-1 and ROUGE-2 at 0.65 and 0.45, respectively.
Meanwhile, traditional hierarchical approaches~\cite{christensen2014hierarchical} produced a ROUGE-1 of 0.27 and ROUGE-2 of 0.18.
In the same task, the percentage of covered unigrams and bigrams based on documents were also compared.
Both Summation and the structured approach covered approximately 4\% unigrams and 2\% bigrams, but dropped below 1\% in both cases when testing the hierarchical approaches.
In the third experiment, all competitors’ outputs were rated based on three measures, including usability in exploring new topics, level of informativeness, and coherency.
Summation’s rate for the first and second criteria was 96\% and 94\%, respectively.
However, it was 34\% for coherency.
We removed all concepts with low similarity to their parents based on a different threshold at each level.
After repeating the same experiment, and rate of coherency increased to 76\%.
 
\begin{figure}[t]
\centering
    \includegraphics[width=\linewidth]{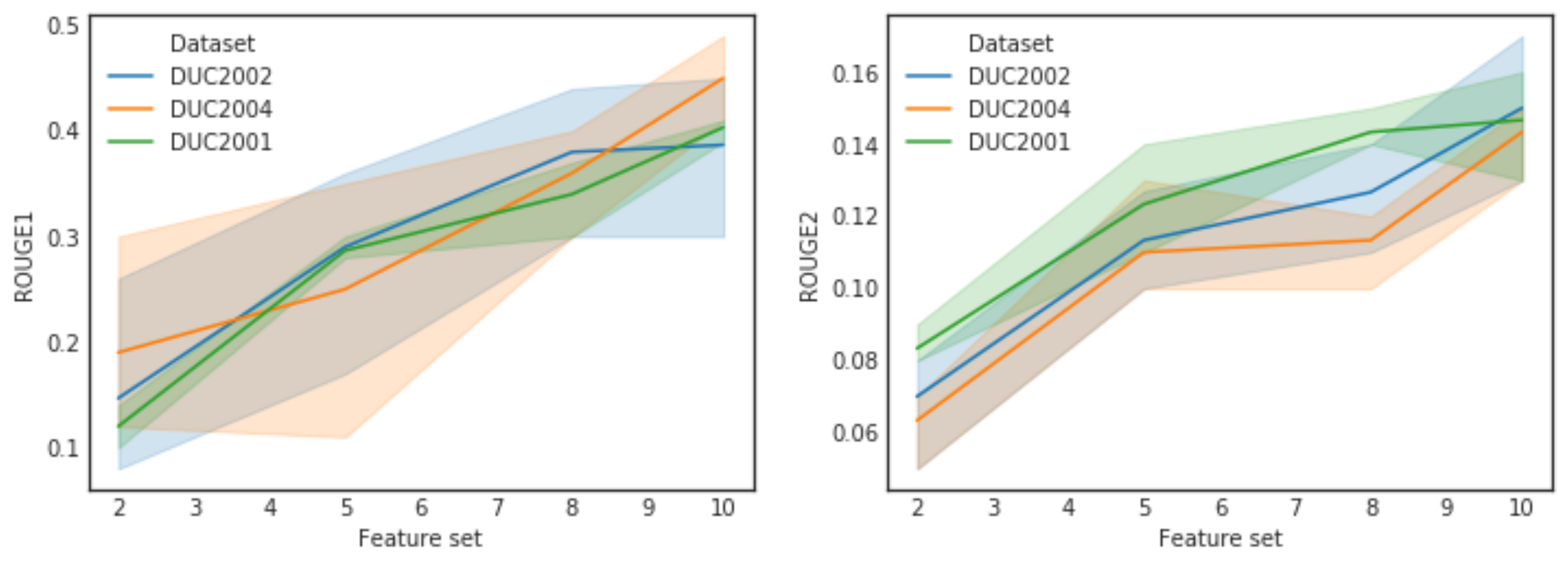}
    \caption{Evaluating different feature sets for estimating the ranker function.}
    \label{featureAnalysis}
\end{figure}

\subsubsection{Feature Analysis} 
\label{feature}
Before evaluating the effect of conceptual preference, it is important to explain the ground-truth concept ranker function ($U$) and the approximate function ($U^*$), indicating the importance of concepts.
To estimate the approximate function ($U^*$), we defined a linear model $U^*(c)=W^T\phi(c)$, where $\phi$ are the features.
To this end, a set of features (whose importance was validated in ExDos) was used, including surface-level and linguistic-level features.
Surface-level features include frequency-based features (TF-IDF, RIDF, gain and word co-occurrence), word-based features (upper-case words and signature words), similarity-based features (Word2Vec and Jaccard measure) and named entities.
Linguistic features are generated using semantic graphs and include the average weights of connected edges, the merge status of concepts as a binary feature, the number of concepts merged with a concept, and the number of concepts connected to the concept.
We defined different combinations of features with different sizes,$\{2,5,8,10\}$, starting from the most critical one.
Then, we repeated the experiments for 10 cluster documents.
We used the concepts included in the reference summary as preferences, and then evaluated the concept coverage in a concept map compared to the reference summaries using ROUGE-1 and ROUGE-2.
The results reported in Fig.~\ref{featureAnalysis} show that the model’s performance improved after adding more features.

\subsubsection{Summary Evaluation}
To avoid subjectivity in the evaluation process, we used the reference summaries as feedback.
The mentioned concepts that exist in reference summaries receive the maximum score by the ranked function.
We compared the summaries produced by three models, including the traditional approach (ExDos), a range of hierarchical approaches~\cite{christensen2014hierarchical}, and a structured summarisation approach~\cite{falke2019automatic}, each tested on randomly selected documents from three datasets using ROUGE-1, ROUGE-2 and ROUGE-L scores based on the references summaries.
The average results reported in Table~\ref{tab:data1} show the supremacy of Summation in selecting specific contents.

\begin{table}[t]
\centering
  \caption{Comparing Summation with Benchmark Datasets}
  \vspace{1mm}
  \label{tab:data1}
  \begin{tabular}{|l|c|c|c|}
    \hline
    Model &ROUGE-1 &ROUGE-2 &ROUGE-L\\
    \hline
    Traditional structured~\cite{falke2019automatic} &0.346  &0.090 &0.251\\
    \hline
    Traditional hierarchical~\cite{christensen2014hierarchical}  &0.211  &0.013 &0.149\\
    \hline 
    Summation &0.731 &0.651 &0.681\\
 \hline
\end{tabular}
\end{table}

\section{Summary}
This chapter provided three solutions to address the problem of personalised summarisation.
First, we proposed an interactive and personalised MDS approach using user feedback.
This included the selection or rejection of concepts, defining the importance of a concept, and users’ confidence level.
We also proposed a summary recommendation framework that interactively learns how to generate personalised summaries based on that feedback.
We provided two structures, first to predict extractive summaries (sentence based), and second to make structured summaries.
We empirically examined the validity of the proposed models using simulated user feedback.
Results verified that the proposed frameworks show promising results in terms of ROUGE score as well as human evaluation. 
The results also showed that user feedback could be integrated into intelligent systems to help them obtain their desired information.

%% file: ch_7/Application.tex
\chapter{Summarisation Applications}
\label{ch_7}
This chapter proposes solutions to the information overload problem across different domains using summarisation. We address three specific applications including:
\begin{itemize}
    \item the use of summarisation in detecting anomalies in network traffic data
    \item the application of summarisation in healthcare analytics problems
    \item the use of summarisation to narrate business process data.
\end{itemize}
Each section discusses each problem, the state-of-the-art approaches in each domain, the proposed solutions, and the experiments conducted.

\section{Detecting Anomalies Through Summarisation}
Monitoring network traffic data to detect hidden patterns of anomalies is a challenging and time-consuming task that requires high computing resources.
To this end, an appropriate summarisation technique is important, as it can effectively substitute for original data.
A network summary can reveal what is happening in a network and managing the network instantly.
However, one drawback is that the summarised data may remove existing anomalies.
Therefore, it is vital to create a summary that can reflect the same pattern as the original data.
For example, the summary should still give insight into most browsing websites, regularly used applications, and incoming traffic patterns.
Three scenarios in which effective summarisation can aid traffic data collection~\cite{hoplaros2014data} include:
\begin{itemize}
    \item providing network administrators with an overview of what is happening in a network
    \item use of summarised network traffic data as input in anomaly detection algorithms to reduce computing costs
    \item summarising intrusion detection alarms, thus, facilitating an administrator’s duties.
\end{itemize}

A concise representation of the data helps both the administrator and the analysis algorithms in all mentioned scenarios.
Different data summarisation techniques are designed for other applications such as transactional data or stream data~\cite{ahmed2019data}, which can also be applied to traffic data.
However, this has some drawbacks regarding anomaly detection:
\begin{itemize}
    \item Clustering is the most used summarisation approach.
    Here, the data centre is considered the summarised data output. 
    The problem is that the cluster centers might not be part of the original data.
    \item Detecting frequent item sets is another approach that only captures frequent items in a summary.
    Therefore, they ignore infrequent anomalies.perform well on summaries, as they do not contain any anomalies.
    \item Semantics-based techniques do not keep the same samples in the summarised data.
    \item Statistical-based techniques such as sampling do not ensure presence of anomalies in summaries since they use a sampling-based summarisation technique.
\end{itemize}

Evidently, not all summarisation approaches are appropriate for anomaly detection.
There is great need for an efficient network traffic summarisation (NTS) technique to ensure that automated summaries more closely resemble the original network traffic data.
In this context, summarisation aims to synopsise original data with interesting patterns, like anomalies, and normal data with the same distribution pattern.

We propose an intelligent summarisation approach for identifying hidden anomalies, called INSIDENT.
The proposed approach guarantees to keep the original data distribution in summarised data, and employs similar ones to ExDos (see Ch.~\ref{ch_4}).
As such, we investigate the adaptation of clustering and KNN algorithms to create a summary.

INSIDENT is an intelligent summarisation approach for detecting anomalies in network traffic datasets, which guarantees the preservation of original data distribution.
The proposed algorithm is used in two scenarios\textemdash (i) as a pre-processing approach for performing anomaly detection, and (ii) to detect anomalies in supervised problems, since the algorithm reveals the hidden structure of data.
The proposed summarisation technique can also be used in various domains in which big data requires mining for interesting and relevant patterns.

\subsection{Related Work}
\label{related}
Anomaly detection techniques perform poorly when the data size increases due to increased false alarms and high computational cost.
Therefore, summarisation can be a beneficial alternative tool.
However, existing summarisation techniques cannot accurately represent the rare anomalies in a dataset.
This section presents the related work on traffic data summarisation, along with anomaly detection techniques.
For anomaly detection purposes, a good summary should be representative of all samples in the original dataset.

\subsubsection{Network Analysis Tools}
Different network analysis tools summarise network traffic data, such as traffic flow analysis tools, flow-tools, network visualisation tools, and network monitoring tools~\cite{ahmed2019intelligent}.
Each produces a graphical report using different measurements, such as network bandwidth or latency.
However, these tools only characterise and aggregate traffic samples regarding a single feature, such as the source address or protocol.
As a result, they are suitable only to extract insights and not for further processing tasks such as anomaly detection.
Besides, the purpose of a summary is to produce an accurate description of the network’s traffic patterns.
Consequently, the summarisation technique should identify traffic patterns according to any desired combinations of features efficiently.

\subsubsection{Statistical Approaches}
Statistical approaches aim to determine the statistical distribution of data that could approximate the original data pattern.
Sampling is a common technique in this category in which a sample is a subset of the dataset.
Sampling techniques include simple random sampling, systematic sampling, cluster random sampling, stratified random sampling, and multistage random sampling~\cite{cochran1977sampling,ghodratnama2009innovative}.
However, summarised data using sampling is under the threat of removing anomalies.
To solve this problem, one recent study~\cite{ahmed2019intelligent} proposed a sampling-based summarisation technique called SUCh, employing sampling and the modified Chernoff bound technique to add anomalous instances in summary.
SUCh is more computationally effective and is also capable of identifying rare anomalies.
However, an essential aspect of summarisation is describing all different traffic behaviour patterns.
Although SUCh ensures the presence of anomalies, it ignores other types of traffic data, as it focuses only on anomalous data.

\subsubsection{Machine Learning Approaches}
Supervised and unsupervised learning have been widely used for knowledge discovery.
Two common machine learning algorithms used in summarising network traffic data are frequent item sets and clustering.

Frequent item sets are sets of items that appear more frequently than other samples. 
Different algorithms are used to detect frequent item sets~\cite{chandola2007summarization}; however, they are most appropriate for detecting frequent items and not rare anomalies.

The two main clustering-based algorithms for network traffic data summarisation include centroid-based and feature-wise intersection clustering algorithms.
In centroid-based summarisation, after clustering samples, centroids are used to form a summary.
Different variations of the k-means algorithm are also to handle high-dimensional data~\cite{ghodratnama2015efficient,wendel2005scalable}.
In a feature-wise intersection-based summarisation algorithm, the summary is generated from each cluster utilizing the feature-wise intersection of the samples after clustering\cite{chandola2007summarization,hoplaros2014data}.
Then, all clusters' summaries are linked to produce the output.
This approach is best suited to datasets with identical attribute values and, therefore, are not appropriate for detecting rare anomalies.

\subsubsection{Semantic-based Approaches}
Semantic-based approaches are not suitable for anomaly detection since the produced summary is not part of the original data.
Examples include linguistic summaries, which are based on fuzzy algorithms. 
These approaches produce text representations that explain essential aspect to enhance human understanding of network traffic summaries~\cite{pouzols2011summarization}.
Attribute-oriented induction is another semantic-based approach that aims to express data in a brief and general manner~\cite{han199616}.
This induction technique is a generalisation process that abstracts a large dataset from a low conceptual level to a relatively high conceptual level.
Other semantic-based approaches include fascicles~\cite{jagadish1999semantic}, using association rules and perform lossy semantic compression.
SPARTAN is another semantic-based summarisation technique~\cite{babu2001spartan} that generalises the fascicles method.

\subsubsection{Anomaly Detection Techniques}
Detecting anomalies is a vital task, aiming to identify anomalous or abnormal data.
Anomalies are any patterns in original data that do not follow the well-defined nature of regular patterns, indicating notable but unusual events that may negatively affect the system.
Therefore, they require prompt critical actions. 
An anomaly can be categorised as a~\cite{ahmed2015investigation}:
\begin{itemize}
    \item point anomaly (when a sample differs from the regular pattern)
    \item contextual (or conditional) anomaly (when a sample behaves anomalously in a special context)
    \item collective anomaly (when a collection of samples behave anomalously).
\end{itemize}

Different supervised, unsupervised and semi-supervised approaches have been proposed for this purpose.
These techniques include classification-based network anomaly detection algorithms such as support vector machines~\cite{balabine2017method}, Bayesian network models~\cite{kruegel2003bayesian}, neural networks~\cite{poojitha2010intrusion}, and rule-based approaches~\cite{yang2013rule}.
Statistical anomaly detection techniques include a mixture model, signal-processing technique~\cite{thottan2003anomaly}, and principal component analysis~\cite{shyu2003novel}.
Other categories include information theory-based and clustering-based methods~\cite{ahmed2019data}.

\subsection{Proposed Approach (INSIDENT)}
\label{proposed}
This section discusses the proposed methodology. We first define the problem and then discuss the algorithm.

\subsubsection{Problem Definition}
$x_i$  is a sample vector and $X = [x_1, x_2,..., x_N]$ is the traffic data consisting of $N$ sample, where $x_i \in R^d$, and d denotes the number of features.
$K$  is the number of clusters, and cluster centroids are denoted by $c$.
Further, $x_=$ is the closest similar sample to $x$, and $s_{\neq}$ is the closest different sample.
An example of network traffic data with few attributes is reported in Table~\ref{tab:sample}.
The goal is to find a cluster of similar samples and representatives for each cluster as the summary $S$, where the same distribution is retained but smaller in size.

\begin{table}[t]
\centering
\caption{Example of Network Traffic Samples}
\label{tab:sample}
    \begin{tabular}{|l|c|c|c|c|}
        \hline
        Source IP & Source port & Destination IP & Destination port  & Protocol \\
        \hline
        192.168.5.10 & 1234 & 192.168.1.1 & 80 & TCP\\
        \hline
        192.168.5.12 & 4565 & 192.168.1.2 & 20 & TCP\\
        \hline
        192.168.5.10 & 20 & 192.168.28.80 & 119 & HTTP\\
        \hline
        192.168.5.10 & 70 & 192.168.1.1 & 50 & TCP\\
        \hline
        211.204.12.10 & 31 & 192.168.28.80 & 119 & HTTP\\
        \hline
        192.168.5.1 & 3214  & 192.168.1.2 & 86 & TCP\\
        \hline
    \end{tabular}
\end{table}

\subsubsection{Methodology}
Previous approaches used different clustering or sampling algorithms to summarise data.
However, there is no guarantee that the summarised data has the same distribution as the original data, and, therefore, cannot substitute the original data.
Thus, we employed the same procedure as in ExDos and investigated the adaptation of clustering and the KNN algorithm to understand the data’s underlying structure.
For this reason, the error rate of the nearest neighbour classifier in each cluster was minimised by locally weighting features in each cluster.
INSIDENT transforms the feature space into a new feature space by weighting features separately in each cluster, where outliers are more easily recognised in the new feature space.
To this end, the weighted Euclidean distance is used where the distance between vectors $x$ and $x_i$ is defined as in Eq.\ref{ch7_eq1}.
\begin{eqnarray}
\label{ch7_eq1}
d_w(x ,x_i) = \sqrt{\sum_{j=1}^{d} w_{j} (x_j-x_{ij})^2},
\end{eqnarray}
and $w_{j}$ is the corresponding weight of $j-th$ feature.
The weights are arranged in a $d\times K$ weight matrix $W=\{w_{ij}, 1\leq i\leq d, 1 \leq j \leq K\}$, where $d$ is the number of features and $K$ is the number of clusters. 
Therefore, for each cluster, there is a vector of weights corresponding to each feature, representing the importance of each feature in each cluster.
The objective function is designed to minimise the error of 1NN in each cluster by regulating the weights of each feature and consequently clustering centres.
To estimate the error of 1NN, the approximation function in Eq.\ref{ch7_eq2} was used~\cite{paredes2006learning}.

\begin{equation}
\label{ch7_eq2}
J(\textbf{W})=\frac{1}{N}\sum_{s\in XS}^{}S_{\beta}(\frac{d_w(x,x_=)}{d_w(x,x_{\neq})}),
\end{equation}

where the sample $x_=$ is the nearest similar sample, and the sample $x_{\neq}$ is the closest different sample to the input sample $x$. 
Respectively, $d_w$ is the weighted Euclidean distance and $S_{\beta}$ is the sigmoid function, defined in Eq.~\ref{ch7_eq3}.
\begin{equation}
\label{ch7_eq3}
S_{\beta}(z)=(\frac{1}{1+{e}^{\beta(1-z)}})
\end{equation}

The objective function of k-means, which aims to minimise the errors in each cluster, is defined in Eq.~\ref{ch7_eq4}.

\begin{equation}
\label{ch7_eq4}
J(\textbf{W},\textbf{C})=\sum_{k=1}^{K}\sum_{i=1}^{\mid N_K \mid} d_{W_{K}}^2 (x_i,c_K)
\end{equation}

Thus, the overall objective function is defined in Eq.\ref{ch7_eq5}.

\begin{equation}
\label{ch7_eq5}
    J(\textbf{W},\textbf{C})=  (\sum_{k=1}^{K}\sum_{i=1}^{\mid N_K \mid} d_{W_{}}^2 (s_i,c_K) + \frac{1}{N}\sum_{k=1}^{K}\sum_{i=1}^{\mid N_K \mid}S_{\beta}(\frac{d_w(x,x=)}{d_w(x,x_{\neq})})),
\end{equation}
where the first term is the objective function of k-means, and the second term is the summation of the classification errors over the $K$ clusters.

Two parameters are optimised in this objective function. 
The first is the weights matrix.
The feature-dependent weights associated with the sample $x_=$ are trained to more closely match $x$ while pushing the sample $x_{\neq}$ further from $x$.
Then, the cluster centroid update is based on the learnt weighted distance.
Since this function is differentiable, we can analytically use gradient descent to estimate the matrix $W$, guaranteeing convergence.
The iterative optimisation of a learning parameter like w is given in Eq.~\ref{ch7_eq6}. 
\begin{equation}
\label{ch7_eq6}
W^{t+1}=W^{t}-\alpha(\frac{J(\textbf{W},\textbf{C})}{\delta(W)})
\end{equation}

To simplify the formula, the function $R(x)$ is defined in Eq.~\ref{ch7_eq7}~\cite{paredes2006learning}

\begin{equation}
\label{ch7_eq7}
R_w(x_i)=(\frac{d_w(x_i,x_{i,=})}{d_w(x_i,x_{i,\neq})})
\end{equation}

The partial derivative of $J(W,C)$ with respect to $W$ is then calculated in Eq.~\ref{ch7_eq8}.

\begin{equation}
\label{ch7_eq8}
    {\frac{\delta J(\textbf{W},\textbf{K})}{\delta W_K}} \cong \sum_{i=1}^{\mid N_K \mid} 2W_K \odot (x_i-C_K)^2+\frac{1}{N}\sum_{i=1}^{\mid N_K \mid} S_{\beta}^{'}(R(x_i))\frac{\delta R(x_i)}{\delta W_k}
\end{equation}

where $\odot$ is the inner product and $\frac{\delta R(x_i)}{\delta W_K}$ is defined in Eq.~\ref{ch7_eq9}.

\begin{equation}
\label{ch7_eq9}
    \frac{\delta R(s_i)}{\delta W_K}=\frac{1}{{}d_{W_K}^2}(x_i,x_{i,\neq})({\frac{1}{R(x_i)}W_K\odot (x_i-x_{i,=})^2}-R(x_i)W_K\odot{(x_i-x_{i,\neq})^2)}
\end{equation}

The derivative of $S_{\beta}(z)$ is defined in Eq.\ref{ch7_eq10}.

\begin{equation}
\label{ch7_eq10}
S_{\beta}(z){'}=\frac{\delta S_{\beta}(z)}{\delta z}=\frac{\beta e^{\beta(1-z)}}{(1+e^{\beta(1-z)})^2}
\end{equation}

The partial derivative of $J(\textbf{W},\textbf{C})$ with respect to $C$, is calculated as in Eq.~\ref{cch7_eq11}.

\begin{equation}
\label{cch7_eq11}
\frac{J(\textbf{W},\textbf{C})}{\delta C_k} \cong \sum_{i=1}^{\mid N_k \mid} -2W_k^2 \odot(x_i-C_k)
\end{equation}

Since we need to optimise the weight of features for each cluster’s samples, along with the centre of the clusters, we first updated $W$ in each cluster and then updated $C$ (centre of clusters).
The INSIDENT algorithm is depicted in Algorithm~\ref{alg1} for added clarification.
Since the algorithm performs in an iterative process using gradient descent, the simplest clustering (k-means) and (KNN) algorithms were used for efficiency.
However, k-means is one of the most reliable and widely used clustering algorithms.
So too is the KNN, which has been successfully used in many pattern-recognition applications~\cite{anava2016k}. 
Thus, similar samples are close to each other in the new feature space, meaning both ‘point’ and ‘contextual’ (or conditional) anomalies are also easily detectable.
In the case of collective anomalies, we selected the number of each cluster’s representative based on its size to maintain the same distribution pattern as the original data.

\subsection{Experiments and Evaluation}
\label{evaluation}
In this section, the dataset, evaluation method and performance of INSIDENT are explained and compared with existing state-of-the-art approaches.

\subsubsection{Data Set}
Experiments on six benchmark datasets were performed. 
The details of datasets and the distribution of normal and anomalous samples in each dataset are reported in Table~\ref{tab:dataset}.
KDD1999 contains collective anomalies, whereas the other five datasets contain only rare anomalies.
These rare anomalous datasets are from the SCADA network, including real SCADA (WTP), simulated anomalies (Sim1 and Sim2), and injected anomalies (MI and MO).

\subsubsection{Evaluation Metrics}
To evaluate the network traffic summary, we explain two widely used summary evaluation metrics: conciseness and information loss~\cite{ahmed2015efficient}.
First, conciseness denotes the size of the summary, which influences the quality of the output.
At the same time, it is important to create a summary that can reflect the underlying data patterns.
Conciseness is defined as the ratio of the input dataset size ($N$) and the summarised dataset size ($S$), defined in Eq.~\ref{ch7_eq12}.

\begin{equation}
\label{ch7_eq12}
    Conciseness=\frac{N}{S}
\end{equation}

Second, information loss is defined as the ratio of the number of samples not present by samples present in a summary, defined in Eq.~\ref{ch7_eq13}.

\begin{equation}
\label{ch7_eq13}
    Information Loss=\frac{L}{T},
\end{equation}
where T is the number of unique samples represented by the summary, and L defines the number of samples not presented in the summary.

\begin{table}[t]
\centering
\caption{Dataset Description}
\label{tab:dataset}
    \begin{tabular}{|l|c|c|c|}
        \hline
        Dataset & Sample number &  Normal percentage & Anomalies percentage\\
        \hline
        KDD1999 &494020 &19.69 &80.310\\
        \hline
        WTP &527 &97.34 &2.66\\
        \hline
        MI &4690 &97.86 &2.14\\
        \hline
        MO &4690 &98.76 &1.24\\
        \hline
        Sim1 &10501 &99.02 &0.98\\
        \hline
        Sim2 &10501 &99.04 &0.96\\
        \hline
    \end{tabular}
\end{table}

To evaluate the performance of the anomaly detection algorithms used in supervised approaches, three measures—accuracy, recall and F1—were used.
Before we define these measures, four values need clarification~\cite{ahmed2015investigation}.
These are:
\begin{itemize}
    \item true positive (TP), or the number of anomalies correctly identified as anomalous
    \item false positive (FP), or the number of normal data incorrectly identified as an anomaly
    \item true negative (TN), or the number of normal data correctly identified as normal
    \item false negative (FN), or the number of anomalies incorrectly identified as normal.
\end{itemize}

Based on these definitions, evaluation metrics are defined in Eq.\ref{ch7_eq14}, \ref{ch7_eq15} and \ref{ch7_eq16}.

\begin{equation}
\label{ch7_eq14}
    Accuracy=\frac{TP+TN}{TP+TN+FP+FN}
\end{equation}

\begin{equation}
\label{ch7_eq15}
    Recall=\frac{TP}{TP+FN}
\end{equation}

\begin{equation}
\label{ch7_eq16}
    F1=\frac{2TP}{2TP+FP+FN}
\end{equation}

\subsubsection{Result Analysis}
This section discusses the INSIDENT performance along with the anomaly detection results.

\begin{table}[t]
\centering
\caption{Real SCADA Dataset (WTP) Results}
\label{tab:WTP}
    \begin{tabular}{|l|c|c|c|}
        \hline
        Model & WTP-Recall &  WTP-Accuracy & WTP-F1\\
        \hline
        KNN & 85.71 &  97.39 & 85.71\\
        \hline
        LOF & 78.57 &  97.38 & 78.57\\
        \hline
        COF & 57.14 &  97.35 & 57.14\\
        \hline
        LOCI & 85.71 &  97.39 & 85.71\\
        \hline
        LoOP & 42.85 &  97.33 & 42.85\\
        \hline
        INFLO & 57.14 &  97.35 & 57.14\\
        \hline
         CBLOF & 92.85 &  97.40 & 92.85\\
        \hline
         LDCOF & 85.71 &  97.39 & 85.71\\
        \hline
         CMGOS & 57.14 &  97.35 & 57.14\\
        \hline
         HBOS & 28.57 &  97.32 & 28.57\\
        \hline
         LIBSVM & 85.71 &  97.39 &  85.71\\
        \hline
        \textbf{INSIDENT}&94.87 &97.91 &94.87\\
        \hline
    \end{tabular}
\end{table}

\textbf{\textit{Anomaly detection evaluation.}}
The baseline algorithms include nearest neighbour-based algorithms( KNN~\cite{ramaswamy2000efficient}, local outlier factor [LOF]~\cite{breunig2000lof}, connectivity-based outlier factor [COF]~\cite{tang2007capabilities}, local correlation integral [LOCI]~\cite{papadimitriou2003loci}, local outlier probability [LoOP]~\cite{kriegel2009loop}, INFLO~\cite{jin2006ranking}), clustering-based approaches (cluster-based LOF [CBLOF]~\cite{he2003discovering}, local density cluster-based outlier factor [LDCOF]~\cite{amer2012nearest}, clustering-based multivariate Gaussian outlier score [CMGOS]~\cite{amer2012nearest}), and statistical approaches (histogram-based outlier score [HBOS], library for support vector machines [LIBSVM]~\cite{amer2013enhancing}. 

These approaches are compared with INSIDENT on different variations of the SCADA dataset, including WTP, MI, MO, and Sim1 and Sim2; their values are reported by Mohiuddin et al.~\cite{ahmed2015investigation}.
Results are reported in Table~\ref{tab:WTP},~\ref{tab:SIM} and ~\ref{tab:MI}.

Table~\ref{tab:WTP} shows that for the real SCADA dataset (WTP), INSIDENT has higher values.
The clustering-based anomaly detection technique (CBLOF) performs next best, and third is the nearest neighbour-based approach.
This is expected and proves that the combination of clustering and KNN can, indeed, perform better.
Conversely, the statistical-based approach (HBOS) did not perform well.

Table~\ref{tab:SIM} displays the results on simulated datasets (Sim1 and Sim2). 
LIBSVM has better recall than the other approaches, and INSIDENT performs second best.
Finally, the clustering-based approaches do not perform well on the simulated datasets.

For the datasets with injected anomalies (MI, MO), INSIDENT, along with the clustering-based approaches, are the best, followed by nearest neighbour-based approaches.
It is interesting to observe that the recall and F1 values are identical for all the anomaly detection techniques. 
This is because the top $N$ anomalies match the actual anomalies in the dataset, meaning the recall and F1 scores are constantly the same.

\begin{table}
\centering
\caption{Simulated SCADA Datasets Results (Sim1 and Sim2)}
\label{tab:SIM}
    \begin{tabular}{|l|c|c|c|c|c|c|}
        \hline
        Model & Sim1-Recall &  Sim1-Acc & Sim1-F1 & Sim2-Recall &  Sim2-Acc & Sim2-F1\\
        \hline
        KNN & 64.7&  99.03& 64.7 &63 &99.05&63\\
        \hline
        LOF & 0&  99.01 &0& 0 &99.03&0\\
        \hline
        COF & 0 &  99.01 & 0& 2&99.03&2\\
        \hline
        LOCI &0 &  99.01 & 0 & 0&99.03&0\\
        \hline
        LoOP & 0.98 &  99.01 & 0.98& 0&99.03&0\\
        \hline
        INFLO & 0 &  99.01 & 0& 0&99.03&0\\
        \hline
         CBLOF & 0 &  99.01 & 0& 0&99.03&0\\
        \hline
         LDCOF & 0 &  99.01 & 0& 0 &99.03&0\\
        \hline
         CMGOS & 18.62 & 99.02 & 18.62& 97&99.05&97\\
        \hline
         HBOS & 30.39 &  99.02 & 30.39 & 27&99.04&6\\
        \hline
         LIBSVM & 74.50 &  99.03 &  74.50 & 68&99.05&68\\
        \hline
        \textbf{INSIDENT} &72.13 &99.07 &72.13 &78.21 &99.05 &78.21\\
        \hline
    \end{tabular}
\end{table}

\begin{table}
\centering
\caption{Simulated SCADA Datasets with Injected Anomalies Results (MI and MO)}
\label{tab:MI}
    \begin{tabular}{|l|c|c|c|c|c|c|}
        \hline
        Model & MI-Recall &  MI-Acc & MI-F1 &  MO-Recall &   MO-Acc &  MO-F1\\
        \hline
        KNN &96 &97.09 &96 &91.37 &98.77 &91.37\\
        \hline
        LOF &38.33 &97.43 &38.33 &55.17 &98.76 &55.17\\
        \hline
        COF &9 &97.82 &9 &25.86 &98.75 &25.86\\
        \hline
        LOCI &91 &97.9 &91 &84.48 &98.77 &84.48\\
        \hline
        LoOP &10 &97.83 &10 &27.58 &98.75 &27.58\\
        \hline
        INFLO &12 &97.83 &12 &43.1 0&98.76 &43.10\\
        \hline
         CBLOF &24 &97.84 &24 &63.79 &98.76 &63.79\\
        \hline
         LDCOF &100 &97.91 &100 &63.79 &98.76 &63.79\\
        \hline
         CMGOS &100 &97.91 &100 &50 &98.76 &50\\
        \hline
         HBOS &98 &97.91 &98 &65.51 &98.76 &65.51\\
        \hline
         LIBSVM &86 &97.9 &86 &91.37 &98.77 &91.37\\
        \hline
        \textbf{INSIDENT} &100 &98.76 &100 &94.21 &99.04 &94.21\\
        \hline
    \end{tabular}
\end{table}

\begin{table}
\centering
\caption{Comparing the Distribution of Anomalies in Summaries and Original Data}
\label{tab:summary}
    \begin{tabular}{|l|c|c|c|}
        \hline
        Dataset &Original data &SUCh Alg. &INSIDENT\\
        \hline
        WTP &2.66 &N/A & 2.33\\
        \hline
        MI &2.14 &2.61 &2.16\\
        \hline
        MO &1.24 &1.46 &1.21\\
        \hline
        Sim1 &0.98 &1.04 &1.01\\
        \hline
        Sim2 &0.96 &0.94 &0.97\\
        \hline
    \end{tabular}
\end{table}

\begin{figure*}
    \centerline{\includegraphics[width=0.65\textwidth]{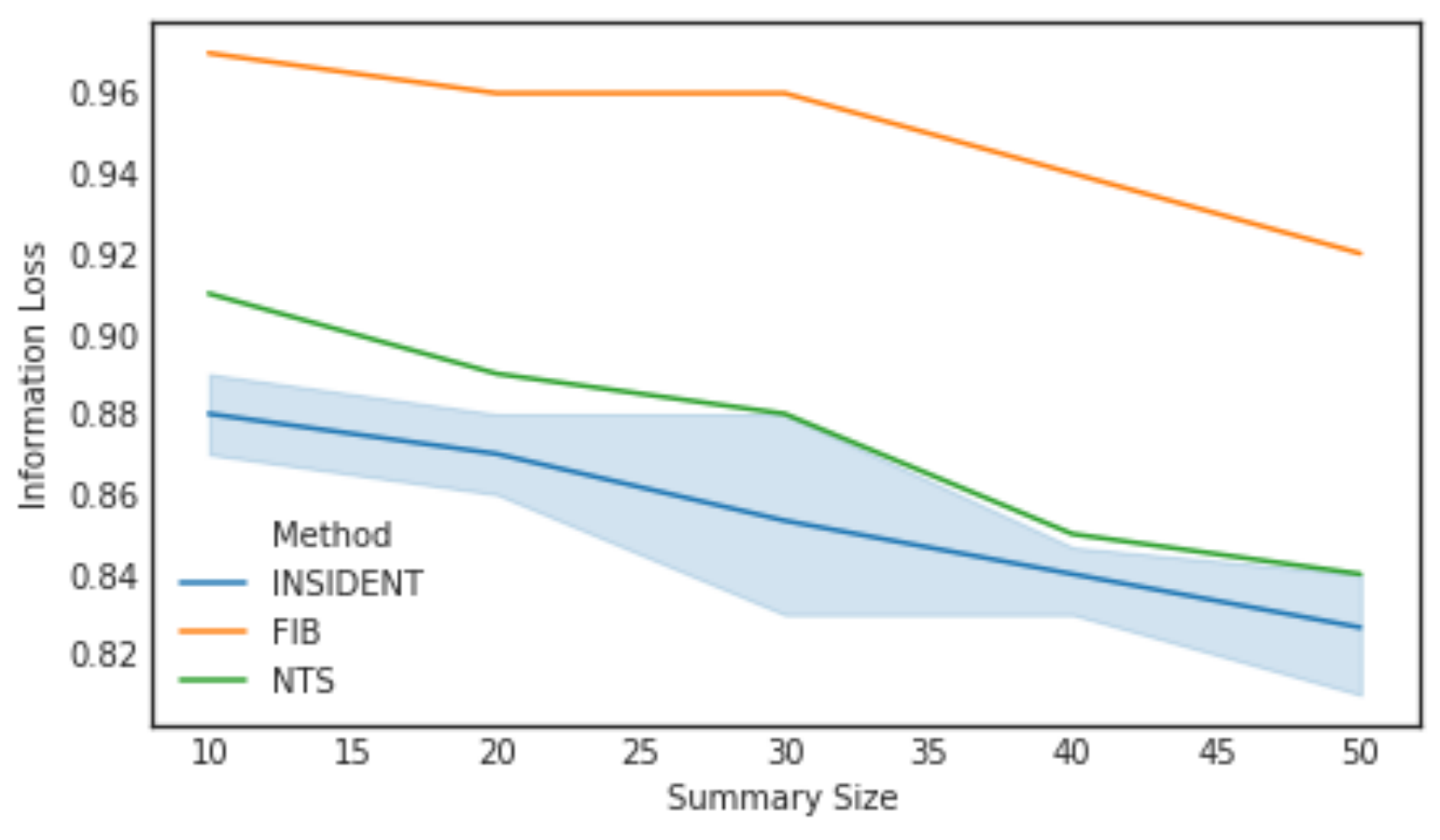}}
    \caption{Result of comparing information loss based on different summary sizes.
    FIB: forwarding information base; NTS: network traffic summarisation}
    \label{fig:infoloss}
\end{figure*}

\textbf{Network Traffic Summarization Evaluation.}
The KDD dataset was used for summarisation evaluation.
Summarisation size, which defines conciseness, is considered a constraint in summarisation algorithms.
When the summary is small, it has maximum information loss.
Conversely, when conciseness is small, the summary contains the whole dataset and, thus, incurs no information loss. 
Therefore, information loss and conciseness are orthogonal parameters.
We used five different summary sizes, with information loss measured for each summary size.
In practise, a network manager/analyst decides the summary size based on the network.
The results are compared with NTS and feature-wise intersection-based summarisation approaches~\cite{ahmed2014novel}.
Since our algorithm is based on k-means, we tested it three times with different initial points for each summary size.
Results are depicted in Fig.~\ref{fig:infoloss}. 
The percentage of anomalies compared with SUCh~\cite{ahmed2019intelligent} is reported in Tablee~\ref{tab:summary}, proving that INSIDENT effectively preserves the percentage of anomalies in generated summaries.

\section{Summarisation in Healthcare Analytics}
The ever-increasing amount of available text data makes it challenging for humans to extract only what they need.
This problem is even more vital in the medical domain, where accessing up-to-date information is essential.
Biomedical information is accessible in various forms.
Examples include biomedical literature), medical records, multimedia documents, information from the web, and a host of other diverse formats such as journal articles and patient records used to progress the latest advances in a particular field of study.
Biomedical literature helps clinicians and researchers to assess, develop and validate their proposed hypotheses and conduct new experiments~\cite{fleuren2015application}.
However, the extensiveness of available resources has introduced greater need for superior information extraction~\cite{DBLP:journals/computing/BeheshtiBVRMW17} and knowledge discovery techniques in academia and medical research. 
Intelligent content summarisation approaches are helpful tools in situations where an overview of a set of documents is needed.
However, in the medical context, the generated summary must be tailored to suit two different user preference types: physicians and patients.

State-of-the-art text summarisation approaches in the biomedical domain cover a wide range of subfields, including biomedical literature~\cite{moradi2017quantifying,moradi2018different,moradi2018cibs}, treatment document summarisation~\cite{zhang2011degree}, evidence-based medical care~\cite{fiszman2009automatic}, drug information~\cite{fiszman2006summarizing},clinical decision support and notes~\cite{morid2016classification,moen2016comparison}, and electronic health records~\cite{pivovarov2015automated}.
However, traditional summarisation approaches are incapable of considering users’ specific needs, uncertainty and interactivity within the information-seeking process.

We propose a novel embedding method, called Summary2vec, in which each summary is presented within a novel architecture by a fixed-length vector covering various aspects of the information space.
Summary2vec is remedial to design automatic services for various analytic purposes that require information-seeking activities.
Specifically, we leverage Summary2vec to produce a hierarchical summarisation structure that helps users better navigate a textual hierarchy and gain more in-depth information upon request.
Then, instead of providing a short and static summary, we present an interactive summarisation approach that engages users in the summarisation process.
The unique contribution of this section is as follows:
\begin{itemize}
    \item We introduce and formalise a theoretically grounded method for embedding summaries, called Summary2vec, and propose a novel neural network architecture.
    Vectorised summaries facilitate exploration of various aspects of the information space. 
    Therefore, the method could be used as inputs to a wide range of machine learning models to solve real-world problems. 
    It also helps in explaining to end users the decisions made by the system.
    \item We propose a hierarchical clustering approach and apply Summary2vec to produce a hierarchical summary structure that can help users navigate and explore different aspects of the information space.
    \item We provide evidence in the form of experiments, in which the model is trained and applied for personalised summarisation.
\end{itemize}

\subsection{Related Work}
There are many practical scenarios where people are facing an extensive collection of text with a particular goal. 
Recently, summarisation application has gained interest among the medical research community due to the tremendous increase of available information.

Physicians and researchers must sift through an increasing amount of published journals, conference proceedings, medical websites, electronic medical records and portals on the internet to pinpoint what they need.
The more documents which one has to face, the more complex the task becomes as the result of information overload. 
However, information is only valuable when it becomes reduced and meaningful, and tailored to the user’s interests.
In these scenarios, information-seeking activities go beyond fact-checking and aim at expanding one’s insight and discovering conceptual knowledge boundaries~\cite{Exploratory1121979}.
As hardware advances and cloud usage empowers immense information processing, document summarisation becomes an essential tool to ensure that users benefit from the information availability.

Document summarisation creates a short, limited-size text, representing the important contents~\cite{nenkova2011automatic}.
Different techniques including extractive and abstractive approaches, single-document summarisation or MDS, and query-focused summarisation are proposed.
This section discusses the key problems of text summarisation in the medical domain.

\subsubsection{Extractive Document Summarisation in the Medical Domain}
MiTAP, or MITRE text and audio processing, is an extractive approach proposed for single and multiple documents~\cite{damianos2002real,mckeown2003leveraging}.
MiTAP intends to monitor infectious disease outbreaks and other biological threats according to epidemiological reports, news wire feeds, email and online news in various languages.

The process begins by filtering and normalising the collected information. 
MiTAP leverages human-created rules to determine the data, source, article title, and body used to select paragraph, sentence, and part-of-speech tags.
Then, a named entity recogniser is used to detect dates, diseases and victim descriptions using human-created rules.
Finally, the document is processed with WebSumm~\cite{mani1999summarizing} to generate summaries.

MUSI, multilingual summarisation for the internet,~\cite{lenci2002multilingual} is a cross-lingual query-based summarisation system that uses Italian and English articles extracted from the Journal of Anaesthesiology to create summaries in French and German using features such as cue phrases, query words and the position of the sentences to form the summary.

To test their approach, Johnson et al.~\cite{johnson2002modeling} used articles published by the Journal of the American Medical Association to rank a series of extracted sentences according to each document’s cluster signature.
The approach takes medical documents filtered by a query as input and clusters them. 
A summary is produced by pairing the cluster signature to sentences to be summarised.
Elsewhere, Kan et al.~\cite{kan2001applying,kan2001domain} proposed an approach called Centrifuser, which is the summarisation engine of the PERSIVAL project.
The input is articles retrieved by its search engine based on patient records and user queries.
For each article, the authors create a topic tree presenting the sectioning of articles.
A composite topic tree is then generated by merging all topic trees, adding detail nodes and matching the nodes with the query.
Next, they choose the representative sentences for each topic.
The last step involves ordering those sentences using topics ordering and physical position sentences.

\subsubsection{Abstractive Document Summarisation in the Medical Domain}
MUSI~\cite{lenci2002multilingual}  generates either extractive summaries or abstractive summaries.
After selecting sentences for extractive summarisation, it maps them into a predicate argument structure representation using tokenising, chunking, shallow syntactic parsing, morphological analysis, dependency analysis, and mapping to the internal representation.
Finally, natural language generation systems are used to create summaries of those extracted sentences.

TRESTLE (text retrieval extraction and summarisation technologies for large enterprises) is another system that relies on named entity recognition for producing single-sentence summaries of pharmaceutical newsletters~\cite{gaizauskas2001intelligent}. 
Drug names and diseases are named entities and are linked to the newsletter from which they have been extracted.

\subsubsection{Personalised Summarisation in the Medical Domain}
Centrifuser and PERSIVAL produce other types of abstractive summary~\cite{elhadad2001towards}.
They create an informative abstractive summary based on the users' preferences (i.e., physicians, patients or relatives).
The system takes three different sources as input, including patient records, consisting of structured and unstructured documents; journal medical articles extracted from online medical journals, mainly in the field of cardiology; and users’ queries~\cite{elhadad2001towards}.

\subsubsection{Multimedia Document Summarisation in the Medical Domain}
Ebadollahi et al.~\cite{ebadollahi2001echocardiogram} and Xingquan et al.~\cite{zhu2002classminer} present systems that perform document summarisation using multimedia content such as echocardiograms (ECGs) and medical videos, respectively~\cite{johnson2002modeling}.
ECGs are usually videotaped and are transcribed into a digital format.
Summarising an ECG involves selecting the most interesting video frames and enabling users to navigate through the ECGs to find their required parts with ease~\cite{ebadollahi2001echocardiogram}.  
The output can be a static summary made by selecting the extracted key frames or a dynamic summary made by a concatenation of the video’s small extracted sequences.

\subsubsection{Summarisation from a Cognitive Science Perspective}
SummIt-BMT (summarise it in bone marrow transplantation) is a query-based summarisation system based on the cognitive model that deals with summarising MEDLINE articles and abstracts for a bone marrow transplant\textemdash a specialised field of internal medicine~\cite{afantenos2005summarization}.
First, users create a search scenario based on the domain ontology concepts. The scenario is converted to a MEDLINE query and its corresponding articles are included. Pieces of text pointing to the query are given as output.
As such, we propose Summary2vec, a summary embedding approach that provides a low-dimensional learnt continuous vector representation of summaries covering various aspects of information space.
Summary2vec is inspired by Word2vec, which represents document vocabulary, wherein words are mapped to vectors of real numbers.

Word2vec can capture the context of a word in a document, the semantic and syntactic similarity, and relations with other words. 
Similar to Word2vec, the whole information space (document clusters) is considered a document, and each summary a word.
We treated each summary as an independent unit of meaning that conveys information about one or more aspects of a large information space, and propose a hierarchical structure using Summary2vec to facilitate personalised summarisation to better analyse the information space.

\subsection{Summary2vec}
\begin{figure*} [t]
    \centering
    \includegraphics[width=\textwidth]{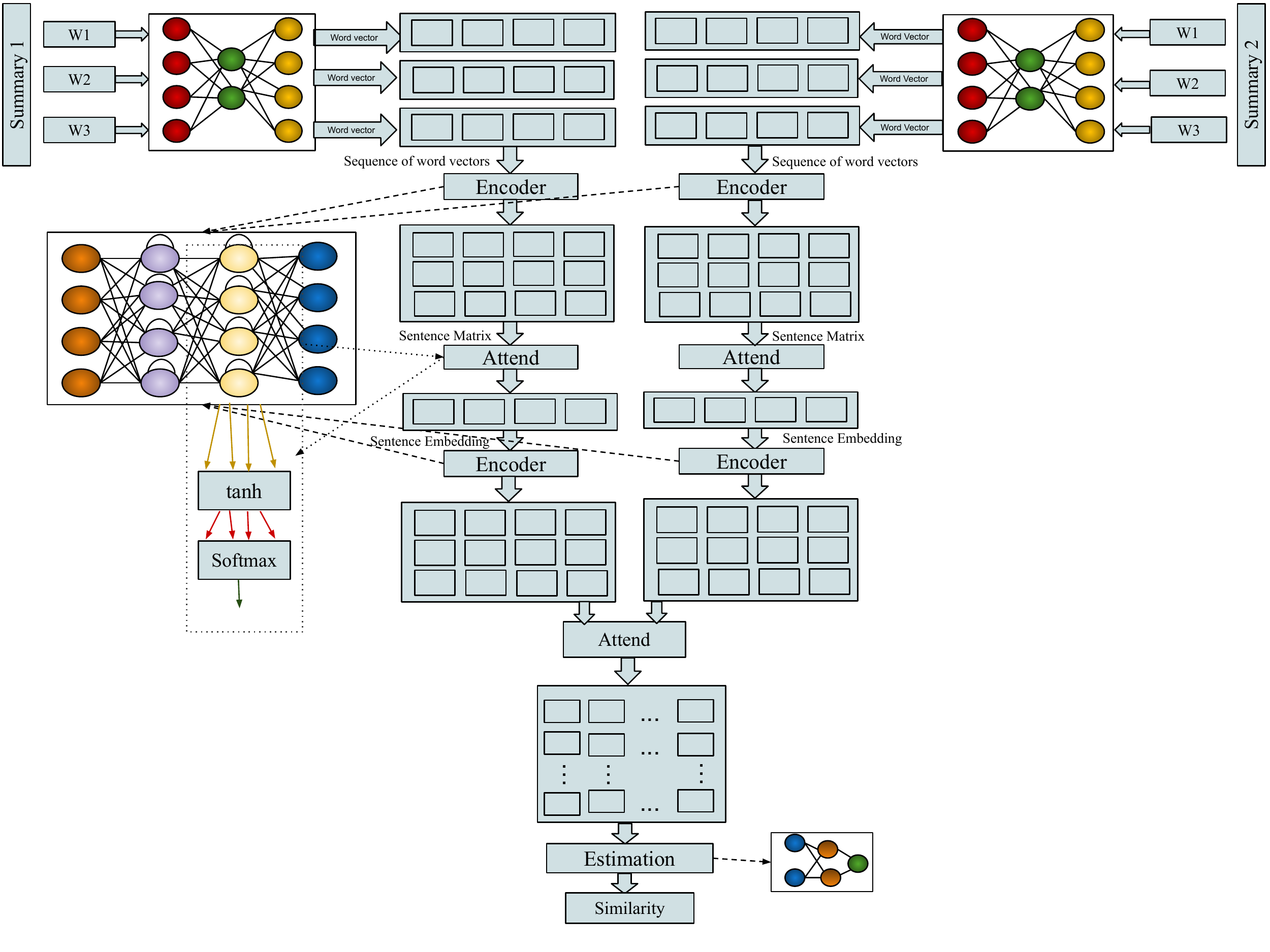}
    \caption{Summary2vec architecture.
    Summaries are transformed into word units to be embedded.
    Sequences of word vectors are encoded to build a sentence matrix. 
    Attention to the sentence matrix is applied to make sentence embedding, which is encoded to make the summary matrix.
    The concatenated output of two parallel networks, corresponding to two summaries, is fed into a fully connected network to estimate the summary’s similarity.}
    \label{fig:hierarchy}
\end{figure*}

The goal of Summary2vec is to create a numeric representation of summaries as an individual related unit of meaning conveying one or more aspects of information space.
Therefore, it is easier for users to obtain the desired information, gain insights and answer complex questions.
Summary2vec is remedial to design automatic services for various analytic purposes that require information-seeking activities.

\subsubsection{Problem Definition}
\label{sum2vec_definition}
We define input $D$  as a set of related documents, and $d$ is a cluster of documents in $D$ .
$Y(d)$ is all the summaries that can be built given document cluster $d$, and $y \in Y(d)$ is a potential summary with limited size $b$.
The goal of Summary2vec as a summary embedding algorithm is to map each input summary y to a vector.
After, any arithmetic operations or comparison is conceivable on summary vectors.

\subsubsection{Methodology}
The proposed architecture depicted in Fig.~\ref{fig:hierarchy} is a nested model composed of four components, including embedding, encoding, attend and similarity estimation.
Each summary in this architecture is a set of sentences, and each sentence a group of words.
Therefore, we began by tokenising the text into words and embedding the words into vectors.

A word embedding is a learnt representation of text such that words with the same meaning have similar representations.
Different techniques are used to learn a word embedding from the text.
Word2vec~\cite{mikolov2013distributed} is an example of a statistical method to learn a distributed representation of words, utilising different architectures.
We used the skip-gram model, which uses the current word to predict the surrounding window of context words by increasing the weights of nearby context words using a neural network model.
More formally, given a sequence of training words $\{w_1, w_2, w_3, ..., w_T\}$, the goal is to maximise the average log probability as in Eq.~\ref{ch7_eq17}~\cite{mikolov2013distributed}.

\begin{equation}
\label{ch7_eq17}
    J=\frac{1}{T} \sum_{t=k}^{T-k} log p(w_{t}|w_{t-k},...,w_{t+k})
\end{equation}

The prediction task is typically done through a multiclass classifier, such as softmax.
Therefore, we have Eq.~\ref{ch7_eq17}.

\begin{equation}
\label{ch7_eq17}
    p(w_{t}|w_{t-k},...,w_{t+k})= \frac{e^{y_{w_t}}}{\sum_i e^{y_{i}}}
\end{equation}

Each of $y_i$ is a normalised log probability for each output word $i$, computed in Eq.~\ref{ch7_eq18}~\cite{mikolov2013distributed}.

\begin{equation}
\label{ch7_eq18}
     y=b+Uh(w_{t-k},...,w_{t+k};W),
\end{equation}
where $U$, $b$  are the softmax parameters, and $h$ is constructed by concatenation or the average of word vectors extracted from $W$.
The neural network-based word vectors are typically trained using stochastic gradient descent, where the gradient is obtained through back propagation~\cite{mikolov2013distributed}.
The embedding $v_{w_{t}}$ is a vector representation of the word $w_t$.
We used Stanford’s GloVe model for word embedding.\footnote{https://nlp.stanford.edu/projects/glove/}

Next, we built a hierarchical model from a sequence of word embeddings for building sentence embeddings using a bidirectional RNN (LSTM).
Given a sequence of word vectors $S=\{v_{w_1},...,v_{w_n}\}$, first the encode step computes a representation, called sentence matrix ($n \times 1$), where each row represents the meaning of each token in the context of the rest of the sentence.
We used LSTM for this purpose, responsible for computing an intermediate representation, denoting the tokens in context.
The vector for each token is computed in two parts\textemdash one part by a forward pass, and another by a backward pass, defined in Eq.~\ref{ch7_eq20}.

\begin{equation}
\label{ch7_eq20}
\begin{split}
      F_i=\mathcal{G} (v_{w_i},\mathcal{G}(S)),  \forall i \in n \\
   B_{l-(i-1)}=\mathcal{G}(v_{l-(i-1)},\mathcal{G}(S)),
\end{split}
\end{equation}
where $\mathcal{G}$ is the feed-forward network using the ReLU activation function.
$F_i$ is the $i$-th forward network, $l$ is the sentence embedding size, and $B_i$ is the $i$-th  backward network.

To obtain the full vector, we concatenated them together as Eq.~\ref{ch7_eq21}.

\begin{equation}
\label{ch7_eq21}
    C_{[l \times 2n]}=[F,B],
\end{equation}
where $[.$ , $.]$denotes the concatenation of two matrices. 
Therefore, matrix C has the shape of $l \times 2n$.

The attend component condenses the matrix representation produced by the encoder to a single vector used as the input to another standard feed-forward network with $tanh$ activation function.
One key advantage the attention mechanism has over other reduction operations is that it characterises input as an auxiliary context vector.
This auxiliary context vector specifies which information to discard.
The reduced vector is tailored to the network, consuming it to compensate for the information loss in reducing the matrix to a vector.

In this architecture, we followed the same attention mechanism proposed by Yang et al.~\cite{yang2016hierarchical}, which takes two matrices and outputs a single vector.
The context vector computed regarding a context vector learnt as a model parameter.
This makes the attention mechanism a pure reduction operation, which could be used in place of any sum or average pooling step.
The attention takes the output of the encoder as in Eq.~\ref{ch7_eq22}.

\begin{equation}
\label{ch7_eq22}
\begin{split}
    e_i=\mathcal{M}(C)\\
    \alpha_i=softmax(e_i)\\
    o_i=\alpha_i\times D_i,
\end{split}
\end{equation}
where $\mathcal{M}$ is a feed-forward network with $tanh$ activation function, and $C$ is the concatenated output matrix of the previous step. 
Vector oi is the output of the attend step to be utilised as the input of the second encode step.
Finally, $D_{[1 \times 2n]}=\sum_{i=1}^{l}C[ik]$, where
$C[ik]$ represents one row of matrix $C$, and $k=2n$.
Hence, $D$ is a single vector that resulted from the summation of all the rows of $C$, as shown in Fig.~\ref{fig:hierarchy}.
A sequence of sentence condensed embedding as the output of the attend step ($o_i$) is given to another encoder, a bidirectional LSTM, to make the summary matrix.
A pair of these networks is set up to produce attention summary vectors from the other summaries, as well be compared and to create an estimation. 
Once the summaries have been compressed (reduced) into a single vector, we can learn the similarity measure between different attention summary vectors using target representation.
The concatenated vector is fed into a fully connected network to estimate the similarity of summaries.
Once summary embeddings are trained, vectorised representations of summaries can be used as input to a wide range of machine learning models.
Sec.~\ref{ch7_persoanlzied} explains the application of Summary2vec in personalised summarisation.

\begin{figure*}[t]
    \centering
    \includegraphics[width=\textwidth]{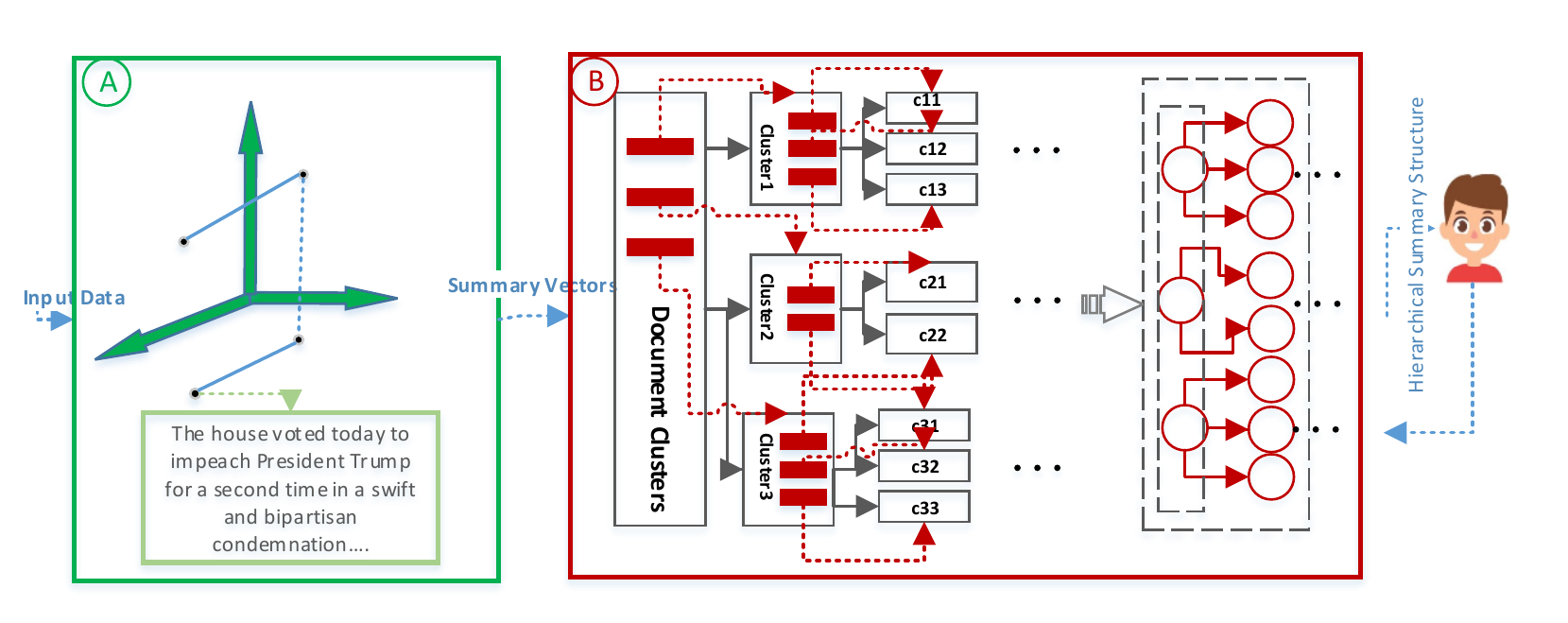}
    \caption{ (A) Summary embeddings are produced. (B) The hierarchical summaries are made based on the proposed hierarchical clustering approach.}
    \label{fig:personalized}
\end{figure*}

\subsection{Personalised Summarisation}
\label{ch7_persoanlzied}
Traditional MDS approaches produce a uniform summary for all users without considering individual interests. 
Therefore, they suffer from two significant limitations.
First, they only provide static summaries that cannot be tailored to specific needs.
Second, most existing summarisation systems cannot explain the generated summary or allow users to explore various aspects of the generated summary.
Consequently, the granularity of the summary cannot be determined, and the output is neither interpretable nor personalised.
Unfortunately, a single summary is unlikely to serve all users in a large population, so the challenge is extracting the desired information from multiple online information sources tailored to a user’s specialty and interest. 
As a result, there is a need for summaries to cater to individual interests and personal background.

Moreover, users’ required information is scattered in a large information space.
Therefore, users must undertake the challenging and time-consuming task of searching that space to find what they desire.
Summary2vec facilitates personalised summarisation and other information-seeking activities required to quickly and easily understand the information space by vectorising each summary as an individual semantic-related entity.
Employing the output of Summary2vec, we propose a hierarchical summarisation structure wherein higher levels of the hierarchy contain the most general aspects and user interaction can occur (depicted in Fig.~\ref{fig:personalized}).

We propose a novel recursive clustering algorithm that uses summary embeddings learnt from a corpus as $S=\{v_{s_1},v_{s_2},...,v_{s_N}\}$.
The hierarchical structure incrementally maintains the most generic information on top (roots).
The parent-to-child links facilitate navigating and drilling down on interesting topics.
Summary2vec facilitates the clustering of summaries using summary vectors.
The hierarchical conceptual clustering minimises the objective function (Eq.~\ref{ch7_eq23}) over all $k$ clusters as  $C=\{c_1,c_2,..,c_k\}$. 

\begin{equation}
\label{ch7_eq23}
J = \sum_{k=1}^{K}\sum_{t=1}^{\mid S \mid} |v_{s_t}- c_k|^2 +\alpha \min _{c\in C} size(c),
\end{equation}
where $c_k$ is the randomly selected centre of $k-th$ cluster.
The second term is the evenness of the clusters added to avoid clusters with small sizes, and $\alpha$ tunes the evenness factor set employing a grid search over the development set.
We implemented hierarchical clustering top-down at each time optimising Eq.~\ref{ch7_eq23}.
Similar summaries can easily be found by comparing users’ selected summary’s vector with any other summary. 
Besides, the output of hierarchical summarisation helps users navigate the hierarchy to find what they need.

\subsection{Evaluation}
The proposed approach is, to the best of our knowledge, the first designed for interactive and hierarchical summarisation of medical data.
Therefore, evaluation of the approach is challenging, as there is no baseline. We evaluated the proposed approach using a PubMed dataset.

PubMed consists of more than 26 million citations for biomedical literature.
The dataset was collected from sources such as MEDLINE, life science journals and published online ebooks.
It also contains links to public access\footnote{https://catalog.data.gov/dataset/pubmed.} text-based content from PubMed Central and other publishers’ websites. 
The data was divided into train, test and validation sets.

We randomly built summaries by selecting sentences and measuring their similarity using ExDos.
We trained the network, defining the word embedding size as 300 and the sentence embedding size as 100.
We then evaluated Summary2vec using the root means square error (RMSE) as the evaluation measure, defined in Eq.~\ref{ch7_eq24}.
\begin{equation}
\label{ch7_eq24}
    RMSE=\sqrt{\frac{1}{P}\sum_{i=1}^P (Similarity_{p_i}-Similarity_{r_i})^2},
\end{equation}
where $P$ is the number of summary pairs, $Similarity_p$ is the predicted similarity by the network, and $Similarity_r$ is the actual similarity value.
If we define $N_train$ rain as the number of train sentences and the summary length as $l$, then the summary set size is
equal to $S_{size}={\binom{N_{train}}{l}}$.
Consequently, the number of summary pairs is equal to $\binom{S_{size}}{2}$.

To evaluate the network, we selected summaries with four different sizes chosen from the set of \{3,5,7,10\}.
The RMSE values are reported in Table~\ref{tab:trainerror}.
Evidently, as the number of sentences in a summary increases, the similarity of the summaries also increases.
Consequently, it is more probable to be over-fitted. 
In our experiments, we set the summary length at $5$.
Since RMSE evaluates the performance of the network trained for estimating summary similarities, we evaluated the produced summary vector using the proposed hierarchical summarisation approach.
We also performed human studies, as the primary purpose of the approach is to help users find what they desire.

One challenge of hierarchical clustering in producing hierarchical summaries is finding the optimal number of clusters.
To tackle this problem, we used the gap statistic~\cite{tibshirani2001estimating} defined in Eq.~\ref{ch7_eq25}.

\begin{equation}
\label{ch7_eq25}
Gap_n(k)= E_{n}{log(W_{k})}{*}-log(W{k}),
\end{equation}
where $W_{k}$ is the clusters' score using objective functions, and $E_{n}{*}$ is the expectation under a sample of size n from a reference distribution.
We performed the gap statistic at each level using values from${5,10,15,20,25,30}$.
The best value at each level was chosen.

Since the purpose of a hierarchical summary is to help users retrieve information, we analysed two aspects of user requirements: (i) information coverage (how much information the summary covers), and (ii) knowledge extraction (how much users can learn from summaries).
We conducted human experiments and designed a series of micro-tasks for each experiment. Thirty MTurk workers were hired to complete the tasks.

Five document clusters were randomly selected from the datasets.
Each evaluator was presented with three documents, to avoid subject bias, and was given two minutes to read each article.
To ensure the human subjects understood the study’s objective, workers were asked to complete a qualification task first.
They were required to write a summary of their understanding and answer questions before undertaking any test.
We manually removed spam from the results.

\begin{table}
\centering
  \centering \caption{valuating Summary Size Effect According to RMSE Value}
  \label{tab:trainerror}
  \tabcolsep=0.11cm
  \begin{tabular}{|c|c|c|}
    \hline
    Summary Size & Train RMSE & Test RMSE\\
   \hline
    3 &0.22&0.31\\
    \hline
    5 &0.09&0.12\\
    \hline
    7 &0.10&0.22\\
    \hline
    10 &0.04&0.25\\
    \hline
\end{tabular}
\end{table}
We analysed the information coverage aspect first automatically using $ROUGE_n$~\cite{lin2004rouge}\footnote{We run ROUGE 1.5.5: http://www.berouge.com/Pages/defailt.aspx with parameters -n 2 -m -u -c 95 -r 1000 -f A -p 0.5 -t 0}.

To automatically evaluate the proposed approaches, we compared the approach with ExDos.
We also compared the percentage of common unigrams and bigrams between these two approaches and the summary written by workers within the same summary size (100 words) at the first level of the hierarchy.
The ROUGE-1 and ROUGE-2 scores were 28\% and 13\% higher than ExDos, at 76\% and 51\%, respectively.

Moreover, for human evaluation of information coverage, we asked MTurk workers to read an article on the same topic and identify the five most common sentences (summary size).
We aggregated responses from 10 workers for each topic, and then analysed the presence of these sentences at the first levels of the hierarchy.
Then, we evaluated them based on the recall measure—the percentage of essential sentences mentioned at the first level.
We repeated this experiment for five topics and averaged the recall measures.
The proposed approach contained 91\% of all important sentences mentioned by workers at the first level and 5\% at the second level.
This experiment illustrates that the first hierarchy level works as a general summarisation tool containing only the most critical sentences.
Users are also allowed to navigate the hierarchy should they require more details.

To assess the possibility of users finding their desired information (knowledge extraction), the workers were asked to answer questions on new topics using hierarchical summarisation. 
Questions were selected to contain both detailed and general topics. Participants’ level of confidence in answering the questions as well as their responses were recorded and assessed by an evaluator for accuracy.
Among the 16 workers, 86\% were completely confident in their answers, but only 52\% answered entirely accurately.
The same experiment was repeated using the ExDos output as a traditional competitor approach.
This time, only 23\% answered completely accurately, and the average confidence level was 31\%.

\section{Summarising Business Process Data}

The large amount of raw data generated by IoT-enabled devices provides real-time intelligence to organisations, and this can enhance knowledge-intensive processes~\cite{schiliro2018icop}. 
The problem with understanding the behaviour of information systems as well as the processes and services they support has become a priority in both medium and large enterprises. 
This is demonstrated by the proliferation of tools for the analysis of process executions, system interactions and system dependencies, and by recent research on process data warehousing, discovery and mining~\cite{van2011process}.
Accordingly, identifying business needs and determining solutions to business problems requires the analysis of business process data~\cite{processBook,DBLP:conf/bpm/BeheshtiBNS11}; this, in turn, will help discover useful information and support decision-making for enterprises.
For example, one intervention that has emerged as a potential solution to the challenges facing law enforcement officers is the use of an interactive constable in a patrol system.
In these processes, it is not sufficient to focus on data storage and analysis and the knowledge workers (e.g., investigators) will need to collect, understand and relate big data (scattered across various systems) to process and communicate their findings, support evidence, and make decisions.
Therefore, we present a scalable and extensible IoT-enabled process data analytics pipeline known as iProcess~\cite{beheshti2018iprocess}.
This helps analysts better absorb data from IoT devices, extract knowledge, and link the two to process (execution) data.

We present novel techniques to summarise the linked IoT and process data to construct process narratives.
Summarisation techniques presented in this section include a novel approach that helps analysts understand and relate big IoT as well as process data to communicate and support their findings with ease.
The proposed approach will enhance data-driven techniques for improving risk-based decision-making in knowledge-intensive processes.
We present a set of innovative, fine-grained and intuitive analytical services to discover patterns and related entities, and further enrich them with complex data structures (e.g., time series, hierarchies and subgraphs) to construct narratives. 
We also propose a framework and algorithms for summarising the (big) process data and constructing process narratives.
This helps us complete the first step towards enabling \emph{storytelling} with process data. 
Before explaining the details of the proposed method, we provide an example of the motivating scenario in Sec.~\ref{ch7_motivating}.

\subsection{Motivating Scenario: Missing People}
\label{ch7_motivating}
Between 2008 and 2015, over 305,000 people were reported missing in Australia (aic.gov.au/), an average of 38,159 reports each year. 
In the United States (nij.gov/), there are as many as 100,000 active missing person’s cases on any given day.
The first few hours following a person’s disappearance are the most crucial.
The sooner police are able to piece together the sequence of events and actions right before a disappearance, the greater the chance of finding a missing person.
This entails gathering information about the person, including physical appearance, their activities on both social media and in the physical/social realm, the data and communication stored in phone calls and emails, and information collected through public means (e.g., CCTV).

The investigation process is a data-driven, knowledge-intensive and collaborative one.
The information associated with an investigation (case process) is usually complex, entailing the collection and presentation of many different types of documents and records.
It is also common that separate investigations may affect other investigation processes.
Nonetheless, the more evidence (knowledge and facts extracted from the data in the data lake~\cite{coredb}) collected, the better related cases could be linked explicitly.
Although law enforcement agencies use data analysis, crime prevention, surveillance, communication, and data sharing technologies to improve their operations and performance, in sophisticated and data-intensive cases such as missing persons there remain many challenges.
For example, fast and accurate information collection and analysis is vital in law enforcement applications~\cite{braga2015police}.
From the policymaker perspective, this trend calls for the adoption of innovations and technologically advanced business processes that can help law enforcers detect and prevent criminal acts.
Enabling IoT data in law enforcement processes will give investigators access to a potential pool of data evidence.
Then, the challenge is to prepare the big process data for analytics, summary, to construct narratives and to help analysts link these narratives to uncover facts with ease.

\begin{figure} [t]
\hspace*{-0.5cm}
\centering
\includegraphics[width=1.1\textwidth]{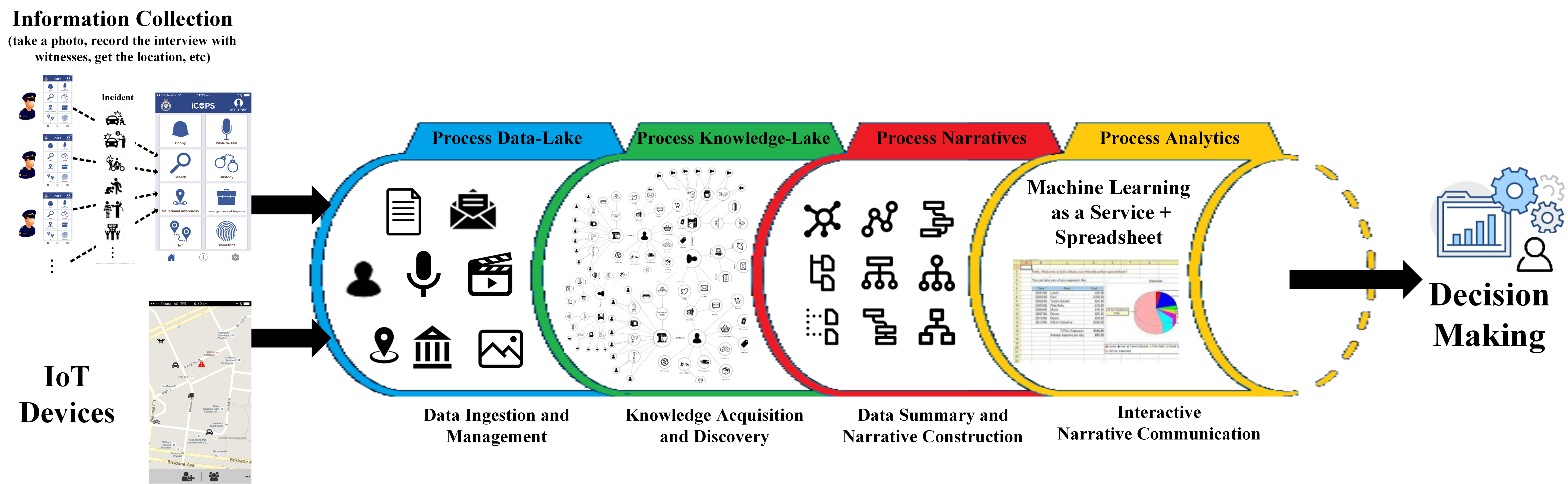}
\caption{IoT-enabled process data analytics pipeline.}
\label{fig:architecture}
\end{figure}

We aim to address this challenge by supplying police officers with internet-enabled smart devices (e.g., phones and watches). 
This will assist them in the process of collecting evidence, provide access to location-based services to identify and locate resources (CCTV, cameras on on-duty officers, police cars, drones and more), organise all these islands of data in a knowledge lake~\cite{coreKG,DBLP:conf/wise/BeheshtiBSS19,DBLP:journals/dpd/BeheshtiBTMBN19}, and feed that information into a scalable and extensible IoT-enabled process data analytics pipeline.
Fig.~\ref{fig:architecture} illustrates this framework, explained in greater detail in Sec.~\ref{DataSpace}.

\subsection{Process Data Lake and Process Knowledge Lake}
\label{DataSpace}
To understand data-driven knowledge-intensive processes, we leveraged CoreDB (a data lake as a service)\cite{coredb} to identify (IoT, private, social and open) data sources and ingest the big process data in the data lake.
CoreDB manages multiple database technologies (from relational to NoSQL), offers a built-in design for security and tracing, and provides a single REST API to organise, index and query the data and metadata in the data lake.
We leveraged the knowledge lake~\cite{coreKG} to transform raw data (unstructured, semi-structured and structured data sources) into a contextualised data and knowledge basis that is maintained by and made available to end users and applications.

\subsection{Data Summaries and Narratives Construction}
\label{Narratives}
\begin{figure} [t]
\centering
\includegraphics[width=1.1\textwidth]{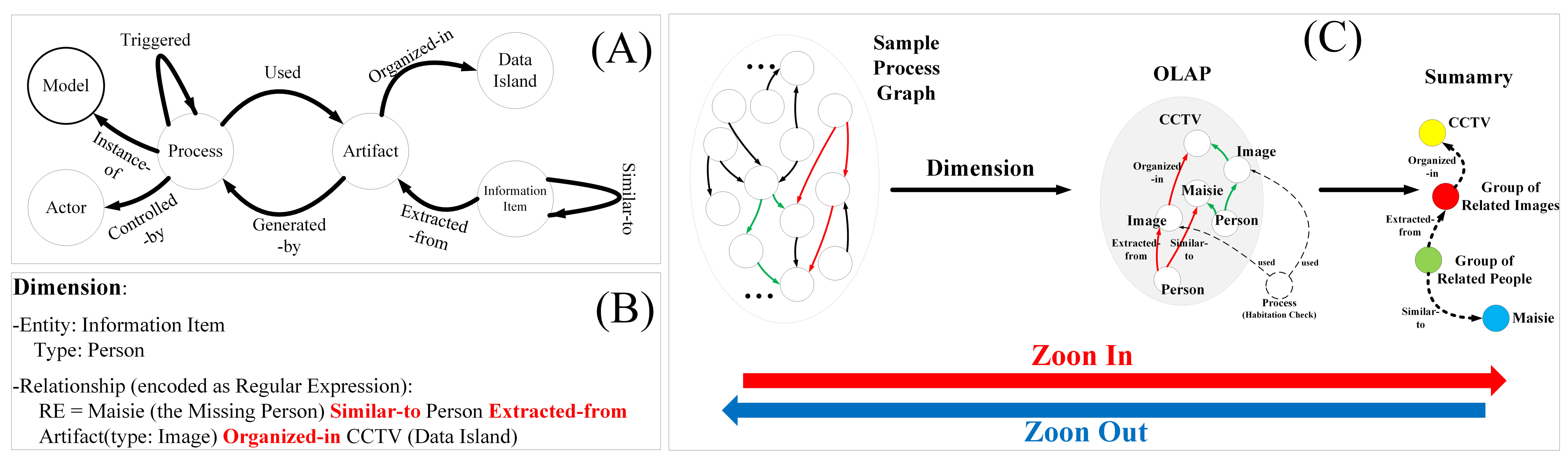}
\caption{Process knowledge graph schema. (A) A sample OLAP dimension, and (B) an interactive graph summary (C).}
\label{fig:schema}
\end{figure}

In this phase, we present an online analytical processing server (OLAP)-style~\cite{DBLP:journals/dpd/BeheshtiBM16} technique as an alternative to query and analysis techniques.
This approach will isolate the process analyst from explicitly analysing different dimensions such as time, location, activity, actor and more.
Instead, the system will be able to use interactive (artefacts, actors, events, tasks, time, location, etc.) summary generation to select and sequence narratives dynamically. 
This novel summarisation method will enable process analysts to choose one or more dimensions (i.e., attributes and relationships) based on their specific goal, and further interact with small and informative summaries. 
This will facilitate the process analysts undergo to analyse data from various dimensions.
Fig.~\ref{fig:schema}(B) illustrates a sample OLAP dimension.

In OLAP~\cite{DBLP:journals/dpd/BeheshtiBM16}, cubes are defined as sets of partitions, organised to provide a multidimensional and multilevel view (where partitions are considered the unit of granularity).
Dimensions are defined as perspectives used for examining the data within constructed partitions.
In police investigation scenarios, such as the 2013 Boston bombing, process cubes can enable effective analysis of the process knowledge graph from different perspectives and with multiple granularities\textemdash for example, by aggregating and relating all the evidence from a person of interest, location of the incident and more.
Following this, we define a process cube as such.

\begin{definition}\label{definition:Relationship} \emph{
(Process Cube)
A process cube is defined to extend decision support on multidimensional networks (e.g., process graphs), considering both data objects and the relationships between them.
We reused and extended the definition for graph cubes proposed in our previous work~\cite{DBLP:journals/dpd/BeheshtiBM16,DBLP:journals/pvldb/HammoudRNBS15}.
In particular, given a multidimensional network $N$, the graph cube is obtained by restructuring $N$ in all possible aggregations of a set of node/edge attributes $A$, where for each aggregation $A'$ of $A$, the measure is an aggregate network $G'$ w.r.t.$A'$. 
We also defined possible aggregations upon multidimensional networks using regular expressions.
In particular, $Q=\{q_1, q_2, ..., q_n\}$ is a set of n process cubes, where each $q_i$ is a process cube (a placeholder for a set of related entities and/or relationships among them) and can be encoded using regular expressions.
In this context, each process cube $q_i$ can extensively support multiple information needs with the graph data model (e.g., Definition~1) and one algorithm (regular language reachability).
The set of related process cubes $Q$ is designed to be customisable by local domain experts (who have the most accurate knowledge about their requirement) to codify their knowledge into regular expressions.
These expressions can describe paths through the nodes and edges in the attributed graph: then, $Q$ can be constructed once and reused for other processes. 
The key data structure behind the process cube is the process knowledge graph (i.e., a graph of typed nodes), which represents process-related entities (such as process instances, models, artefacts, actors, data sources and information items), and typed edges, which label the relationships of the nodes to one another (illustrated in Fig.~\ref{fig:schema}(A)).
We leveraged the graph mining algorithms in our previous work~\cite{DBLP:journals/dpd/BeheshtiBM16} to walk the graph from one set of interesting entities to another via the relationship edges, and to discover which entities are ultimately transitively connected.
We then grouped them in folder nodes (set of related entities) and path nodes (set of related patterns).
We used correlation conditions~\cite{motahari2011event} to partition the process knowledge graph based on a set of dimensions derived from the attributes of node entities.
We used a path condition~\cite{DBLP:journals/dpd/BeheshtiBM16} as a binary predicate defined on the attributes of a path.
This allowed us to identify whether two or more entities are related through that path.
}\end{definition}

\begin{definition}\label{definition:Relationship} \emph{
(Dimensions)
Each process cube $q_i$ has a set of dimensions $D=\{d_1, d_2, ..., d_n\}$, where each $d_i$ is a dimension name.
Each dimension $d_i$ is represented by a set of elements (E), where elements are the nodes and edges of the process knowledge graph.
In particular, $E=\{e_1, e_2, ..., e_m\}$ is a set of $m$ elements, where each ei is an element name.
Each element ei is represented by a set of attributes (A), where $A=\{a_1, a_2, ..., a_p\}$ is a set of p attributes for element $e_i$, and each $a_i$ is an attribute name. 
A dimension $d_i$ can be considered a given query that requires grouping graph entities in a certain way.
Correlation conditions and path conditions can be used to define such queries.
}\end{definition}

A dimension uniquely identifies a subgraph in the process knowledge graph, which we call a ‘summary’.
Now, we introduce the new notion of ‘narrative’.

\begin{definition}\label{definition:Relationship} \emph{
(Narrative)
A narrative $N=\{S,R\}$ is a set summaries $S=\{s_1, s_2, ..., s_n\}$ and a set of relationships $R=\{r_1, r_2, ..., r_m\}$ among them, where $s_i$ is a summary name and $r_j$  is a relationship of type ‘part-of’ between two summaries. 
This type of relationship enables the zoom-in and zoom-out operations (see Fig.~\ref{fig:schema}(C)) to link different pieces of a story, and further enables analysts to interact with narratives.
Each summary $S=\{Dimension,View-Type,Provenance\}$, identified by a unique dimension $D$, relates to a view type $VT$ (e.g., process, actor or data view), and is assigned to a provenance code snippet $P$  to document the evolution of the summary over time.
(More nodes and relationships can be added to the process knowledge graph over time.) We leveraged our work~\cite{DBLP:conf/caise/BeheshtiBN13}  to document the evolution of summaries over time.
}\end{definition}

\begin{figure} [t]
\hspace*{-0.2cm}
\centering
\includegraphics[width=1.15\textwidth]{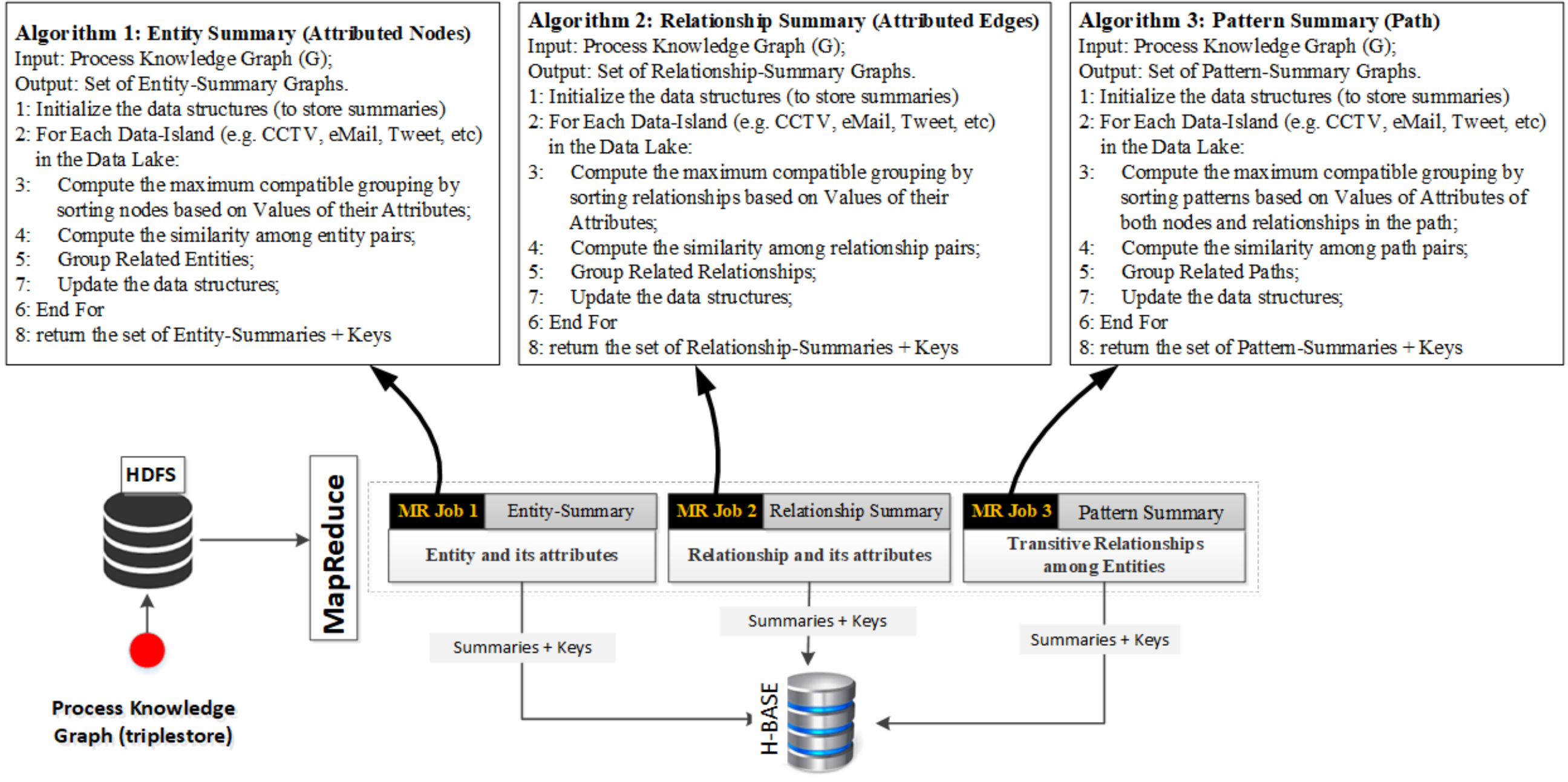}
\caption{Scalable summary generation.}
\label{fig:summary}
\end{figure}

The formalism of the summary $S$ will enable users to consider different dimensions and views of a narrative, including the event structure (narratives are about something happening), the purpose of a narrative (narratives about actors and artefacts), and the role of the listener (narratives are subjective and depend on the perspective of the process analyst).
Also considered was the importance of time and provenance, as narratives may have different meanings over time.
As such, we developed a scalable summary generation algorithm that supports three types of summaries. 
Fig.~\ref{fig:summary} illustrates the scalable summary generation process.
The three summary types are as follows:

\begin{itemize}
  \item \textit{Entity summaries} use correlation conditions to summarise the process knowledge graph based on a set of dimensions deriving from the attributes of node entities.
  In particular, a correlation condition is a binary predicate defined on the attributes of attributed nodes in the graph, allowing users to identify whether two or more nodes are potentially related. 
  Algorithm~1 in Fig.~\ref{fig:summary} will generate all possible entity summaries. 
  For example, one possible summary may include all related images captured in the same location. 
  Another summary may include all related images captured in the same timestamp.

  \item \textit{Relationship summaries} use correlation conditions to summarise the process knowledge graph based on a set of dimensions deriving from the attributes of attributed edges. 
  Algorithm~2 in Fig.~\ref{fig:summary} will generate all possible relationship summaries.
  For example, one possible summary may include all related relationship types ‘controlled-by’ and have the following attributes: ‘Controlled-by (role=‘Investigator’; time=‘$\tau_1$’; location=‘255.255.255.0’)’.
  Within the relationship summaries we also store the nodes from and to the relationship (in this context, the process instance and the actor).
  
  \item \textit{Path summaries} use path conditions to summarise the process knowledge graph based on a set of dimensions deriving from the attributes of nodes and edges in a path.
  (A path is a transitive relationship between two entities showing a sequence of edges from the start entity to the end.) In particular, a path condition must be defined on the attributes of nodes and edges, allowing users to identify whether two or more entities (in a given process knowledge graph) are potentially related through that path.
  Algorithm~3 in Fig.~\ref{fig:summary} will generate all possible path summaries.
  For example, one possible relationship summary includes all related images captured in the same location and contains the same information item (e.g., a missing person).
  Another relationship summary includes all related tweets or emails sent on timestamp $\tau_1$ and includes the keyword ‘Maisie’ (the missing person).
\end{itemize}

\subsection{Process Analytics}
\label{Analytics}

\begin{figure}[t]
\hspace*{-0.7cm}
\centering
\includegraphics[width=\textwidth]{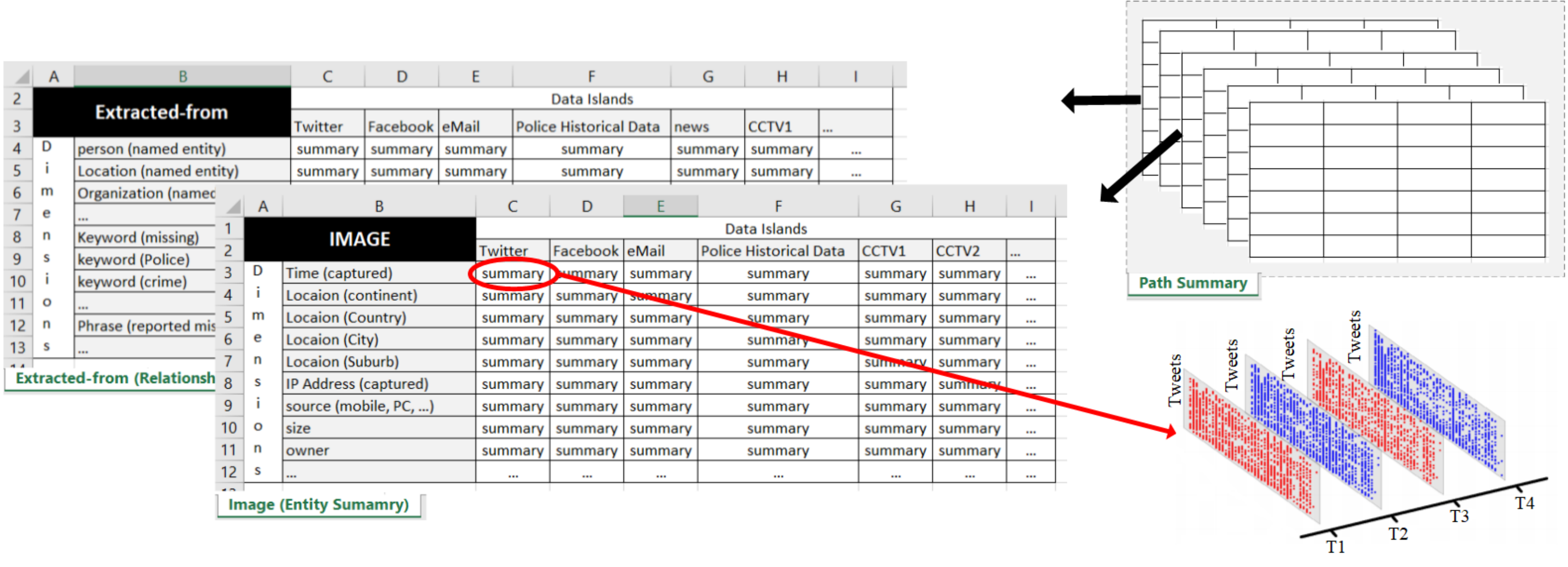}
\caption{Presenting a spreadsheet-like interface on top of the scalable summary generation framework~\cite{amouzgar2018isheets}.}
\label{fig:spreadsheet}
\end{figure}
In this phase, we present a spreadsheet-like interface on top of the scalable summary generation framework.
The goal is to enable analysts to interact with the narratives and control the resolutions of summaries. 
A narrative $N$ can be analysed using three operations.
These are:
\begin{itemize}
    \item~roll up—to aggregate summaries by moving up along one or more dimensions, and to provide a smaller summary with fewer details
    \item~drill down—to disaggregate summaries by moving down dimensions, and to provide a larger summary with more details
    \item~slice and dice—to perform selection and projection on snapshots.
\end{itemize}

To achieve this goal, we used spreadsheets and organised all the possible summaries in the rows and columns of a grid. Each tab in the spreadsheet defines a summary type (e.g., entity, relationship or path summary), the rows in a tab are mapped to the dimensions (e.g., attributes of an entity), and the columns in a tab are mapped to various data islands in the data lake. Each cell contains a specific summary.

We created a set of machine learning algorithms available as a service, purposed to help analysts manipulate and use the summaries in spreadsheets. This supports the following functions:

\begin{itemize}
    \item~The \textit{roll-up} operation performs aggregation on a spreadsheet tab, either by climbing up a concept hierarchy (i.e., rows and columns, which represent the dimensions and data islands accordingly) or by climbing down a concept hierarchy (i.e., dimension reduction).
    \item~The \textit{drill-down} operation is the reverse of roll up. This function navigates from less detailed summaries to more detailed summaries.
    It can be realised either by stepping down a concept hierarchy or introducing additional dimensions.
    For example, in Fig.~\ref{fig:spreadsheet}, applying the drill-down operation on the cell intersecting time (dimension) and CCTV1 (data source) will provide a more detailed summary, grouping all the items over different points in time.
    As another example, applying the drill-down operation on the cell intersecting country (dimension) and Twitter (data source) will provide a more informative summary, grouping all the tweets sent in different countries.
    \item~The \textit{slice} operation performs a selection on one dimension of the given tab, resulting in a sub-tab. The dice operation defines a sub-tab by performing a selection on two or more dimensions.
    This will enable analysts, for example, to see tweets from two dimensions, such as time and location.
    The slice-and-dice operation can be simply seen as a regular expression that groups together different entity and/or relationship summaries (presented in the spreadsheet tabs), and weaves them together to construct path summaries, illustrated in Fig.~\ref{fig:spreadsheet}.
\end{itemize}

\subsection{Implementation and Evaluation}
\label{implementation}

\begin{figure} [t]
\centering
\includegraphics[width=\textwidth]{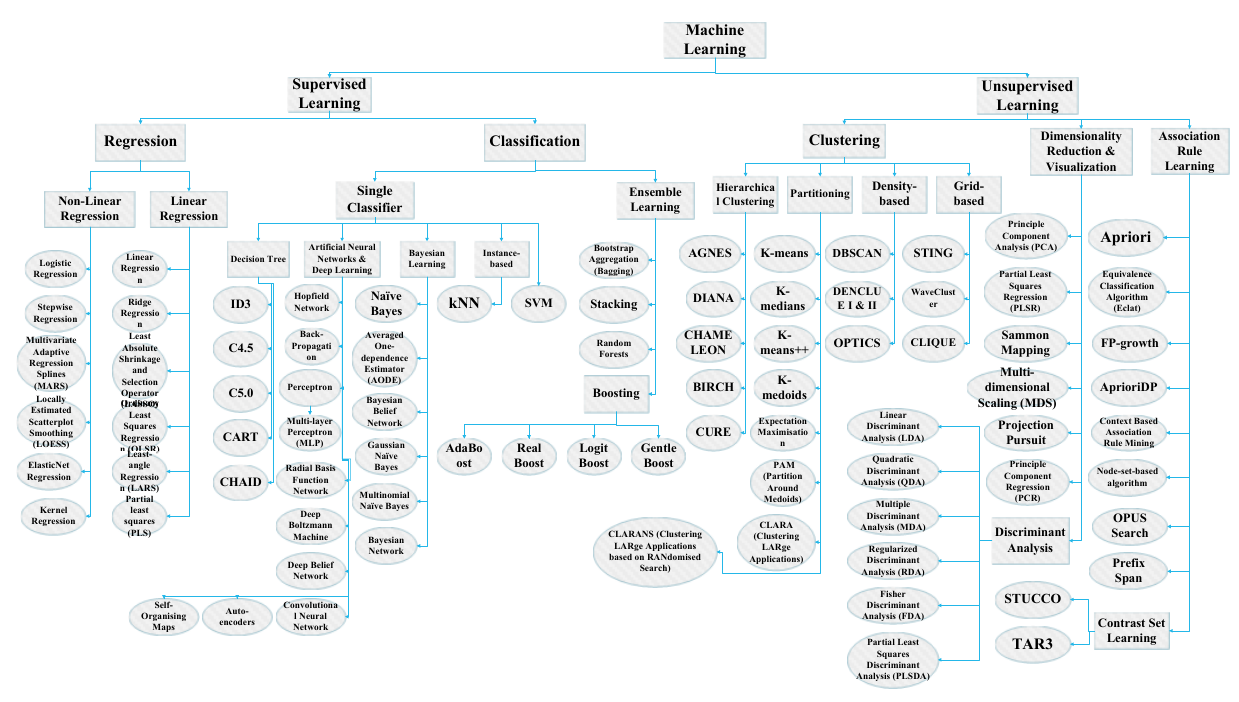}
\caption{Taxonomy of the machine learning algorithms used as a service to streamline the process of acquiring knowledge when interacting with the summaries.}
\label{fig:iML}
\end{figure}

We focus on the motivating scenario to assist knowledge workers in the domain of law enforcement to collect information from an investigation scene, and from IoT-enabled devices of interest (including mobile phones), with greater ease.
The goal here is to contribute to the research and thinking towards making police work more effective and efficient, while augmenting officers’ knowledge and decision management processes through superior information and communication technology.

We developed ingestion services to extract the raw data from IoT devices such as CCTVs, location sensors in police cars and smart watches (to detect the location of people on duty), and police drones. These services will persist the data in the data lake. 
Next, inspired by Google Knowledge Graph (developers.google.com/knowledge-graph/), we focused on constructing a policing process knowledge graph—an IoT infrastructure that can collaborate with internet-enabled devices to collect data, understand events and facts, and assist law enforcement agencies in analysing and understanding the situation to choose the best next step in their processes.
There are many systems that can be used at this level to extract information items from artefacts (such as emails, images and social items), including our curation APIs~\cite{wwwCuration}, Google Cloud Platform (cloud.google.com/) and Microsoft’s Computer Vision API (azure.microsoft.com/).

We have identified many useful machine learning algorithms that helped us summarise knowledge graphs and extract complex data structures, such as time series, hierarchies, patterns and subgraphs, each subsequently linked to such entities as business artefacts, actors and activities.
Fig.~\ref{fig:iML} illustrates the taxonomy of these services. We used a spreadsheet-like dashboard that permits easy interaction for knowledge workers accessing the summaries.
The dashboard enables monitoring the entities (e.g., IoT devices, people and locations) and searching for facts (e.g., suspects, evidences and events).
A set of services has been developed to link the dashboard to the knowledge graph and the data summaries.

\begin{figure} [t]
\centering
\includegraphics[width=1.0\textwidth]{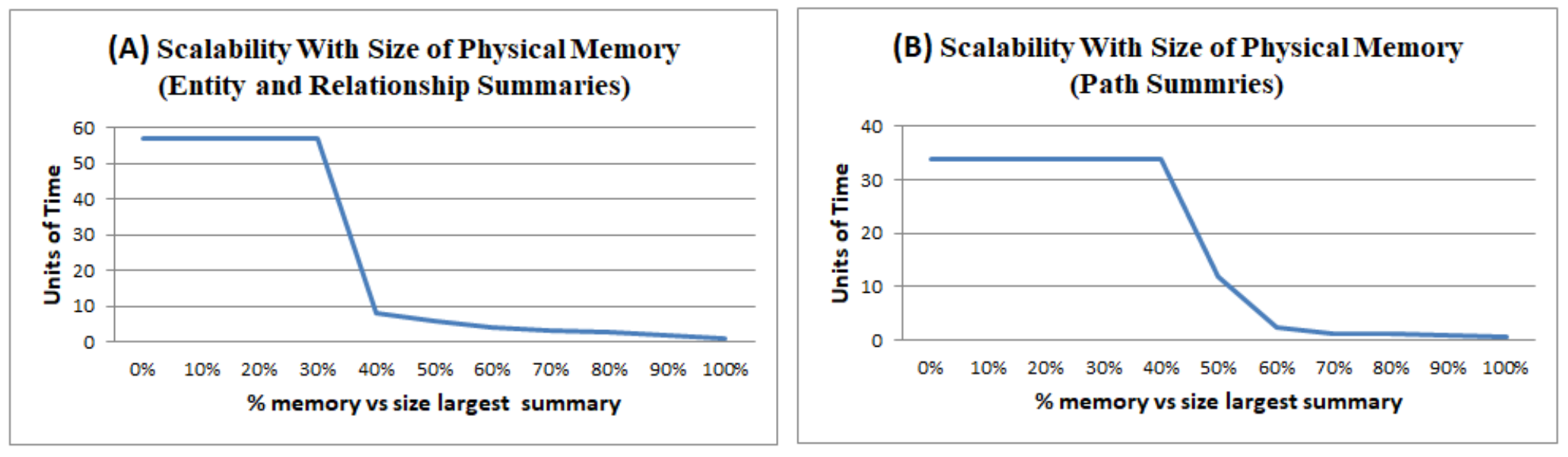}
\caption{Scalability with the size of physical memory for entity and relationship summaries. (B) Scalability with the size of physical memory for path summaries.}
\label{fig:evaluation}
\end{figure}

Fig.~\ref{fig:evaluation} plots the performance of our access structure as a function of available memory for the entity/relationship and path summaries.
These summaries have been generated from a Twitter dataset containing over 15 million tweets, persisted and indexed in the MongoDB (mongodb.com) database in our data lake.
For the path summaries, we have limited the depth of the path to form a maximum of three transitive relationships between the start and end nodes.
This experiment was performed on the Amazon EC2 platform using instances running the Ubuntu Server 14.04. 
The memory size is expressed as a percentage of the size required to fit the largest partition of data in the hash access structure in the physical memory.
For efficient access to single cells (i.e., a summary), we built a partition-level hash access structure, where the partitions will be stored as memory and the operations will be evaluated one partition at a time.
If a summary does not fit in the memory, we incur an I/O if a referenced cell is not cached. In the case of an entity/relationship summary (see Fig.~\ref{fig:evaluation}(A)), this occurs when the available memory is around 40\% of the largest summary.
For the path summary (see Fig.~\ref{fig:evaluation}(B)), this occurs when the available memory is around 30\% of the largest summary.

\section{Summary}
This chapter proposed solutions to the information overload problem in different domains through summarisation, proving its importance and flexibility. We explained and provided both the challenges and state-of-the-art approaches in each domain.
We also offered solutions to each of the applications and evaluated the proposed models, addressing three specific applications in the process. 
These include:
\begin{itemize}
    \item the use of summarisation in detecting anomalies in network traffic data
    \item application of summarisation in healthcare analytic problems
    \item using summarisation to narrate business process data.
\end{itemize}

%% file: ch_8/Conclusion.tex
\chapter{Conclusion and Feature Work}
\label{ch_8}
This chapter summarises the contributions and findings in this dissertation and outlines possible directions for future research.

\section{Summary of Contributions}
This dissertation has shown that personalised and human-in-the-loop summarisation is an important task, but it is fraught with many challenges that have not been addressed adequately in the field.
Ch.~\ref{Ch_2} provides an overview of document summarisation and proposes a new categorisation schema for state-of-the-art summarisation approaches based on existing gaps in the field.
Ch.~\ref{ch_3} summarised the experimental setting and the dataset used to evaluate the models.
Overall, this thesis made three major contributions towards addressing the problem of personalised and human-in-the-loop summarisation, discussed as follows.

Ch.~\ref{ch_4} introduced the central problem of feature engineering in document summarisation.
We proposed a general-purpose extractive approach for summarising documents.
As shown, the algorithm achieved better results than most state-of-the-art methods in terms of efficiency and performance, and found that features are not equally important.
Besides, the post-trained weights represent the importance of each feature in discriminating against each class.

We also proposed NARS, a novel personalised interactive hierarchical summarisation approach that enables users to explore the information they desire with minimal reading required (Ch.~\ref{ch_5}).
NARS is in contrast to a generic summary that is unique for all users, who, by navigating the personalised hierarchy, can search for information based on their own needs and interests.
Two variants of the approach, a SNARS and a FNARS, were also proposed.
FNARS provides a more concise overview of information, while SNARS offers greater detail.
Conversely, FNARS is a fully structured model that can be used for further analysis.
Overall, the proposed approaches help users with general knowledge about a topic to explore a wide range of information.
We evaluated our approach using both automatic and human evaluation, considering four aspects: information coverage, knowledge extraction, effectiveness and user preference.

Given the limited amount of work on interactive and personalised MDS, we studied this problem by optimising summaries based on user feedback (Ch.~\ref{ch_6}).
We subsequently proposed a summary recommendation framework that interactively learns how to generate personalised summaries.
Employing user feedback and domain expert knowledge in a single framework demonstrated the proposed approach’s ability to generate personalised summaries.
We also studied predicting structured and personalised summaries that can help tackle the variety and volume of big generated data.
We highlighted the benefit of using RL algorithms in personalised summarisation, but realised that defining the reward to capture user feedback poses a significant challenge.
That said, evaluating a personalised summarisation approach remains an even greater challenge.

Finally, Ch.~\ref{ch_7} proposed new summarisation models to tackle the problem of information overload in various domains, including network traffic data, health analytics, and business process data.
We proposed Summary2vec, a summary embedding algorithm that can offer a solution to feature engineering in document summarisation.

\section{Open Challenges and Future Directions}
Many possible directions have emerged for future study as the output of the research discussed in this dissertation. 
These are as follows.

\subsection{Personalised Feature Engineering for Summarisation}
Recent advances in neural network approaches eliminate the need for feature engineering in different applications.
However, for personalised approaches, there is still a need for feature engineering~\cite{ghodratnama2021intelligent}. 
One research direction concerns personalised approaches for feature engineering as the basis for personalised summarisation using techniques based on crowd knowledge.

Besides, existing research on summarization incorporating human-in-the-loop systems is limited due to current challenges~\cite{summary2vec}.
First, capturing users’ interests is a significant challenge in providing effective personalised summaries.
This is because users are generally reluctant to specify their preferences, as entering lists of interests may be a tedious and time-consuming task.
Besides, people’s interests are not static and change over time. Therefore, techniques that extract implicit information about users’ preferences are a next-step approach for generating effective personalised summaries.
Another potential direction is to use human feedback history on new domains using transfer learning.

\subsection{Summary Representation and Visualisation}
One alternative direction to approaching interactive summarisation concerns summary representation.
Different forms of feedback to extract user interests need to be evaluated from users’ perspectives and cognitive load~\cite{DBLP:conf/intellisys/SchiliroBM20,zakershahrak2020we,zakershahrak2020order,zakershahrak2020online}.
Examples include learning the process of graph construction or navigation from users’ perspectives.
Another extension of this work is to take advantage of users’ queries in generating personalised summaries, using approaches such as storytelling with data~\cite{DBLP:conf/www/BeheshtiTB20,DBLP:conf/wise/TabebordbarBB19}.
Identifying the appropriate summary size and the number of feedback loops is another crucial aspect of the summarisation process.

One promising direction is to refine the user interface and combine the summarisation ideas in this dissertation with different visualisation techniques.
These can then be used as storytelling tools.
Designing intelligent user interfaces can help researchers better analyse different effects and obstacles when interacting with users and, thus, improve interaction quality, reducing users’ cognitive load and avoiding additional noise.

\subsection{Real-time Summarisation and Performance Boosting}
The largest stumbling block facing the proposed methodologies concerns the processing time, which was not prioritised in this study.
Therefore, one possible extension is to employ scaling techniques to make personalised real-time summaries, which can process streaming input data quickly and rapidly alter the summary based on recent data~\cite{ghodratnama2020rare}.

\subsection{Datasets and Evaluation Metrics}
The most interesting finding of this dissertation is that evaluating a personalised summary is the most challenging part of designing a summarisation model.
Evaluating a personalised summarisation approach is difficult due to the high subjectivity of the problem.
There is also an equal need for test datasets that provide sufficient contextual information~\cite{DBLP:conf/momm/BeheshtiHY19,ghodratnama2020adaptive}.
Future work in this domain could focus on using crowdsourcing techniques~\cite{DBLP:conf/caise/BeheshtiVBT18,DBLP:journals/fgcs/SerhaniKSNBB20} to facilitate personalised summarisation, both for evaluation and creation purposes in various domains.

\subsection{Summarizing Business Process Data}
In today's knowledge-, service-, and cloud-based economy, businesses accumulate massive amounts of data from a variety of sources~\cite{DBLP:conf/assri/ShahbazBNQPM18,DBLP:conf/caise/BeheshtiBN13}. In order to understand businesses one may need to perform considerable analytics over large hybrid collections of heterogeneous and partially unstructured data that is captured related to the process execution~\cite{DBLP:conf/icsoc/SunBBB15,DBLP:journals/spe/BeheshtiBM18}. This data, usually modeled as graphs, increasingly come to show all the typical properties of \emph{big data}: wide physical distribution, diversity of formats, non-standard data models, independently-managed and heterogeneous semantics.
Few related work~\cite{DBLP:journals/dpd/BeheshtiBM16,DBLP:conf/wise/BeheshtiBNA12}, focused on summarizing big process data, by extending Online analytical processing (OLAP) approach to discover concept hierarchies for entities based on both data objects and their interactions in process graphs.

As an ongoing and future work, we are extending our summarization framework to enable explorative querying and understanding of big process data. This is an important line of future work, as understanding process data requires scalable and process-aware methods to support querying, exploration and analysis of the process data in the enterprise because~\cite{DBLP:journals/corr/abs-2105-10911,DBLP:conf/edbt/BeheshtiBM16}: (i)~with the large volume of data and their constant growth, the process data analysis and querying method should be able to scale well; and (ii)~the process data analysis and querying method should enable users to express there needs using process-level abstractions. 
A novel applications would be summarizing process data to personalize recommendations~\cite{DBLP:journals/algorithms/BeheshtiYMGGE20,DBLP:journals/access/YakhchiBGO020}.
In particular, modern Recommendation Systems will require to access and understand the big data built on top of the large data islands. This is important as the growing enhancement in interconnection, storage, as well as data management has made it possible to connect to data deluge from the big data, which in turn, can lead to making intelligent and accurate personalization and recommendations~\cite{elahi2021recommender}.

\subsection{Different Domains and Applications}
Our proposed models to tackle information overload have shown encouraging results, and we conceive their application in many problems where require human and computers to work cooperatively.
Different solutions in various domains were employed in Ch.~\ref{ch_7}, including traffic data, health data and business process data.
The proposed embedding algorithm, Summary2vec, is perhaps well suited in this context as well in many other real-world scenarios. 
Moreover, the produced summaries are all in text format. Making summary vectors for other types of content such as video and/or image is also required.